# SPATIAL PYRAMID CONTEXT-AWARE
# MOVING OBJECT DETECTION AND TRACKING
# FOR FULL MOTION VIDEO AND WIDE AERIAL MOTION IMAGERY

---

A Thesis presented to

the Faculty of the Graduate School

at the University of Missouri

---

In Partial Fulfillment

of the Requirements for the Degree

Doctor of Philosophy

---

by

Mahdieh Poostchi

Dr. Kannappan Palaniappan, Thesis Supervisor

May 2017

The undersigned, appointed by the Dean of the Graduate School, have examined the dissertation entitled:

# SPATIAL PYRAMID CONTEXT-AWARE
# MOVING OBJECT DETECTION AND TRACKING
# FOR FULL MOTION VIDEO AND WIDE AERIAL MOTION IMAGERY

presented by Mahdieh Poostchi,

a candidate for the degree of Doctor of Philosophy,

and hereby certify that, in their opinion, it is worthy of acceptance.

_______________________________________

Dr. Kannappan Palaniappan

_______________________________________

Dr. Filiz Bunyak

_______________________________________

Dr. Jeffrey Uhlmann

_______________________________________

Dr. Guilherme DeSouza

Dedicated to my amazing family for the love, unconditional support,

and constant encouragement I have gotten over the years.

In particular, I would like to thank my dad, my mother, and my sister Hanieh.

You are the salt of the earth, and I undoubtedly could not have done this without you.

# ACKNOWLEDGMENTS

I would like to express my special appreciation and thanks to my advisor Dr. Kannappan Palaniappan, who has been a tremendous mentor for me. He is a talented teacher and passionate scientist. His patient guidance, encouragement and endless enthusiasm showered me in torrents of inspiration throughout my time as his student. I would also like to thank Dr. Guna Seetharaman for his financial support, scientific advice and insightful discussions.

I would like to thank the members of my doctoral committee, Dr. Filiz Bunyak, Dr. Jeffrey Uhlmann and Dr. Guilherme De Souza for the valuable comments they had to improve my dissertation work. I would like to thank Dr. Filiz Bunyak for her valuable insight into my work and for her endless support. Her enthusiasm and love for teaching and mentoring is contiguous. I have had the fortune of being Dr. Uhlmann's student and attend his class where I truly learned about the principles of data fusion techniques and the concept of motion estimation. He is and remains my best role model for a teacher. I would also thank Dr. Guilherme De Souza for his support and valuable suggestions in revising my dissertation.

I would like to thank Dr. Stefan Jaeger and Dr. George Thoma for providing me the opportunity to work on malaria screening project at the Communications Engineering Branch of the Lister Hill National Center, NIH. I had an amazing internship experience working with great people and conducting research in image analysis and machine learning technique for biomedical and clinical systems.

I would like to express my gratitude to Jodie Lenser and Sandra Moore for helping me with my appointment and academics at the Department of Computer Science, University of Missouri-Columbia.

# TABLE OF CONTENTS

# LIST OF TABLES

# LIST OF FIGURES

viii

xi

ABSTRACT

A robust and fast automatic moving object detection and tracking system is essential to characterize target object and extract spatial and temporal information for different functionalities including video surveillance systems, urban traffic monitoring and navigation, robotic, medical imaging, etc. A reliable detecting and tracking system is required to generalize across huge variations in object appearance changes due to camera viewpoint, pose, scale, lighting conditions, imaging quality or occlusions and achieve real-time performance. In this dissertation, I present a collaborative Spatial Pyramid Context-aware moving object detection and Tracking system (SPCT). The proposed visual tracker is composed of one master tracker that usually relies on visual object features and two auxiliary trackers based on object temporal motion information that will be called dynamically to assist master tracker. SPCT utilizes image spatial context at different level to make the video tracking system resistant to occlusion, background noise and improve target localization accuracy and robustness. We chose a pre-selected seven-channel complementary features including RGB color, intensity and spatial pyramid of HoG to encode object color, shape and spatial layout information. We exploit integral histogram as building block to meet the demands of real-time performance. A novel fast algorithm is presented to accurately evaluate spatially weighted local histograms in constant time complexity using an extension of the integral histogram method. Different techniques are explored to efficiently compute integral histogram on GPU architecture and applied for fast spatio-temporal median computations and 3D face reconstruction texturing. We proposed a multi-component framework based on semantic fusion of motion information with projected building footprint map to significantly reduce the false alarm rate in urban scenes with many tall structures. The experiments on extensive VOTC2016 benchmark dataset and aerial video confirm that combining complementary tracking cues in an intelligent fusion framework enables persistent tracking for Full Motion Video (FMV) and Wide Aerial Motion Imagery (WAMI).

# Chapter 1

# Introduction

Image and video content analysis are becoming more popular and complex with the advent of accessible high quality camera sensors and various video analytics applications that enable us to automatically process image or video to characterize and extract temporal and spatial information for different functionalities including filtering [1, 2, 3], video surveillance systems [4, 5], video content retrieval [6, 7], urban traffic monitoring and navigation [8, 9, 10, 11], behavior and activity recognition [12, 13, 14, 15], 3D reconstruction [16, 17], etc.

One of the most common and challenging tasks in a video analytics framework is detecting and tracking moving objects. A robust and fast moving object detection and tracking system is essential to determine meaningful events and suspicious activities in a video surveillance system or to automatically annotate and retrieve video contents in a video content retrieval system. In an urban traffic monitoring system, reliable tracking results can be applied to automatically compute the flux of the vehicles for further road traffic congestion analysis and so many other applications. The objective of this dissertation is to develop a robust, accurate and high performance moving object detection and tracking system for Full Motion Video (FMV) as well as Wide Aerial Motion Imagery (WAMI).

## 1.1    Problem Statement

A robust visual moving object detection and tracking system needs to generalize across huge variations in object appearance that may be observed through full motion videos. These variations usually arise from three sources including object movements, camera motion and background dynamics. As target object moves through the field of view of a camera, the object appearance may change dramatically due to variations in pose, shape and scale or being partially or fully occluded. The 2D shape and appearance of an object may also change substantially when the camera's viewpoint is altered. Capturing images using moving cameras will impose extra challenges on the system unlike imaging with a fixed camera mounted on a wall or a land-pose. Motion blur may occur at the moment of "shot" and can be further magnified by slow exposure [18]. Local sensor noise and compression artifacts are some other difficulties that may impose to system by camera model. Moreover, environment complexities like dynamic background, sudden illumination changes, background clutter, occlusion or shadow interferences are some of the typical challenges that can make moving object detection and tracking processing more complicated [19, 20, 21, 22]. Figure 1.1 elaborates some of these challenges for six selected sequences from 2016 Visual Object Tracking challenge dataset (VOTC2016) [19].

Aerial video provides a global picture of the ground scene over different time

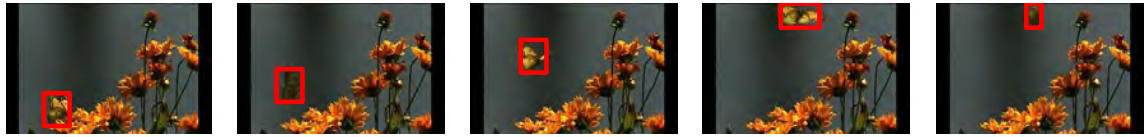

(a) Seq. butterfly, Pose Variation

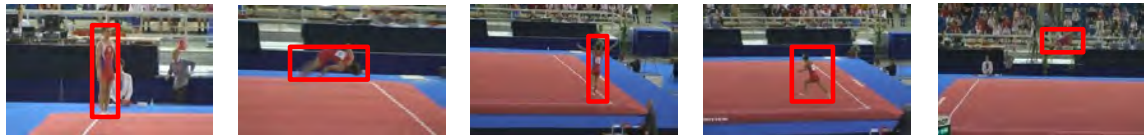

(b) Seq. gymnastics1, Shape Deformation

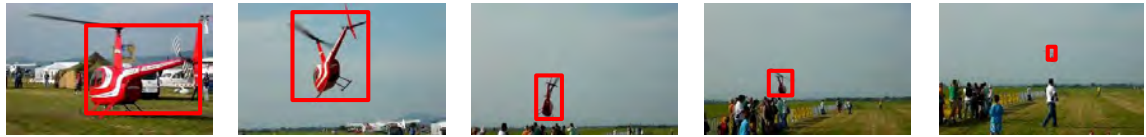

(c) Seq. helicopter, Scale Variation

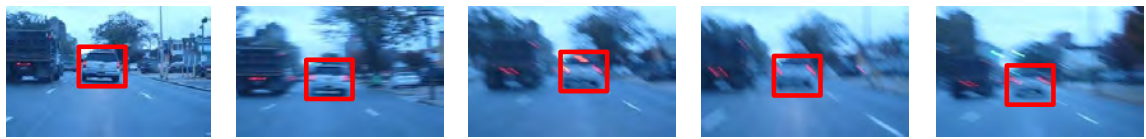

(d) Seq. car1, Motion Blur

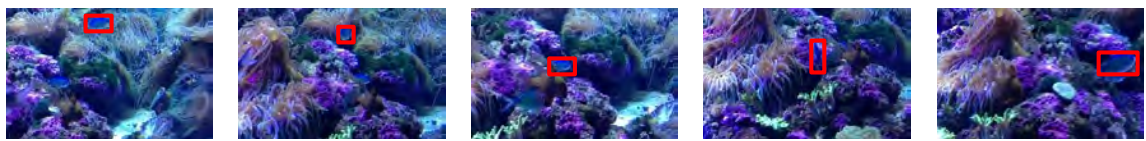

(e) Seq. fish1, Background Clutter

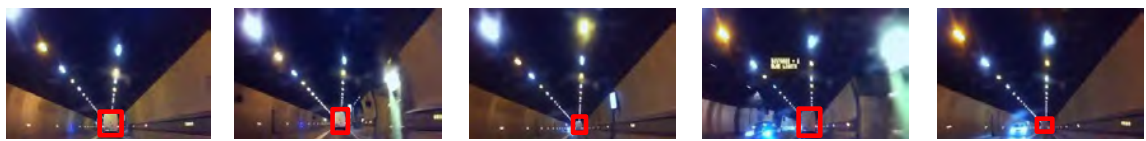

(f) Seq. tunnel, Illumination Changes

Figure 1.1: Typical image processing challenges of a moving object detection and tracking system.

scales. For example, urban aerial imagery captures large scale activity analysis of vehicles and pedestrians in urban settings. Airborne imagery enables understanding the simultaneous behavior of multiple drivers sharing the same road using multi-object tracking, covers a greater variety of interactions between road-users than would be encountered by any one single user, and facilitates routing around accidents to improve traffic flow [23, 24, 25, 8].

However, detecting and tracking moving objects in aerial imagery is impacted by more challenges due to small and low target resolution, large object displacement due to low frame rate, congestion and occlusions, motion blur and parallax effect of tall structures, camera vibration, camera exposure and varying viewpoints, low quality metadata and geo-registration errors [26, 5, 27, 28] in addition to other challenges that we usually face in FMV tracking [29, 30, 31, 32].

Many visual tracking systems are developed to address these challenges including correlation filter-based trackers [33, 22, 34, 35], fragment-based trackers [36, 37, 38], learning-based trackers [39, 40] and Likelihood of features-based tracker [27]. [38, 41] present Flock of Tracker (FoT) that perform target localization by combining displacement estimations provided by independent local trackers laid out on a regular grid covering the object. Despite numerous algorithms that have been proposed, it remains a challenging task to develop a video tracking system that performs persistent tracking accurately and efficiently for both standard Full Motion Video (FMV) and Wide Aerial Motion Imagery (WAMI). Correlation based trackers are usually search for the object within a small neighborhood around the previous estimated location and therefore will lose the object quickly when there is large object displacement due to low frame rate or fast motion particularly in WAMI dataset. Learning-based trackers usually requires a large dataset in order to achieve good performances. My dissertation focused on developing a robust and fast moving object detection and tracking system for Full Motion Videos (FMV) as well as Wide Aerial Motion Imagery (WAMI).

## 1.2 Research Objective

The objective of this dissertation is to develop a robust, accurate and high performance moving object detection and tracking system for Full Motion Video (FMV) as well as Wide Aerial Motion Imagery (WAMI). The input of the systems varies from a standard definition video to very large scale airborne imagery collected over urban areas. The output will be target tracklets that are computed using object visual features and object temporal motion information. Motion prediction will be used to localize the object when being partially or fully occluded by trees or tall structures. The estimated motion detection mask can be fused intelligently with visual object features to increase tracking localization accuracy or being applied to initialize and perform persistent multi-object tracking. Disciplined or informed intelligent fusion of different kinds of information is useful for a general purpose tracking system across modalities and different computer vision tasks. The proposed detection and tracking system needs to be robust to object visual appearance changes due to scale, pose, orientation and illumination. Our developed system main objective is to accommodate to object appearance changes and perform persistent tracking under background noises (clutter, dynamics) and occlusion as well as camera motion effects. Two measures are used to analyze the performance of the visual tracking: Accuracy and Robustness. Accu-racy is the average overlap between the predicted and ground truth bounding boxes during successful tracking periods. Robustness measures the number of times that tracker loses the target during tracking [42, 19, 43]. We weight robustness more than accuracy since the ultimate goal of visual tracking is performing persistent tracking. It is also required to achieve real-time performance on low-power computing platforms (laptops, PCs).

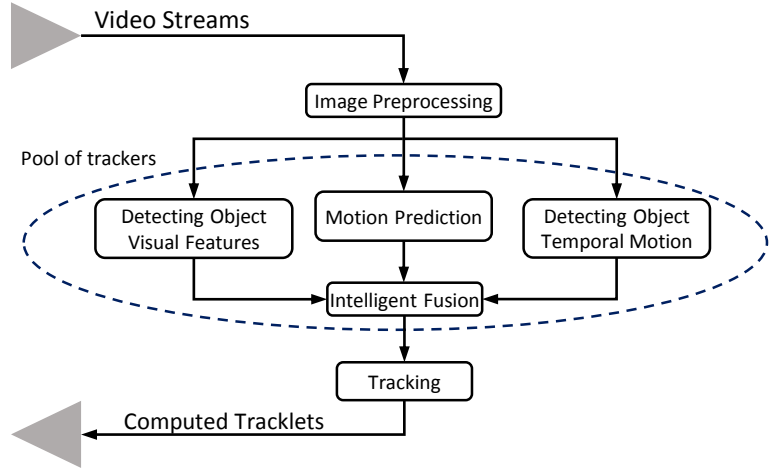

Figure 1.2: Key components of proposed collaborative moving object detection and tracking system. Disciplined or informed intelligent fusion of different kinds of information is useful for a general purpose tracking system across modalities and different computer vision tasks.

## 1.3 Dissertation Layout

In this dissertation we propose a collaborative tracking system consists of a master tracker and two auxiliary trackers. The main idea is to have a pool of trackers that are working together in an intelligent fusion framework to improve tracking performance by being called dynamically. Figure 1.2 illustrates the collection of trackers. *Master* tracker is assigned to the cue that has the most contribution in target localization. The visual feature-based tracker usually takes the lead as long as object is visible and presents discriminative visual features. Otherwise, tracker will be assisted by motion information. For example, motion prediction will be used to localize the object when being partially or fully occluded by trees or tall structures. Accurate temporal motion information can be used to distinguish moving objects from static background in full motion video and filter out false object detections.

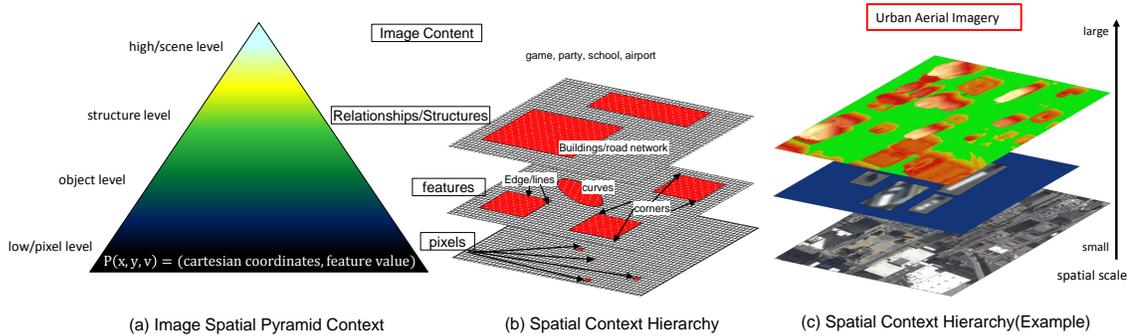

Figure 1.3: Image spatial context can be modeled as a hierarchy of abstractions by increasing the spatial scale. The most left figure shows image spatial context pyramid. Middle image generalizes the spatial scale concept: at the lowest level are raw pixels with color and spatial information in 2D Cartesian coordinates (pixel level). Further processing within a neighborhood yields features that can be interpreted as objects (object-level). At the next higher level, structures are emerged composed of one or more objects and relationships among them and finally at the largest scale (image size) information regarding image content is provided. The right most image presents image spatial context hierarchy for a sample urban aerial imagery.

Image spatial context can be modeled as a hierarchy of abstractions by increasing the spatial scale. Figure 1.3 describes the hierarchy. At the lowest level are raw pixels with color and spatial information in 2D Cartesian coordinates (pixel level). At a higher layer, further processing within a neighborhood yields features such as corners, edges, lines, curves, and color regions. One may combine and interpret these features as objects and their attributes (object-level). At the next higher level, structures are emerged, composed of one or more objects and relationships among them (structure level). Finally, at the largest scale (image size) information regarding image content is provided. We utilizes image spatial context at different level to make our video visual tracking system resistant to occlusion and background noise and improve the robustness. Pixel-level spatial information are used to build intensity spatially weighted histogram or compute object foreground and background color histogram tensor. Spatial layout of image fragments are preserved when constructing the spatial pyramid of HoG. The structure-level spatial context (i.e. road network, building maps) can be applied to filter out the false object detections by distinguishing background from moving objects in full motion videos.

Therefore, our proposed visual tracker is named Spatial Pyramid Context-aware Tracker (SPCT). We chose a 7-channel complementary low-level feature set including RGB color(3), intensity(1), gradient orientation and magnitude(2) and edges(1) to encode target object color, shape and spatial layout information so that to accommodate object appearance changes due to illumination, pose and orientation and maintain real-time performance. Figure 1.4 illustrates the principle tasks of SPCT including tracker cues, target localization and fusion scheme.

The main contribution of the work are summarized as follows:

- **Pool of trackers with a smart context-based fusion scheme**: A collaborative tracking system consists of a master tracker and two auxiliary trackers is developed based on

    - **Multi-channel Features (*Master/Auxiliary*):** that model target appearance, accommodate to appearance changes, incorporating spatial con-

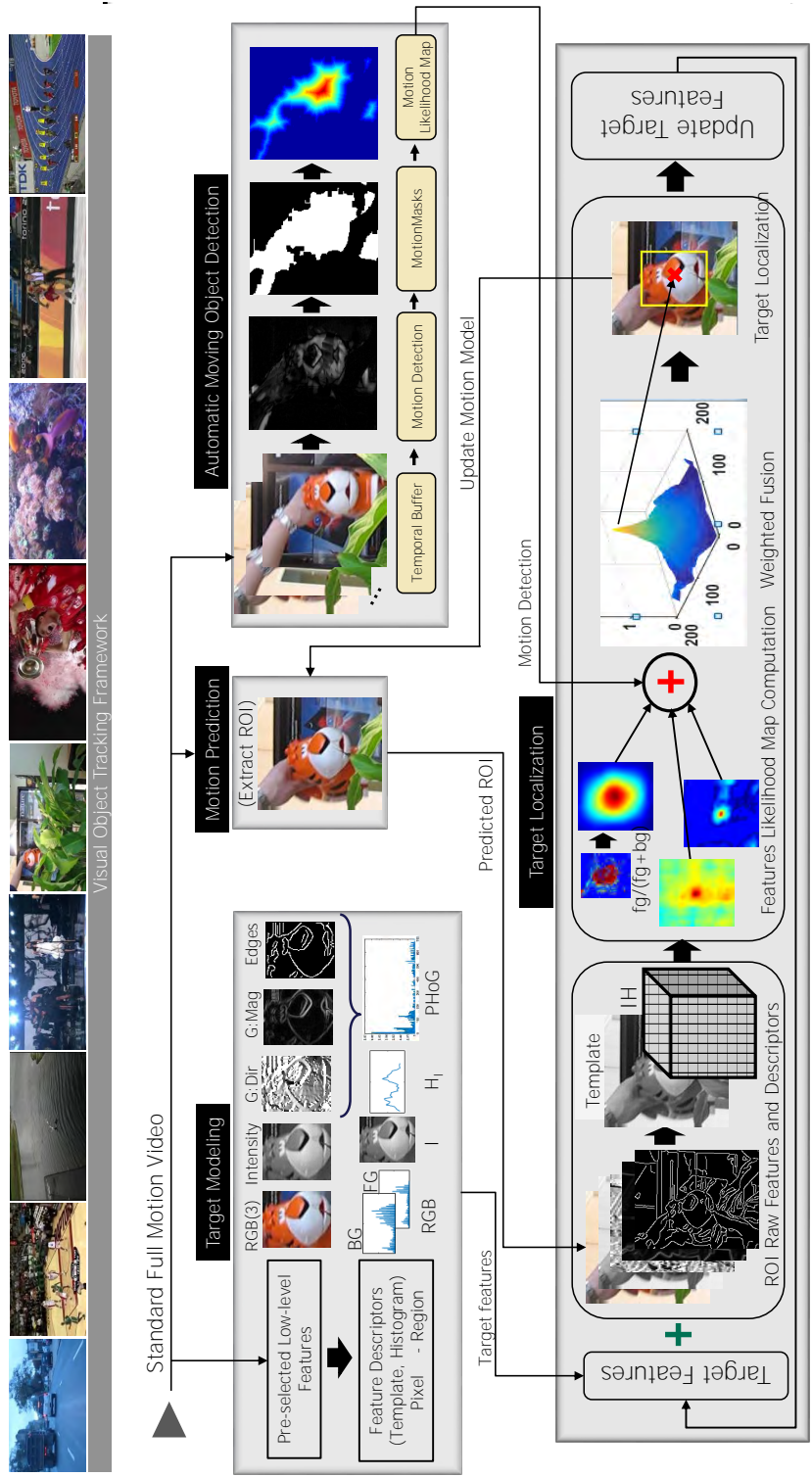

Figure 1.4: Fundamental tasks of the proposed spatial pyramid context-aware moving object detection and tracking system for full motion video.

text and is simple and fast to compute.

- **Motion Detection (*Master/Auxiliary*):** that distinguishes moving objects from the static scene and filters out false detections.

- **Motion Prediction (*Master/Auxiliary*):** that models target motion dynamics to automatically detect Region of Interest (ROI) and improve target localization accuracy within and intelligent fusion framework.

- **Spatial pyramid appearance tracking:** we utilize spatial Pyramid of Histogram of Gradient Orientation (PHoG) to encode object local shape and spatial layout of the shape so that to make tracking resistant to occlusion and invariant to illumination changes.

- **Spatially weighted local histograms in O(1) using weighted integral histogram:** we proposed a novel fast algorithm to accurately evaluate spatially weighted local histograms in O(1) time complexity using an extension of the integral histogram method (SWIH) that encode both spatial and feature information.

- **Parallel GPU implementation of integral histogram:** we utilize integral histogram as the building block to encode candidate regions feature information and achieve fast, multi-scale histogram computation in constant time. Although integral histogram enables fast exhaustive search but it is still considered as the most compute intensive image processing task for the presented tracking system. The sequential implementation of the integral histogram uses an $O(N)$ recursive row-dependent method, for an image with $N$ pixels. Therefore, I explored different techniques to efficiently compute integral histograms on GPU architecture using the NVIDIA CUDA programming model [44, 31].

- **Context-based semantic fusion of motion information with projected building footprint information:** we proposed a multi-component framework based on semantic fusion of motion information with projected building footprint information to significantly reduce the false alarm rate in urban scenes with many tall structures. Moving object detection in wide-area aerial imagery is very challenging since fast camera motion prevents direct use of conventional moving object detection methods and strong parallax induced by tall structures in the scene causes excessive false detections [45, 46].

- **Orientation-Aligned Template Matching by Learning the Object Direction:** Experimental results show that most of the orientation sensitive features fail to detect the object when object template and search window are not aligned for example when computing features likelihood maps using normalized cross correlation of target template and search window. I proposed an orientation-aligned template matching particularly for vehicle detection in wide aerial imagery using vehicle's non-holonomic constraints.

- **Target object initialization refinement using CAMSHIFT:** Many of the tracking systems rely on ground truth annotations for object tracking initialization and re-accusation of the object in case of tracker failure. These information can be used to build the object visual appearance model and localize the target neighborhood. However, the automatic and manual generation of the ground truth is still one of the most tedious and error-prone aspects when developing benchmark data set. If the ground truth bounding box an-notated around the object is not tight, not oriented aligned or not centered around the object (drifted), it will contain background information that will be incorporated into object descriptors which is not desirable. Incorporating background information will lead to less accurate target localization and rapidly loss of the target being tracked. Hence, I used the Continuously Adaptive Mean Shift (CAMSHIFT) algorithm to partially correct the drifted and loose ground truth bounding box and improve tracking robustness.

- **Offline feature selection test-bed using tracking context:** A separate test-bed is developed for filtering-based feature selection in order to decouple feature performance from the rest of the tracking system where the final outcome depends not only on the features used but also on the other parameters like the predictor performance. Based on this experiment, a 7-channel complementary features including RGB(3), gradient orientation and magnitude (2) and edges(1) are chosen to characterize the object appearance model [25, 27].

- **Automatic Detection of candidate regions using motion prediction:** Automatic prediction of ROI in a complex image or video is a key task for visual tracking that enables fast search and avoids background clutter, particularly for large scale aerial imagery. When target motion dynamics is linear or approximately linear during the intervals between observations then a motion prediction filter like Kalman filter can be used to automatically determine the search window in the next image.

- **Top performance on benchmark datasets including VOTC2016 and WAMI data**

Algorithms and methods that are presented in this dissertation are published as our contributions [44, 27, 25, 31, 42, 45, 46, 19, 43, 47, 48, 49, 50, 51, 52, 53].

Chapter 2 presents the key components of the visual tracking system that mainly improved tracking robustness and accuracy. Chapter 3 explains the novel fast algorithm we proposed to accurately evaluate spatially weighted local histograms in O(1) time complexity using an extension of the integral histogram. Chapter 4 discusses different methods that have been studied to distinguish moving objects from static background for full motion video as well as aerial imagery. The GPU implementation of the integral histogram will be presented in Chapter 5. The last chapter concludes work and discusses future directions.

# Chapter 2

# Spatial Pyramid Context-aware Visual Object Tracking

Automatic visual object tracking is an active research area which has many practical applications including video surveillance, video compression, video editing, traffic monitoring, vision-based control, human-computer interfaces, medical imaging, augmented reality, robotics, content-based indexing and retrieval. However, tracking moving objects in videos is very challenging due to background variations, illumination changes, shadow interference, camera vibration and occlusions, local sensor noise and compression artifacts [19].

Tracking in Wide Aerial Motion Imagery (WAMI) is even harder than traditional tracking using standard ground-based Full Motion Video (FMV) due to the problems associated with small and low resolution targets, large moving object displacement due to low frame rate, low quality metadata and georegistration errors, motion blur and parallax effect, camera exposure and camera varying viewpoints [25, 27, 28].

This dissertation presents a visual object tracking system that is composed of one *master* tracker that usually relies on visual object features and two *auxiliary* trackers based on object temporal motion information that will be called dynamically to assist *master* tracker failure. The visual feature-based tracker usually takes the lead as long as object is visible and presents discriminative visual features. Otherwise, tracker will be assisted by temporal motion information. For example, motion prediction will be used to localize the object when being partially or fully occluded by background. This chapter describes the key components of the visual feature-based tracker and the path prediction cue.

## 2.1   Image Pre(Post)-Processing Operations

Some of the challenges that are related to image quality including noise, low contrast, jitters, etc. which are imposed to the system by camera low quality, camera motion and background dynamics can be addressed before modeling the object or extracting motion information to avoid further complexities. Image filtering, contrast enhancement, image stabilization and aerial imagery geo-registration are typical image pre-processing operations that can significantly increase the reliability of the visual tracking system.

**Image Pre-processing Operation**: Gaussian averaging, Median filtering, Bilateral filtering and histogram equalization are the operators that have been applied to the images if required to enhance the image quality. We used a state-of-the-art structure from motion (SfM) and registration algorithm called *MU BA4S* in order to orthorectify image sequences in a global reference system and maintain the relative movement between the moving camera platform and the fixed scene [54, 55].

**Image Post-processing Operation**: Parallax effects (which are particularly severe in dense urban scenes) along with spatial camera-to-camera registration and georegistration errors prevent direct use of detection algorithms relying on motion information through variations of background subtraction and optical flow analysis [28]. Urban scene contextual information is employed to filter out motion parallax induced flow responses and enhance robustness of the system.

## 2.2 Target Tracking Initialization Refinement Using CAMSHIFT

Many of the tracking systems rely on ground truth annotations for object tracking initialization and re-accusation of the object in case of tracker failure. These information can be used to build the object visual appearance model and localize the target neighborhood. However, the automatic and manual generation of the ground truth is still one of the most tedious and error-prone aspects when developing benchmark data set. If the ground truth bounding box annotations around the object is not tight, oriented aligned or not centered around the object (drifted), it will contain background information that will be incorporated into object descriptors which is not desirable. Incorporating background information will lead to less accurate target localization and rapidly loss of the target being tracked.

I used the Continuously Adaptive Mean Shift (CAMSHIFT) algorithm to partially correct the drifted and loose ground truth bounding box and improve tracking robustness. CAMSHIFT is indeed an adaptation of Mean Shift algorithm which is a

---

**Algorithm 1:** CAMSHIFT Algorithm [56]

---

**Input :** Likelihood Map $L$, threshold $\delta$, initial point $(x_0, y_0)$

**Output :** $x_{peak}$

  Estimate $x_{(t=0)} = (x_0, y_0)$, $d_t = \infty$

  Form a Search Window (SW) of $L$ centered at $x_{(t=0)}$

  **while** $d_t < \delta$ **do**

   Calculate zero and first SW moments

   $\quad M_{00} = \sum_x \sum_y P(x, y)$

   $\quad M_{01} = \sum_x \sum_y x \times P(x, y)$

   $\quad M_{10} = \sum_x \sum_y y \times P(x, y)$

   Find $x_t = \frac{M_{10}}{M_{00}}$, $y_t = \frac{M_{01}}{M_{00}}$

   Compute $d_t = (x_t, y_t) - (x_{t-1}, y_{t-1})$

   Form new SW center at $(x_t, y_t)$

   t=t+1

  **end**

  $x_{peak} = (x_t, y_t)$

---

robust, non-parametric method that climbs density gradients to find the peak of the probability distribution [56]. Algorithm 1 describes CAMSHIFT processing. Given a likelihood map, CAMSHIFT will converge to the mean (mode) of the probability distribution by iterating in the direction of maximum increase in probability map. Since the maximum of the likelihood map is around the center of the target, the convergence of the CAMSHIFT will happen around object centroid as well. I will use the location of the peak computed by CAMSHIFT as the new object centroid to refine the tracking initialization information using GT annotations and then recompute the object scale and orientation. The performance of the CAMSHIFT highly relies on the probability map. Target RGB color 3D histogram back-projection is used to compute the likelihood map that associates the pixel values in the image with the value of the corresponding histogram bin. The pixel value of the probability map is in fact the ratio of the target foreground region normalized histogram bin value to the background normalized histogram bin value.

Figure 2.1 shows the extracted target template using the ground truth annotation information and the target initialization refinement process using CAMSHIFT algorithm. Figure 2.2 presents the target bounding box annotation refinement results for some of the selected VOTC2016 sequences.

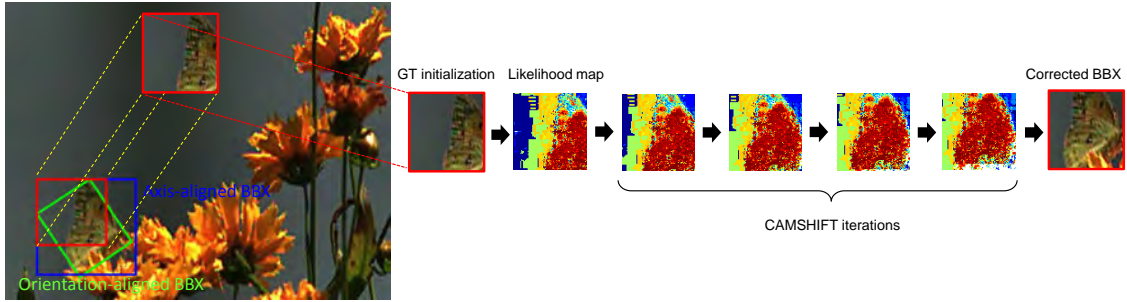

Figure 2.1: Tracking Target initialization refinement using CAMSHIFT algorithm for sequence *butterfly* of VOTC2016 challenge dataset.

I evaluated the overall performance of SPCT on VOTC2016 60 sequences using the refined tracking initialization information. Using refined centered target template bounding box for tracking initialization enable us to improve tracking average robustness performance from 1.4 to 1.3.

## 2.3 Multi-Channel Features for Appearance Modeling

Object model representation is one of the main components in many generative and discriminative visual tracking algorithm. A good tracking system needs to generalize across huge variations in object appearance due to viewpoint, pose, scale, lighting conditions, imaging quality or occlusions. In addition, these tasks should preferably meet the demands of real-time performance on low-power low-latency computing plat-forms. A wide range of features and appearance models have been described in video and target tracking literature to address these challenges [19, 27, 25]. Using a rich collection of features increases tracker robustness but is computationally expensive for real-time applications and localization accuracy can be adversely affected by including distracting features in the feature fusion. I evaluated the performance of a rich set of complementary low level image-based feature descriptors incorporating intensity, edge, texture, and shape information using filtering-based feature selection method. I developed a standalone test-bed to explore offline feature subset selections for video tracking to reduce the dimensionality of the feature space and to discover relevant representative lower dimensional subspaces for online tracking [25].

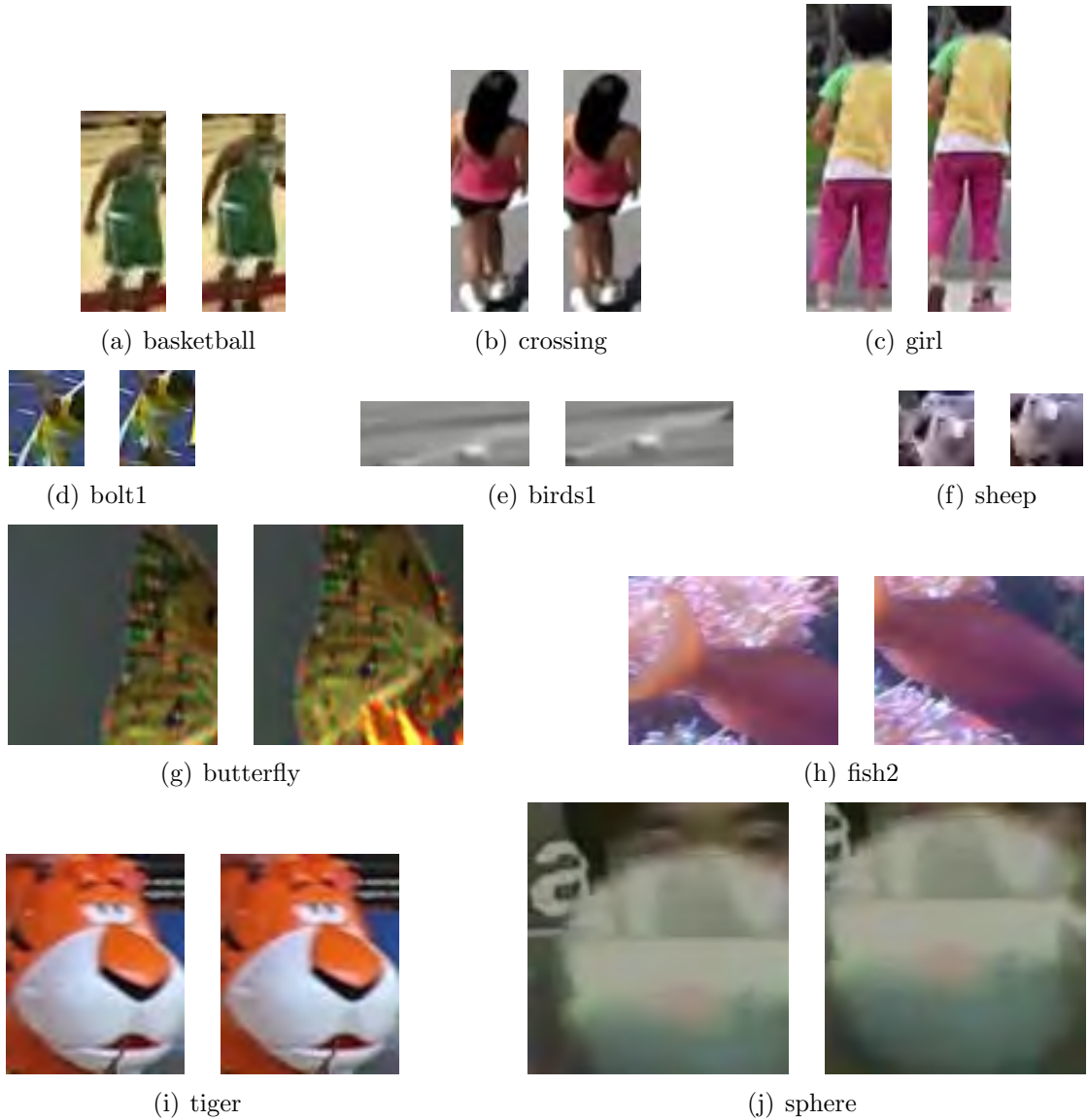

(a) basketball

(b) crossing

(c) girl

(d) bolt1

(e) birds1

(f) sheep

(g) butterfly

(h) fish2

(i) tiger

(j) sphere

Figure 2.2: Tracking Target initialization refinement using CAMSHIFT. Each pair of images presents target template using ground truth annotations on the left and the refined centered target template using CAMSHIFT on the right for some of the sequences from VOTC2016 benchmark dataset [19].

### 2.3.1 Low Level Features

2D Image visual content can be modeled as a hierarchy of abstractions. At the lowest level, we have the raw pixels with color or brightness information. Many statistical measures can be derived by further processing of these features. I applied image fusion using the separable two dimensional steerable Gaussian filter bank operator with pixel-based correlations of image intensity to extract a rich set of multi-scale features. Several features based on structure tensor (first derivatives) and Hessian (2nd derivatives) image operator are computed. These features are grouped into four categories including color, edge-based, local shape-based and texture-based features.

Many local edge and corner descriptors can be directly obtained from image first derivatives like gradient orientation and gradient magnitude as well as from the two eigenvalues $\lambda_1, \lambda_2$ of the structure tensor matrix $\mathbf{J_C}$

$$\mathbf{J_C} = \begin{bmatrix} \sum \left( \dfrac{\partial \mathbf{I}_i}{\partial x} \right)^2 & \sum \dfrac{\partial \mathbf{I}_i}{\partial x} \dfrac{\partial \mathbf{I}_i}{\partial y} \\ \sum \dfrac{\partial \mathbf{I}_i}{\partial x} \dfrac{\partial \mathbf{I}_i}{\partial y} & \sum \left( \dfrac{\partial \mathbf{I}_i}{\partial y} \right)^2 \end{bmatrix} \tag{2.1}$$

$$\lambda_{1,2} = \frac{1}{2}(J_C(1,1) + J_C(2,2) \pm \sqrt{(J_C(1,1) - J_C(2,2))^2 + (2J_C(1,2))^2}) \tag{2.2}$$

Some of the common edge and corner detectors that can be derived from structure tensor are Beltrami [57], Harris [58], Shi-Tomasi [59] and Cumani [60] detectors computed as

$$Beltrami(\mathbf{I}) = 1 + \mathbf{trace}(\mathbf{J_C}) + \mathbf{det}(\mathbf{J_C}) = 1 + (\lambda_1 + \lambda_2) + \lambda_1\lambda_2 \tag{2.3}$$

$$Harris(\mathbf{I}) \;\; = \mathbf{det}(\mathbf{J_C}) - k\,\mathbf{trace^2}(\mathbf{J_C}) = \lambda_1\lambda_2 - k(\lambda_1 + \lambda_2)^2 \tag{2.4}$$

$$ShiTomasi(\mathbf{I}) = min(\lambda_1, \lambda_2) \tag{2.5}$$

$$Cumani(\mathbf{I}) = max(\lambda_1, \lambda_2) \tag{2.6}$$

Local shape-based features are calculated using the eigenvalues $\lambda_{1,2}$ of the Hessian matrix $\mathcal{H}$, of the intensity field $I(x, y)$, that describes the second order structure of local intensity variations around each point of the image

$$\mathcal{H} = \begin{bmatrix} L_{xx} & L_{xy} \\ L_{xy} & L_{yy} \end{bmatrix}, \lambda_{1,2} = \frac{1}{2}(L_{xx} + L_{yy} \pm \sqrt{(L_{xx} - L_{yy})^2 + (2L_{xy})^2}) \tag{2.7}$$

Shape Index (SI) (Eq.2.8) and Normalized Curvature-Index (NCI) (Eq.2.9) are two features that we derived from the eigenvalues, $\lambda_1 \geq \lambda_2$, of $\mathcal{H}$,

$$\theta(x,y) = \tan^{-1} \frac{\lambda_2(x,y)}{\lambda_1(x,y)} \tag{2.8}$$

$$\phi(x,y) = \tan^{-1} \frac{sqrt\lambda_1(x,y)^2 + \lambda_2(x,y)^2}{1 + I(x,y)} \tag{2.9}$$

Another very popular shape measure is the magnitude weighted histogram of Hessian eigenvector orientation. We considered two simple texture measures including gradient magnitude and local binary pattern (LBP) histogram [61] to characterize the quantized local intensity variability. Table 2.3 summarizes the described features.

## 2.3.2 Feature Selection Test-bed Using Tracking Context

I explored offline feature subset selection for video tracking to reduce the dimensionality of the feature space and to discover relevant representative subspaces for online tracking. Good features should be discriminative, robust and easy to compute so that result in improved quantitative and more efficient computational performance, and increased system adaptability and flexibility.

A test-bed that decouples evaluation of the feature selection module from the rest of the tracking system is being developed [27, 25]. The proposed test-bed performs three tasks:

- Computes individual likelihood maps for each feature

- Construct fused likelihood maps for the selected feature subsets

| Candidate Features Set | Features Label | Normalization Range | Number of bins | Equation |
|---|---|---|---|---|
| Intensity histogram | 1: HI | [0,255] | 10 | $I(x,y)$ |
| Gradient Magnitude | 2: HGM | [0,100] | 10 | $GM = \|\nabla_I\| = \sqrt{I_x^2 + I_y^2}$ |
| Shape Index | 3: HSI | [-3*pi/4, pi/4] | 10 | $SI(x,y) = \theta(x,y) = ATAN2(\lambda_{min}, \lambda_{max})$ <br> $\lambda_{min}, \lambda_{max}$ : eigenvalues of Hessian matrix <br> $H = \begin{bmatrix} I_{xx} & I_{xy} \\ I_{xy} & I_{yy} \end{bmatrix}$ |
| Normalized Curvature Index | 4: HNCI | [0, pi] | 10 | $\theta(x,y) = \tan^{-1}\left(\frac{\sqrt{\lambda_{max}(x,y)^2 + \lambda_{min}(x,y)^2}}{1 + I(x,y)}\right)$ <br> $\lambda_{min}, \lambda_{max}$ : eigenvalues of Hessian matrix |
| Hessian Eigenvector Orientations | 5: HEO | [-90 90] | 10 | $\tan^{-1}\left(\frac{V_y}{V_x}\right);$   $V_y = I_{xy}, V_x = \lambda_{max} - I_{yy}$ |
| Gradient Orientation | 6: HOG | [-90 90] | 10 | $\tan^{-1}\left(\frac{I_y}{I_x}\right)$ |
| LBP | 7: HLBP | 16 sampling points on a circle of radius 2 | 17 | $LBP_{P,R} = \sum_{p=0}^{P-1} s(g_p - g_c)2^p, s(x) = \begin{cases} 1, if x \geq 0; \\ 0, otherwise \end{cases}$ |
| Intensity Normalized Cross-correlation | 8: INCC | [0,255] | 10 | $\frac{1}{n}\sum_{x,y}\frac{(I(x,y) - \bar{I})(T(x,y) - \bar{T})}{\sigma_I \sigma_T}$ |
| Gradient Magnitude Normalized Cross-correlation | 9: GMNCC | [0,100] | 10 | $\frac{1}{n}\sum_{x,y}\frac{(GM(x,y) - \overline{GM})(T(x,y) - \bar{T})}{\sigma_{GM} \sigma_T}$ |
| ARST Orientation | 10: HARST | [-90 90] | 10 | based on ST, $J = \begin{bmatrix} I_x^2 & I_x I_y \\ I_x I_y & I_y^2 \end{bmatrix}$ |
| Shape Index Correlation | 11: SINCC | [-3*pi/4, pi/4] | 10 | $\frac{1}{n}\sum_{x,y}\frac{(SI(x,y) - \overline{SI})(T(x,y) - \bar{T})}{\sigma_{SI} \sigma_T}$ |

Figure 2.3: Collection of appearance-based candidate features.

- Evaluate feature subsets based on the corresponding fused likelihood map using the filtering-based methods

In this work, WAMI Persistent Surveillance Systems (PSS) imagery acquired from an eight camera array for Philadelphia is used. Each camera in the array produces an 11 megapixel 8-bit gray scale image typically $4096 \times 2672$ at one to four frames per second. These raw images are georegistered to a $16384 \times 16384$ image mosaic with a ground sampling distance of about $25cm$ for the imagery used in this experiment. Figure 2.4 presents details of the WAMI PSS dataset that have been used for feature performance evaluation.

| Car | Number of frames | Size of data (Number 0f frames * 22) | Car Template |
|---|---|---|---|
| 1 | 254 | 5588 |  |
| 2 | 138 | 3036 |  |
| 3 | 118 | 2596 |  |
| 4 | 266 | 5852 |  |
| 5 | 88 | 1936 |  |
| 7 | 104 | 2288 |  |
| 8 | 92 | 2024 |  |
| 10 | 145 | 3190 |  |
| 11 | 184 | 4048 |  |
| 12 | 35 | 770 |  |
| 13 | 110 | 2420 |  |
| 14 | 167 | 3674 |  |
| 15 | 148 | 3256 |  |
| 16 | 46 | 1012 |  |
| 17 | 47 | 1034 |  |
| Total 15 PSS cars | Total frames = 1942 | Size of data = 42724 | |

Figure 2.4: WAMI Persistent Surveillance System (PSS) dataset details.

In order to decouple feature evaluation from the rest of the tracking system, at each frame $t$, the search window for the target is set to an $m \times m$ region around the true target ground truth position for that frame instead of the predicted target position in the tracking system. Then the performance of the feature subsets can be readily evaluated since the true target location is known and the search region is the same for all feature subsets.

Likelihood maps for individual features are computed using Normalized Cross Correlation (NCC) (Eq. 2.10) or sliding window histogram distance matching using

City-Block distance computation metric (Eq. 2.11, $p = 1$).

$$\gamma[u,v] = \frac{\sum_{k,l}(f[x,y] - \bar{f}_{u,v})(t[x-u, y-v] - \bar{t})}{(\sum_{x,y}(f[x,y] - \bar{f}_{u,v})^2 \sum_{x,y}(t[x-u, y-v] - \bar{t})^2)^{0.5}} \qquad (2.10)$$

$$Dist(H_1, H_2) = \left(\sum_i |H_1(i) - H_2(i)|^p\right)^{\frac{1}{p}} \qquad (2.11)$$

Joint likelihood maps are constructed by fusing individual likelihood maps using weighted sum. Equal weight fusion is used in this study in order to minimize the influence of likelihood fusion approach on feature selection performance. The joint likelihood map for feature subset $X^*$ for frame $t$ is estimated as:

$$L_{X^*}(t) = \sum_{i=1}^{\text{card}(X^*)} w_i \ L_{X_i^*}(t), \quad w_i = \frac{1}{\text{card}(X^*)} \qquad (2.12)$$

where $L_{X^*}(t)$ is the fused likelihood map for feature subset $X^*$ at time $t$, $L_{X_i^*}(t)$ is likelihood map for feature $i$ and $card(X^*)$ is the number of features in the subset. The filtering score for a feature subset is determined by the target localization of its corresponding likelihood map. The likelihood map scoring is done as follows: Likelihood maps typically contain a number of peaks/local maxima. The height of a peak $L(p,t)$ is the likelihood that the target is located at peak position $p$. In the ideal case the highest score for a feature set is when the most likely (highest) peak in the corresponding likelihood map is located on the target (*i.e.* zero distance to target). The scoring process ranks peaks in the fused likelihood map in decreasing order of their heights. The highest peak is labeled as rank 1 and higher ranks are assigned to the other lower confidence peaks (Figure 2.5). Once the peaks are ranked, the score of a likelihood map $L(t)$ is determined by the rank of the highest peak inside the target ground truth region, or is penalized as a miss if there is no local peak present in the target region,

$$score(L(t)) = \begin{cases} rank\left(\arg\max_{p \in R_{GT}}(L(p,t))\right) & \text{if } \exists p \in R_{GT} \\ k+1 & \text{otherwise} \end{cases} \quad (2.13)$$

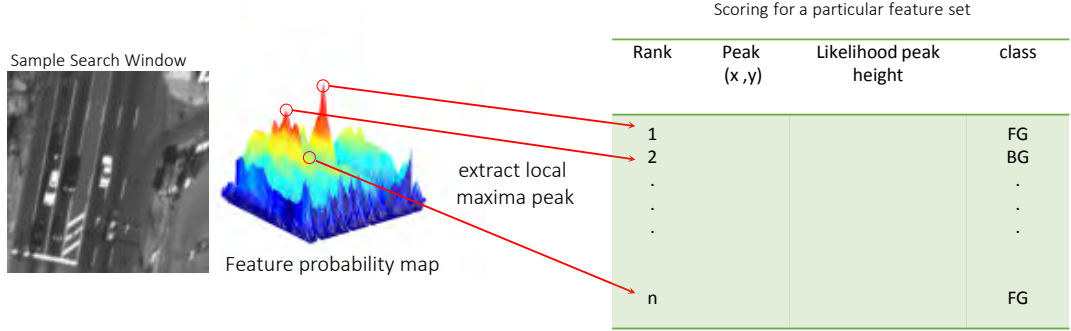

Figure 2.5: Evaluation of a fused likelihood map produced by a feature set. From Left to right: search window, target to sliding window match likelihood map, local maxima (peaks) in the likelihood map and corresponding rank, position and class information.

The score of a feature subset $X^*$ is computed as the average likelihood score over the total number of processed frames where occluded frames are ignored.

$$score(X^*) = \frac{\sum_{t=1}^{\text{card}(frames)} score(L_{X^*}(t))}{\text{card}(frames)} \quad (2.14)$$

I analyzed four selection methods with varying levels of optimality and computational cost. The performance of the exhaustive FOCUS algorithm is compared to the sequential heuristic SFFS, SFS and RELIEF feature selection methods for 11 features described in Figure 2.3. FOCUS which is not practical for larger feature sets because of its exhaustive search, is feasible in this study and produces the best results. The SFFS-based selection has performance similar to greedy SFS selection and both outperform the RELIEF method for the vehicle tracking application. The linear RELIEF unlike the other feature selection methods which produce the same results each time, uses random class sampling so the resulting weights change from run to run. Overall SFFS and SFS perform very well, close to the optimum determined by FOCUS, but RELIEF does not work as well for feature selection in the context of appearance-based object tracking.

The observed results illustrate that there are many factors that affect the performance of a particular feature. Image resolution, target size, target color (i.e. light or dark car) and background complexities change from one car to the other car. The obtained results validate that the intensity and the gradient magnitude (both histogram and correlations) are the most discriminative features for the light cars. While in the case of dark cars, additional feature descriptors like histogram of gradient orientation (HOG) and histogram of eigenvector orientation (HEO) and Linear Binary Pattern (LBP) are required.

Therefore, I have chosen the most efficient features including RGB color, intensity and HoG to accommodate the object appearance changes instead of using a collection of multiple features that will provide similar information or sometimes distracting other features during the fusion.

### 2.3.3   Target Appearance Modeling

Appearance-based tracking algorithm that use color and HOG features achieve state-of the art performance while maintaining real-time speed. Image color histogram is one of the simplest method to achieve robustness to deformation. Color or intensity histograms are insensitive to shape variation, but are often not sufficient to discriminate object from background and are also sensitive to significant illumination changes. On the other hand, gradient magnitude weighted histogram of orientation presents to be a very strong feature that is robust to illumination changes and represents image shape. Therefore, a 7-channel low-level complementary features including RGB color(3), intensity(1), gradient magnitude(1), gradient orientation(1) and edges(1) are selected to fully characterize the object and accommodate to appearance changes.

Either pixel-wise (target template) or region-wise (histogram) descriptors are used to represent target model. Templates provide good spatial localization and discriminative power but are sensitive to change in pose, viewing angle and scale. Therefore, I incorporate histogram-based descriptors to obtain global information about the object that are robust to change due to motion, pose or viewing angle using appropriate normalization and alignment operations.

**Target Template**: Target template carries both spatial and appearance information of the object. However, templates only encode the object appearance that is generated from a single view. Therefore, it is more suitable for problems where the viewing angle of the camera and the object pose remains constant or change very slowly like modeling moving objects in aerial imagery.

**3D Histogram of RGB color**: Our RGB color descriptor consists of two three dimensional histograms of size $32 \times 32 \times 32$ to represent foreground object and its background information. This information are used later for pixel-wise RGB color likelihood map computation using foreground to background histogram ratio (Figure 2.6).

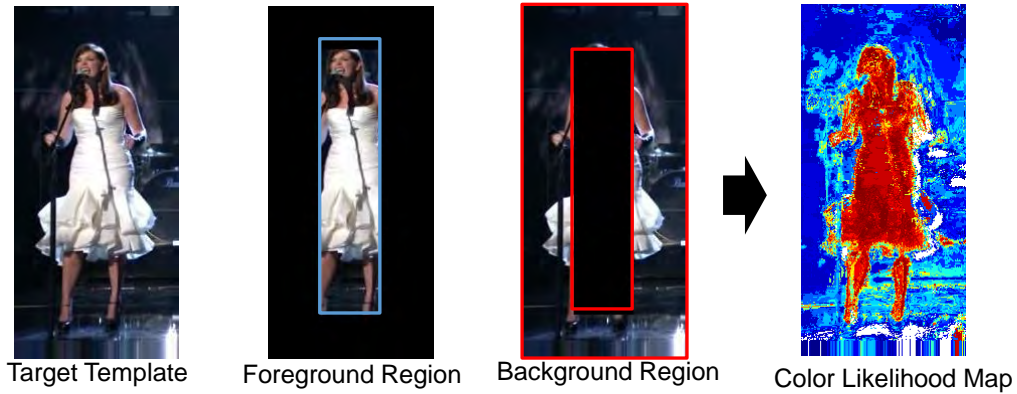

Target Template   Foreground Region   Background Region   Color Likelihood Map

Figure 2.6: Target foreground and background histogram computations to compute pixel-wise color likelihood map using foreground to background color histogram ratio.

**Spatial Pyramid of Histograms of Gradient Orientation (PHoG)**: Using higher resolution feature descriptors for representing the object is essential to improve the system performance, particularly to address occlusion and appearance changes. With this approach, coarse feature descriptor covers the entire object at the lowest level $L_1$, while smaller patches of the object are covered at higher resolution pyramid level. Therefore, the matching process does not only rely on the entire object but the smaller patches and their spatial order. We have used *PHoG* to represent the object by its local shape and spatial layout of the shape [62, 63]. Local shape is obtained by the distribution over edge orientation within a region, and spatial layout by tiling the image into regions at multiple resolutions. The descriptor consists of a magnitude weighted histogram of gradient orientation over image subregion at different resolution level. Figure 2.7 illustrates the spatial pyramid of HoG feature computation for sequence *Singer1* from *VOTC*2016 dataset.

**Spatially Weighted Histogram of Intensity**: Instead of plain histograms that are sensitive to noise and occlusions, we modeled the object brightness information by a spatially weighted histogram. The idea is to assign lower weights to the pixels that most likely belong to background or occluding objects. We proposed a novel fast algorithm to accurately evaluate spatially weighted local histograms in O(1) time complexity using an extension of the integral histogram method (SWIH) that encodes both spatial and feature information (refer to Chapter 3).

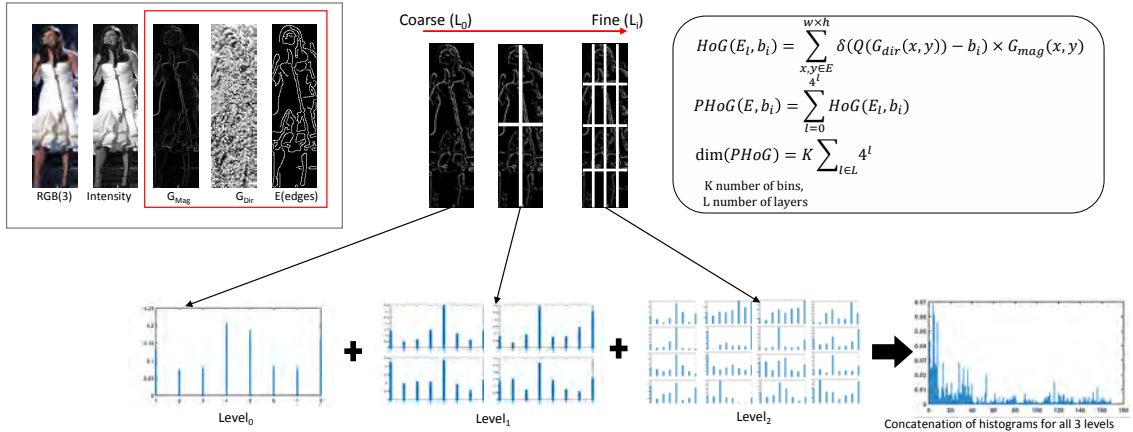

Figure 2.7: Compute Spatial Pyramid Magnitude Weighted Histogram of Gradient Orientation over Edge Image.

### 2.3.4    Candidate Region Descriptors

The candidate region or search window is represented either by its template for pixel-wise matching or an integral histogram tensor will be constructed for each region-based feature. Integral Histogram is our building block to extract multi-scale local histograms in constant time.

Integral histogram is a recursive propagation preprocessing method used to compute histograms over arbitrary rectangular regions in constant time [64]. The integral histogram at position $(x, y)$ in the image holds the histogram for all the pixels contained in the rectangular region defined by the top-left corner of the image and the point $(x, y)$ as shown in Figure 2.8. The integral histogram for a rectangular 2D region defined by the spatial coordinate $(x, y)$ and bin variable $b$ is defined as:

$$H(x, y, b) = \int_0^x \int_0^y Q(I(r, c), b) dr \; dc \approx \sum_{r=0}^x \sum_{c=0}^y Q(I(r, c), b) \qquad (2.15)$$

where $Q(I(r, c), b)$ is the binning function that evaluates to 1 if $I(r, c) \in b$ for the bin $b$, and evaluates to 0 otherwise. Given $H$, computation of the histogram for any candidate region $R$ (Fig. 2.8) reduces to the combination of four integral histograms:

$$h(R, b) = \; H(r^+, c^+, b) - H(r^-, c^+, b) - H(r^+, c^-, b) + H(r^-, c^-, b) \qquad (2.16)$$

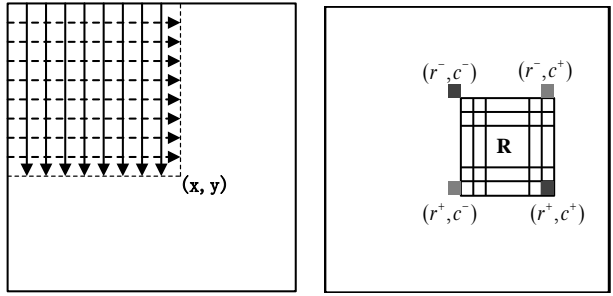

Figure 2.8: Computation of the integral histogram up to location $(x, y)$ using a cross-weave horizontal and vertical scan on the image (left). Computation of the histogram for an arbitrary rectangular region $R$ is shown on right image (origin is the upper-left corner with y-axis horizontal).

However, the conventional local histogram computations using integral histogram does not directly apply to compute spatially weighted local histograms and requires more processing to build the spatial L-level pyramid high resolution histograms of gradient orientations.

## Spatial L-level Pyramid Histograms of Gradient Orientations Using Integral Histogram

The main integral histogram property is to compute multi-scale local histograms with two subtraction and one addition operations. Using this feature enables us to construct spatial L-level pyramid of HoG in few steps. Let's assume that we want to generate the $L$ level pyramid histograms of sliding windows of size $w \times h$ within a search window of size $W \times H$(Figure2.9).

At the lowest level $L_0$, we compute the coarse local histograms of size $w \times h$. For the higher resolution histograms at each pyramid level $L_i$, first we compute the smaller subregions histograms of size ($w_i = \frac{w}{2^i}, h_i = \frac{h}{2^i}$) in constant time by three vectorized arithmetic operations. In the second step, for each region of size $w \times h$, we combine the corresponding subregions histograms of size ($w_i, h_i$) considering their translations

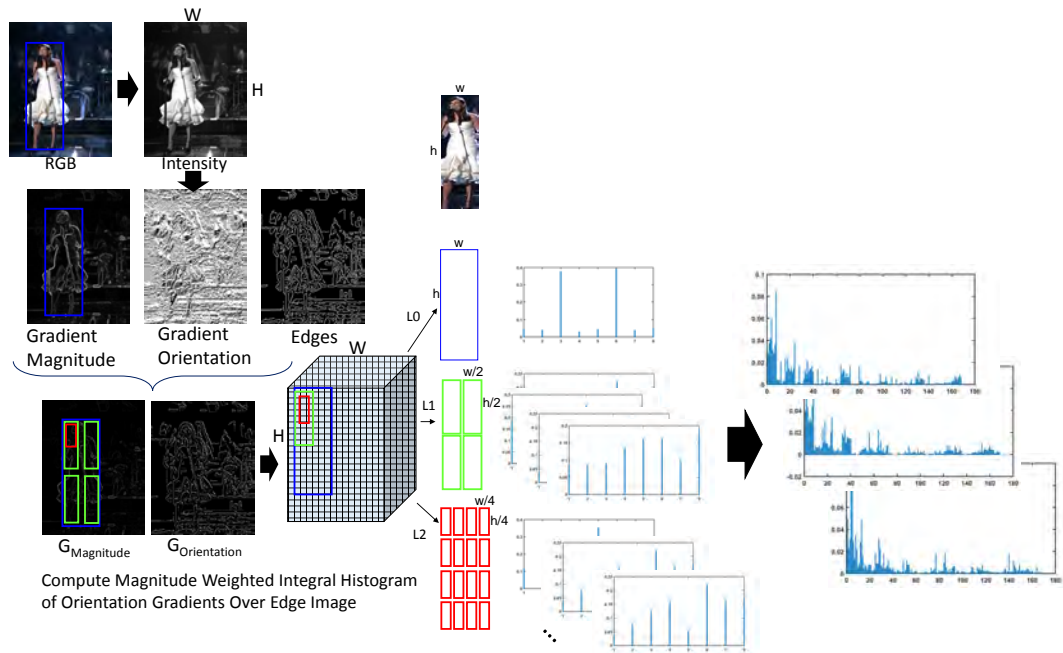

Figure 2.9: Spatial Pyramid HoG Feature Computation For Search Window.

**Algorithm 2:** Computing Spatial L-level Pyramid Histograms of Gradient Orientations Using Integral Histogram

**Input :** Search Window(SW) of size $H \times W$, image chip of size $w \times h$, number of bins $b$, number of pyramid level $L$

**Output :** $SPHoG$: Spatial L-level Pyramid Local Histograms of Gradient Orientations

1: Compute SW edges using Canny edge detection
2: Compute SW gradient magnitude and orientations
3: Quantize gradient orientations over edge regions into $b$ bins
4: $IH =$ Compute gradient magnitude weighted integral histogram of gradient orientation over edges
5: //Compute spatial L pyramid level gradient histogram of orientation
6: **for** $l = 0 : L$ **do**
7:     $s = 2^l$
8:     $h_i = \frac{h}{s}; w_i = \frac{w}{s};$
9:     $H_l =$ Compute-Local-Histograms-Using-IH($IH, w_i, h_i$)
10:     $ind_i = 1; ind_j = 1;$
11:     **for** $i = 1 : s$ **do**
12:        **for** $j = 1 : s$ **do**
13:           $sub_{H_l} = H_l(ind_i : h + ind_i - 1, ind_j : w + ind_j - 1, :)$
14:           $SPHoG = SPHoG + sub_{H_l}$
15:           $ind_i = ind_i \times h;$
16:        **end for**
17:        $ind_i = 1;$
18:        $ind_j = ind_j \times w;$
19:     **end for**
20: **end for**

from the region center. Therefore, we will use the same integral histogram tensor $L$ times to construct the spatial L-level pyramid local histograms. Algorithm 2 describes the local pyramid histograms computations.

## 2.4 Feature Likelihood Maps Computation

For every frame $I_t$, matching likelihood maps for individual features are computed based on the feature characteristics and descriptors. Pixel-based normalized cross correlation of image chip provides good spatial localization and discriminative power but are sensitive to change in pose, viewing angle and scale. Therefore, we incorporate histogram-based descriptors to obtain global information about the object that are robust to change due to motion, pose or viewing angle using appropriate normalization and alignment operations. Pixel-wise foreground to background histogram ratio and sliding window histogram similarity operator are used to compute the histogram-based likelihood maps.

### 2.4.1 Normalized Cross Correlation

Normalized Cross Correlation (NCC) is one of the most popular methods to perform template matching. Providing a reference image of an object (the template image) and a candidate image chip, correlation-based methods identify all candidate regions that match the predefined template to localize the object. NCC computation is simple, effective and invariant to linear brightness and contrast variations. We used Matlab fast normalized cross correlation function based on integral images [65] to perform target template similarity matching (Eq. 2.17)

$$\gamma[u,v] = \frac{\sum_{k,l}(f[x,y] - \bar{f}_{u,v})(t[x-u, y-v] - \bar{t})}{(\sum_{x,y}(f[x,y] - \bar{f}_{u,v})^2 \sum_{x,y}(t[x-u, y-v] - \bar{t})^2)^{0.5}} \qquad (2.17)$$

It is been seen that correlation-based likelihood maps provide more accurate peak-like response for object localization compare to histogram-based matching. The reason is that NCC uses local sums to normalize the cross-correlation, giving high likelihoods within a small local region around the center of the object window and low likelihood values for the rest of the object window [66]. We applied NCC to compute intensity feature likelihood map. However, NCC fails to localize the target when there is significant rotation and scale changes. Therefore, we incorporate histogram-based matching that are robust to translation, pose or viewing angle changes.

## 2.4.2   RGB Color Foreground to Background Histogram Ratio

Pixel-wise color Likelihood maps are simple and fast to compute, robust to deformation and discriminative when background color contrasts sufficiently with object.

Bays classifier is used to compute the pixel-wise RGB color likelihood map using object foreground and background region information prior knowledge, $R_{FG}$ and $R_{BG}$ respectively, derived from the ground truth position of the target $X_{GT} = (C_x, C_y)$ (Figure. 2.6). RGB color histogram $H_{FG}(b_R, b_G, b_B)$ and $H_{BG}(b_R, b_G, b_B)$ of size $32 \times 32 \times 32$ over $R_{FG}$ and $R_{BG}$ region are computed to model foreground object and its background

$$H_{ROI}(b_R, b_G, b_B) = \sum_{(x,y) \in R_{ROI}} \delta((R(x,y) - b_R) \times (G(x,y) - b_G) \times (B(x,y) - b_B)) \quad (2.18)$$

so that the RGB color components at location $(x, y)$ are encoded by the same bin index set $(b_R, b_G, b_B)$. The probability of foreground and background color information is approximated using foreground and background histogram information and the areas $|R_{FG}|$ and $|R_{BG}|$:

$$P(b_{RGB}(x,y)|(x,y) \in R_{BG}) = \frac{H_{BG}(b_{RGB}(x,y))}{|R_{BG}|} \quad (2.19)$$

$$P(b_{RGB}(x,y)|(x,y) \in R_{FG}) = \frac{H_{FG}(b_{RGB}(x,y))}{|R_{FG}|} \quad (2.20)$$

Then for a given pixel at location $(x, y)$, the probability of belonging to foreground

region is

$$L(x, y) = P((x, y) \in R_{FG} | b_{RGB}(x, y)) =$$

$$\frac{P(b_{RGB}(x, y) | (x, y) \in R_{FG}) \times P((x, y) \in R_{FG})}{P(b_{RGB}(x, y) | (x, y) \in R_{FG}) \times P((x, y) \in R_{FG}) + P(b_{RGB}(x, y) | (x, y) \in R_{BG}) \times P((x, y) \in R_{BG})}$$

$$\frac{H_{FG}(b_{RGB}(x, y))}{H_{FG}(b_{RGB}(x, y)) + H_{BG}(b_{RGB}(x, y))} \tag{2.21}$$

We applied the same method to evaluate the performance of different color spaces including RGB, HSV and YCBCr. The results are presented in Section 2.8.2.

### 2.4.3 Sliding Window Histogram Matching

Histogram-based representation is widely applied in many image processing tasks including detection, tracking and recognition due to its simplicity and rich discriminative information. Given two image chips, the reference and the candidate image, we compute the histogram of each and then different histogram distance matching operators including bin-to-bin distances and cross-bin distances will be applied to compute the similarity between two images. The bin-to-bin distances between two histograms are based on the differences of the corresponding bins in the histograms including $\ell_1$ and $\ell_2$ distances, histogram intersection and Bhattacharyya coefficient between two histograms [67]. The cross-bin similarity measures perform cross bin comparison between two histograms to obtain more robust measure of their similarities including earth movers distance (EMD) [68].

We applied the city block distance metric to compute intensity histogram likelihood map that is a special case of the Minkowski bin-to-bin similarity metric (Eq. 2.22), where p=1.

$$Dist(H_1, H_2) = \left( \sum_i |H_1(i) - H_2(i)|^p \right)^{\frac{1}{p}} \tag{2.22}$$

### 2.4.4 Spatial Pyramid Matching Using the Mercer Kernel

Using high resolution L-level PHoG descriptor enables us to encode the entire object feature information at the lowest level $L_0$ and the small object patches at higher levels. The image grid at level $l$ has $2^l$ cells along each dimension, for a total of $D = 2^{k \times l}$, where $k$ is the histogram bin size. The final PHoG descriptor is the concatenation of HoG vectors for the grids across all levels with dimensionality of $k \sum_{l \in L} 4^l$.

Similarity between a pair of PHoGs $X$ and $Y$ is computed using Mercer Kernel method as described in [63]. Spatial pyramid matching using the Mercer kernel assign higher weights to the matches that are found at finer resolution than matches found at coarser resolution. Let $H_l^X$ and $H_l^Y$ represents the histograms of $X$ and $Y$ at level $l$ and $H_l^X(i)$ and $H_l^Y(i)$ are the number of points from $X$ and $Y$ that falls into $i^{th}$ cell of the grid. Then, matches at level $l$ is obtained by the histogram intersection function as

$$I^l = I(H_l^X, H_l^Y) = \sum min(H_l^X(i), H_l^Y(i)) \tag{2.23}$$

Finally, the pyramid match kernel is computed as follows:

$$\kappa^L(X, Y) = I^L + \sum_{l=0}^{L-1} \frac{1}{2^{L-l}}(I^l - I^{l+1}) = \frac{1}{2^L}I^0 + \sum_{l=1}^{L} \frac{1}{2^{L-l+1}}I^l \tag{2.24}$$

## 2.5 Orientation-Aligned Template Matching by Learning the Object Direction

Experimental results show that most of the orientation sensitive features fail to detect the object when object template and search window are not orientation aligned for example when computing features likelihood maps using normalized cross correlation of target template and search window. Figure 2.10 illustrates the template matching likelihood map results using normalized cross correlation of target template and region of interest (ROI) for a vehicle detection and tracking system. As it is shown in the first row of Figure 2.10, at time $t = 1$ the detection response perfectly localize the object since both target template and ROI are vertically aligned. However, the direction of the vehicle may gradually change as it moves along the image sequences. Assuming that at frame $t$, the target template and ROI lose their alignment due to vehicle turn. As it is shown in second row of Figure 2.10, NCC fails to detect and localize the vehicle since it is sensitive to orientation changes.

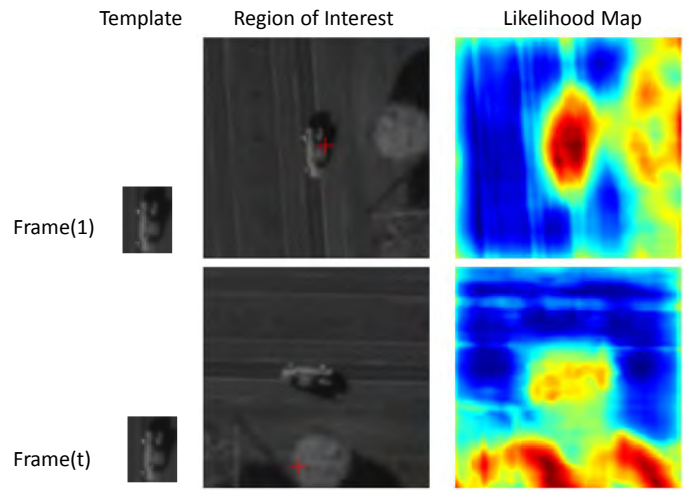

Figure 2.10: Orientation sensitivity performance evaluation of intensity Normalized Cross Correlation (NCC) likelihood map. NCC fails to detect and localize the object when the target template and region of interest lose their orientation alignment due to object movements and orientation changes.

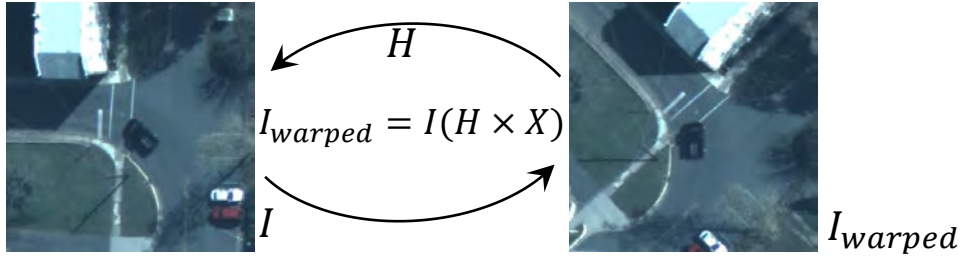

Figure 2.11: Compute transformed ROI using target true position and orientation.

We conducted two different experiments to evaluate the performance of orientation sensitive features by rotating the target template or ROI. In both experiments ROI is cropped around the true target ground truth position so that the ideal detection response and the likelihood map peak should located at the center of ROI. Initially, target template is being aligned horizontally or vertically based on its original direction. Different methods are investigated to estimate the vehicle orientation including radon transform, object motion dynamics and CAMSHIFT algorithm. In the first experiment, at every iteration target template is rotated 5 degree CCW while the ROI is fixed and aligned with target true orientation. In the second run, target template remains fixed and at every iteration the ROI is rotated 5 degree CCW. Inverse warping is used if needed to align search window using homography transformation matrix $H = H_1 \times H_2 \times H_3$ based on target true position $(C_x, C_y)$ and orientation $\theta$, as shown in Figure 2.11, where

$$H_1 = \begin{bmatrix} 1 & 0 & -C_x \\ 0 & 1 & -C_y \\ 0 & 0 & 1 \end{bmatrix}, H_2 = \begin{bmatrix} cos(\theta) & -sin(\theta) & 0 \\ sin(\theta) & cos(\theta) & 0 \\ 0 & 0 & 1 \end{bmatrix}, H_3 = \begin{bmatrix} 1 & 0 & +C_x \\ 0 & 1 & +C_y \\ 0 & 0 & 1 \end{bmatrix} \quad (2.25)$$

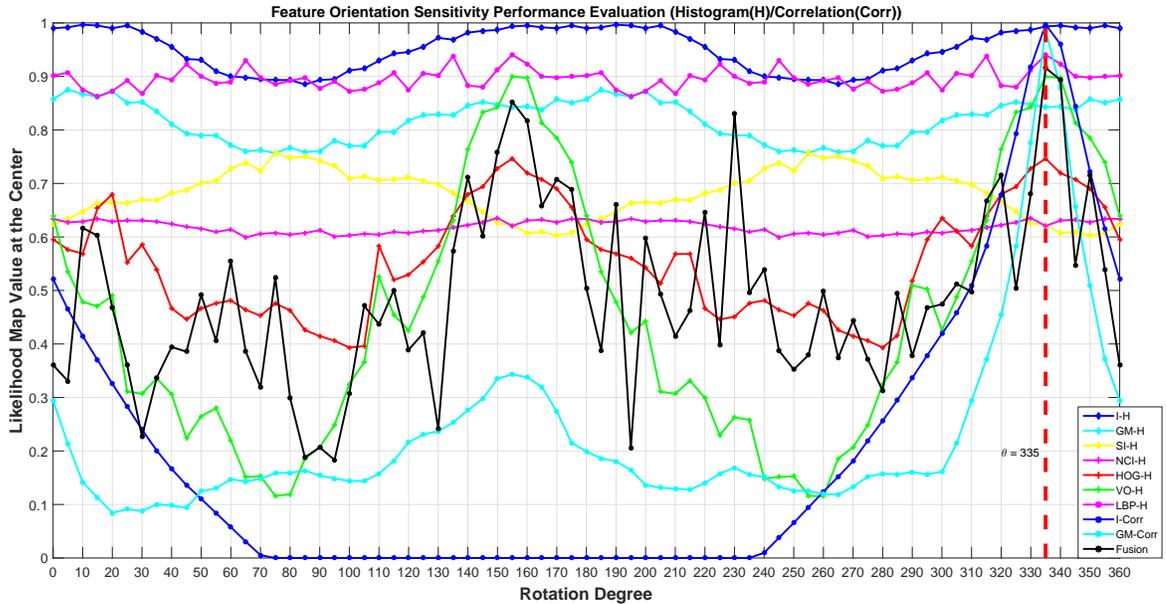

Figure 2.12: Features orientation sensitivity performance evaluation.

Figure 2.12 presents orientation sensitivity performance evaluation results using sliding window histogram matching and normalized cross correlation template matching. The experiments are conducted on frame 1 of sequence *C5_4_1R* from CLIF dataset using fixed aligned target template and rotating ROI. Radon transform is used to estimate the vehicle orientation. We considered 9 individual features including histogram of intensity (I-H), gradient magnitude (GM-H), shape-index(SI-H), normalized curvature index (NCI-H), eigen-vector orientation (VO-H), linear binary pattern (LBP-H), correlation of intensity (I-Corr) and gradient magnitude (GM-Corr). Moreover, we fused features likelihood maps using equal weighting and presents the results (*Fusion*). The estimated object orientation is 335 degree and as we expected the orientation sensitive operations including NCC performs well when the template and ROI are getting aligned at 335 degree ( Figure 2.12, vertical red-line).

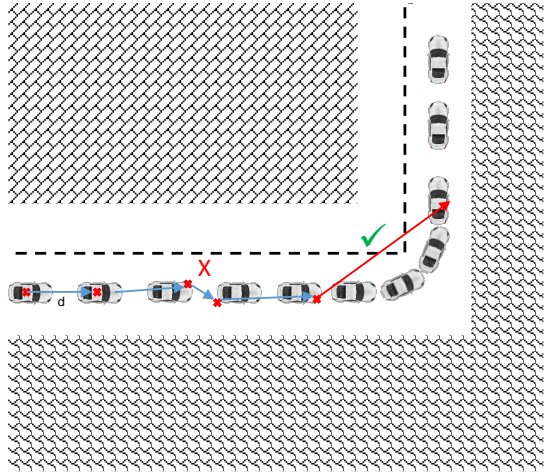

Figure 2.13: A vehicle displacement is always parallel to the path due to its non-holonomic constraints.

Experimental results proved the idea that likelihood map will have the maximum peak value at the center when target template and ROI are completely aligned particularly for rigid vehicle detection in aerial imagery. Therefore, we proposed an orientation-aligned template matching module based on vehicle's non-holonomic constraints (Figure 2.13).

## 2.5.1 Learning the vehicle direction

Accurate object orientation estimation is the most challenging task that affect the aligned matching performance. I have tested radon transform, CAMSHIFT algorithm and motion dynamics to estimate the orientation of the vehicle and perform orientation aligned matching at every iteration of the tracking. However, each of them has its own drawback. Background clutter and noise will adversely affect orientation estimation using radon transform and CAMSHIFT algorithm.

As it is shown in Figure 2.13, a vehicle displacement is always parallel to the path

**Algorithm 3:** Compute vehicle new direction

**Input :** $Prev_{Dir}, minDist$
**Output :** $New_{Dir}$

1: //compute distance between current target location and the previous frame
2: $curr_{dist}$ = compute_distance(position($curr_{frame}$),position($curr_{frame}$-1));
3: // set previous frame two frames before the current frame
4: $prev_{frame} = curr_{frame} - 2$;
5: // compute distance to the previous frame until pass minDist
6: **while** $curr_{dist} <$ minDist and $prev_{frame} \geq Init_{frame}$ **do**
7:     $curr_{dist}$ = compute_distance($curr_{frame}$,$prev_{frame}$)
8:     $prev_{frame} = prev_{frame} - 1$;
9: **end while**
10: // compute New_Dir if conditions passed
11: **if** $prev_{frame} > init_{frame}$ **then**
12:     $New_{Dir}$ = compute_direction(rvel,cvel);
13:     **if** $Prev_{Dir} == $ -1 **then**
14:         $Prev_{Dir} = New_{Dir}$;
15:     **end if**
16: **end if**

due to its non-holonomic constraints. Therefore if we estimate the direction of the path, in fact we have estimated the orientation of the vehicle that can be used if required to transform the target reference template.

Algorithm 3 describes the proposed vehicle direction learning algorithm using estimated target tracklets. We experimentally defined a minimum distance threshold value to avoid inaccurate orientation estimation due to small displacement. This threshold value set to be double of target vehicle maximum length ($minDist = 2 \times max(w, h)$). At every iteration the new orientation of the vehicle is estimated using the most displacement along $X$ and $Y$ axis. Then if the new computed direction is not equal to the previous direction, the target template orientation update function will be called. The proposed vehicle direction learning method significantly improves the visual tracking performances when car start turning.

## 2.6 Tracking Using Motion Prediction

In many visual tracking applications users typically concentrate on some specific monitored objects, such as a vehicle, a person, a face, a ball etc. Hence, only the region around this object is important that can be considered as a fixed or dynamic region of interest (ROI) within the image. Automatic prediction of ROI in a complex image or video is a key task for visual tracking that enables fast search and avoids background distractors, particularly for large scale aerial imagery [27, 28]. Many of the state-of-the art trackers perform their tracking procedure in a search area centered at the position estimated in the previous image, assuming that the object doesn't have large displacement. On the other hand, when target motion dynamics is linear or approximately linear during the interval between observations then a motion prediction filter can be used to determine the search window in the next frame. Kalman filter and its extensions are common motion prediction methods that are being used in many tracking systems [69, 70, 71]. Kalman filter relies on a motion model that represents how object moves in time. One can estimate system's dynamical evolution characteristics, matrix $F$, based on at least two first frames. Let's assume that our system has an state estimate $X$ and covariance matrix $P$ and constant velocity, then at time $t = 0$:

$$X(0|0) = [\ C_x \quad C_y \quad V_x \quad V_y\ ]^T \tag{2.26}$$

$$P(0|0) = \begin{bmatrix} \alpha & 0 & 0 & 0 \\ 0 & \alpha & 0 & 0 \\ 0 & 0 & \alpha & 0 \\ 0 & 0 & 0 & \alpha \end{bmatrix} \tag{2.27}$$

where $\alpha$ is a small value at the beginning when initializing the tracker. An additional covariance matrix $Q$ is considered to account for the amount of uncertainty associated with motion model, $F$. The process noise matrix $Q_{const}$ is a powerful parameter that can be adjusted to allow a linear model to adopt to the dynamics changes of a highly nonlinear system. Since the time interval between two consequences frame is 1 ($\triangle t = 1$), Q is a constant matrix set to small values that will be added to state covariance matrix $P$ to tune the filter. Now, we can estimate the location of the object at time $t + 1$ using Kalman prediction:

$$
\begin{aligned}
X(t+1|t) &= FX(t|t) \\
P(t+1|t) &= FP(t|t)F' + Q_{const}
\end{aligned}
$$

(2.28)

$X(t + 1|t)$ will be used to detect the region of interest in frame $t + 1$. Then target location $Z(t + 1)$ and related covariance matrix $R(t + 1)$ will be estimated using computed fused features likelihood map on detected ROI:

$$
Z(t+1) = [Z_x, Z_y]
$$

(2.29)

$$
R(t+1) = \frac{1}{conf} \times \begin{bmatrix} \beta & 0 \\ 0 & \beta \end{bmatrix}
$$

(2.30)

If the state observation confidence measure, *conf*, based on features likelihood map is low due to weak features responses then Kalman prediction result will be fused with feature-based tracker result to improve localization accuracy using Kalman filter

fusion:

$$S = HP(t+1|k)H' + R(t+1)$$
$$W = P(t+1|k)H'S^{-1}$$
$$P(t+1|t+1) = P(t+1|t) - WSW' + Q_{const} \qquad (2.31)$$
$$X(t+1|t+1) = X(t+1|t) + W(Z(t+1) - HX(t+1|t))$$

## 2.7   Target Localization

Possible target locations within the search window are denoted by peak locations in the fused posterior features likelihood map. Fused Features likelihood map is computed using weighted averaging. Given a case where feature fused likelihood maps indicates low probability of the target location (due to inadequacy of features to localize the object, occlusions, lighting conditions, shape deformation, etc.), a low confidence measure will be reported to call dynamically other cues for assistance. Then temporal moving object detection mask will be used to filter out false feature-based detections of background. The low confidence measure computed on features fused likelihood map will be used to update and enlarge observation covariance matrix. Finally filtered feature-based estimation will be fused with Kalman predicted position to localize the object. At the end of every iteration $t$, object's appearance descriptors $f_k$ will be updated using blending approach, where $\alpha$ is a small value (Eq. 2.32).

$$f_k(t+1) = \alpha \times f_k(t) + (1-\alpha) \times f_k(t-1) \qquad (2.32)$$

## 2.8 Visual Tracking Experimental Results

The proposed tracking algorithm has been implemented in Matlab and being tested for all 60 sequences of VOTC2016 benchmark full motion videos [72] and two aerial video including ARGUS and ABQ WAMI images. VOTC2016 is considered as the largest and most challenging benchmark for single-object short term tracking due to the number of submitted and tested state-of-the-art trackers [19]. The VOTC2016 dataset contains 60 sequences with several visual attributes and challenges including illumination changes, scale variations, motion change, camera motion and occlusion. ARGUS WAMI dataset collected by DARPA Autonomous Realtime Ground-Ubiquitous Surveillance-Imaging System contains more than 4000 wide aerial images. ARGUS-IS system is capable of imaging an area of greater than 40 square kilometers with a Ground Space Distance (GSD) of 15 cm at video rates of greater than 12 Hz [73]. ABQ aerial urban imagery were collected by TransparentSky [74] using an aircraft with on-board IMU and GPS sensors flying 1.5 km above ground level of downtown Albuquerque, NM on September 3, 2013. Imaging was done at frame rate of 4Hz and 2.6 km orbit radius. This dataset contains 1071 raw ultra high resolution images ($6400 \times 4400$) with nominal ground resolution of 25cm which are orthorectified using *MU BA4S* registration approach. For evaluation, we carried out the experiments on the first 200, $2000 \times 2000$ cropped images from location $(4761, 5800)$ upper left corner in the $12K \times 12K$ orthorectified images for which the ground-truth are provided by Kitware.

This section evaluates the accuracy and robustness performance of the proposed collaborative spatial context-aware tracking system (SPCT).

### 2.8.1 Evaluation Methodology

VOTC 2015 reset-based strategy is applied to evaluate the performance of the proposed Spatial Context-aware Tracking system (SPCT) and other trackers. The reset-based methodology is a supervised approach that detects a failure whenever tracker estimated bounding box result has zero overlap with the ground truth. If so tracker will be re-initialized five frames after the failure occurred. Two measures are used to analyze the proposed tracking system performance: Accuracy and Robustness. Accuracy is the average overlap between the predicted and ground truth bounding boxes during successful tracking periods. Robustness measures the number of times that tracker loses the target during tracking [42, 19, 43].

### 2.8.2 Color Model Performance Evaluation

An image is represented by a 2D array of pixels which are made of combinations of primary color channels. Color feature information presents rich discriminative power and are widely used in many detection and tracking systems. RGB color model is the most widely used color space that consists of three independent image plane including Red, Green and Blue. The color components of an 8-bit RGB image are integers in the range [0, 255]. Affine linear transformations of RGB color model create different color spaces including HSI, YCbCr, CYK etc. Conversion between the RGB model and the HSI model is as follows:

$$I = \frac{R + G + B}{3}, \quad S = 1 - \frac{min(R, G, B)}{I}, \quad H \in [0°, 360°] \tag{2.33}$$

Intensity (I) is just the average of the red, green and blue components range in $[0, 255]$. Saturation (S) is indicating the amount of white present and range in $[0, 1]$. Hue(H) describes the color itself in the form of an angle between $[0°, 360°]$, where $0°$ means red, $120°$ means green and $240°$ means blue.

YCbCr is scaled and offset version of YUV color space where $Y$ is defined to have a nominal 8-bit range of $[16, 235]$, Cb and Cr are defined between $[16, 240]$.

We have evaluated the performance of three color model including RGB, HSI and YCbCr in our proposed tracking system SPCT. We generated a 3D histogram tensor for each of the color models that are quantized in a fixed range. Our RGB and YCbCr color descriptors consist of two $32 \times 32 \times 32$ histogram tensor to represent object foreground and background information and HSI information are encoded in $18 \times 32 \times 32$ histogram tensor. Table 2.1 and 2.2 report SPCT performance results on VOTC2016 dataset for each of the color models with respect to robustness and accuracy respectively. As it is reported RGB color model achieves the best performance with the lowest robustness of 1.3 and accuracy of 0.458 followed by HSI with robustness of 1.43 and accuracy of 0.46.

Table 2.1: SPCT robustness performance evaluation using different color spaces on VOTC2016 dataset.

| VOTC2016 | | | | Robustness | | | |
|----------|-----|-----|------|----------|-----|-----|------|
| Sequence | **RGB** | **HSI** | **YCbCr** | Sequence | **RGB** | **HSI** | **YCbCr** |
| bag | 0 | 0 | 0 | handball1 | 3 | 3 | 3 |
| ball1 | 0 | 0 | 0 | handball2 | 4 | 3 | 3 |
| ball2 | 1 | 2 | 1 | helicopter | 1 | 1 | 1 |
| basketball | 1 | 1 | 1 | iceskater1 | 0 | 1 | 1 |
| birds1 | 1 | 6 | 4 | iceskater2 | 0 | 0 | 0 |
| birds2 | 0 | 1 | 1 | leaves | 2 | 2 | 2 |
| blanket | 1 | 1 | 1 | marching | 0 | 0 | 1 |
| bmx | 1 | 0 | 0 | matrix | 1 | 2 | 2 |
| bolt1 | 0 | 0 | 0 | motocross1 | 1 | 1 | 3 |
| bolt2 | 1 | 1 | 1 | motocross2 | 1 | 1 | 1 |
| book | 2 | 2 | 2 | nature | 2 | 2 | 3 |
| butterfly | 2 | 2 | 1 | octopus | 1 | 0 | 0 |
| car1 | 1 | 1 | 1 | pedestrian1 | 1 | 1 | 1 |
| car2 | 0 | 0 | 1 | pedestrian2 | 0 | 0 | 0 |
| crossing | 0 | 0 | 0 | rabbit | 4 | 4 | 6 |
| dinosaur | 3 | 4 | 3 | racing | 0 | 0 | 0 |
| fernando | 2 | 3 | 2 | road | 0 | 0 | 0 |
| fish1 | 4 | 3 | 5 | shaking | 2 | 2 | 5 |
| fish2 | 4 | 4 | 5 | sheep | 0 | 0 | 1 |
| fish3 | 0 | 0 | 1 | singer1 | 1 | 1 | 1 |
| fish4 | 1 | 1 | 1 | singer2 | 0 | 0 | 0 |
| girl | 1 | 1 | 1 | singer3 | 0 | 0 | 0 |
| glove | 2 | 2 | 2 | soccer1 | 3 | 2 | 5 |
| godfather | 0 | 0 | 0 | soccer2 | 3 | 4 | 4 |
| graduate | 3 | 4 | 6 | soldier | 0 | 0 | 1 |
| gymnastics1 | 5 | 5 | 5 | sphere | 0 | 0 | 0 |
| gymnastics2 | 5 | 5 | 5 | tiger | 0 | 0 | 0 |
| gymnastics3 | 2 | 2 | 2 | traffic | 1 | 1 | 1 |
| gymnastics4 | 0 | 0 | 0 | tunnel | 0 | 0 | 0 |
| hand | 4 | 4 | 5 | wiper | 0 | 0 | 0 |
| | | | | **Average** | **1.300** | 1.433 | 1.700 |

Table 2.2: SPCT accuracy performance evaluation using different color models on VOTC2016 dataset.

| VOTC2016 | Accuracy | | | | | | |
|---|---|---|---|---|---|---|---|
| Sequence | **RGB** | **HSI** | **YCbCr** | Sequence | **RGB** | **HSI** | **YCbCr** |
| bag | 0.475 | 0.476 | 0.474 | handball1 | 0.436 | 0.470 | 0.439 |
| ball1 | 0.761 | 0.743 | 0.739 | handball2 | 0.430 | 0.429 | 0.420 |
| ball2 | 0.520 | 0.310 | 0.564 | helicopter | 0.378 | 0.377 | 0.377 |
| basketball | 0.657 | 0.655 | 0.651 | iceskater1 | 0.508 | 0.512 | 0.495 |
| birds1 | 0.440 | 0.466 | 0.442 | iceskater2 | 0.507 | 0.516 | 0.505 |
| birds2 | 0.451 | 0.435 | 0.436 | leaves | 0.257 | 0.266 | 0.268 |
| blanket | 0.425 | 0.416 | 0.295 | marching | 0.608 | 0.614 | 0.623 |
| bmx | 0.239 | 0.214 | 0.225 | matrix | 0.446 | 0.545 | 0.417 |
| bolt1 | 0.388 | 0.385 | 0.386 | motocross1 | 0.368 | 0.375 | 0.377 |
| bolt2 | 0.492 | 0.500 | 0.484 | motocross2 | 0.415 | 0.530 | 0.444 |
| book | 0.326 | 0.389 | 0.386 | nature | 0.280 | 0.276 | 0.339 |
| butterfly | 0.395 | 0.409 | 0.375 | octopus | 0.306 | 0.289 | 0.290 |
| car1 | 0.656 | 0.654 | 0.653 | pedestrian1 | 0.562 | 0.555 | 0.560 |
| car2 | 0.709 | 0.719 | 0.683 | pedestrian2 | 0.177 | 0.171 | 0.172 |
| crossing | 0.455 | 0.439 | 0.443 | rabbit | 0.333 | 0.384 | 0.314 |
| dinosaur | 0.378 | 0.361 | 0.377 | racing | 0.347 | 0.347 | 0.346 |
| fernando | 0.395 | 0.424 | 0.406 | road | 0.525 | 0.522 | 0.522 |
| fish1 | 0.374 | 0.354 | 0.317 | shaking | 0.577 | 0.565 | 0.518 |
| fish2 | 0.343 | 0.330 | 0.371 | sheep | 0.333 | 0.342 | 0.434 |
| fish3 | 0.460 | 0.477 | 0.421 | singer1 | 0.538 | 0.497 | 0.539 |
| fish4 | 0.408 | 0.409 | 0.409 | singer2 | 0.672 | 0.679 | 0.672 |
| girl | 0.527 | 0.527 | 0.528 | singer3 | 0.207 | 0.193 | 0.206 |
| glove | 0.552 | 0.549 | 0.547 | soccer1 | 0.484 | 0.482 | 0.506 |
| godfather | 0.424 | 0.400 | 0.414 | soccer2 | 0.576 | 0.631 | 0.557 |
| graduate | 0.326 | 0.312 | 0.335 | soldier | 0.481 | 0.483 | 0.514 |
| gymnastics1 | 0.372 | 0.383 | 0.385 | sphere | 0.767 | 0.772 | 0.767 |
| gymnastics2 | 0.485 | 0.487 | 0.491 | tiger | 0.721 | 0.726 | 0.726 |
| gymnastics3 | 0.283 | 0.290 | 0.282 | traffic | 0.643 | 0.660 | 0.660 |
| gymnastics4 | 0.494 | 0.495 | 0.497 | tunnel | 0.320 | 0.339 | 0.339 |
| hand | 0.414 | 0.448 | 0.424 | wiper | 0.630 | 0.614 | 0.614 |
| | | | | **Average** | 0.458 | 0.460 | 0.457 |

## 2.8.3 Motion Prediction Performance Evaluation

One of the main tracking cues that we have considered in our collaborative tracking system particularly for path prediction is Kalman filter. SPCT utilizes Kalman prediction information to automatically detect the candidate region in the coming frames and to assist feature-based tracker when it fails to localize the object due to background clutter, occlusion or appearance changes within an informed fusion framework. This section presents performance evaluation of SPCT with and without using Kalman prediction on ARGUS WAMI dataset.

Table 2.3: Kalman prediction performance evaluation on ARGUS WAMI dataset.

| ARGUS WAMI Dataset | | | | | | | | | |
|---|---|---|---|---|---|---|---|---|---|
| | | without KF | | | using KF | | | | |
| Seq. | Track Len | A | R | MFR | A | R | MFR | KF Fused | **KF%** |
| 0 | 40 | 0.730 | 0 | 0.000 | 0.748 | 0 | 0.000 | 0 | 0% |
| 2 | 997 | 0.285 | 0 | 0.000 | 0.288 | 1 | 0.001 | 47 | 5% |
| 3 | 329 | 0.472 | 0 | 0.000 | 0.470 | 0 | 0.000 | 21 | 6% |
| 4 | 806 | 0.535 | 7 | 0.009 | 0.673 | 5 | 0.006 | 17 | 2% |
| 5 | 97 | 0.726 | 2 | 0.021 | 0.660 | 0 | 0.000 | 14 | 14% |
| 6 | 131 | 0.650 | 3 | 0.023 | 0.641 | 2 | 0.015 | 14 | 11% |
| 7 | 128 | 0.654 | 3 | 0.023 | 0.607 | 1 | 0.008 | 60 | 47% |
| 8 | 172 | 0.613 | 6 | 0.035 | 0.553 | 3 | 0.017 | 39 | 23% |
| 10 | 107 | 0.629 | 5 | 0.047 | 0.594 | 1 | 0.009 | 18 | 17% |
| 11 | 147 | 0.561 | 1 | 0.007 | 0.528 | 0 | 0.000 | 69 | 47% |
| 12 | 114 | 0.588 | 2 | 0.018 | 0.558 | 0 | 0.000 | 17 | 15% |
| 13 | 97 | 0.595 | 1 | 0.010 | 0.529 | 0 | 0.000 | 23 | 24% |
| 14 | 112 | 0.582 | 0 | 0.000 | 0.584 | 0 | 0.000 | 25 | 22% |
| 15 | 273 | 0.617 | 0 | 0.000 | 0.615 | 0 | 0.000 | 0 | 0% |
| 16 | 143 | 0.630 | 2 | 0.014 | 0.581 | 0 | 0.000 | 41 | 29% |
| 17 | 141 | 0.695 | 4 | 0.028 | 0.559 | 0 | 0.000 | 22 | 16% |
| 18 | 124 | 0.558 | 0 | 0.000 | 0.533 | 0 | 0.000 | 22 | 18% |
| 19 | 249 | 0.502 | 0 | 0.000 | 0.532 | 0 | 0.000 | 63 | 25% |
| 20 | 141 | 0.390 | 3 | 0.021 | 0.425 | 2 | 0.014 | 38 | 27% |
| avg | $\sum$= 4348 | 0.580 | 2.053 | 0.013 | 0.562 | **0.789** | **0.004** | $\sum$=550 | 18% |

Table 2.3 reports SPCT performance with and without incorporating Kalman prediction. Kalman Filter (KF) is called 550 times along 4348 ARGUS images which means that 18% of the time KF is being fused with feature-based results. Using KF improves the overall performance of SPCT and enables us to reduce the robustness from 2% to 0.4%.

### 2.8.4 SPCT Performance Evaluation on FMV

We compared SPCT performance results on VOTC2016 full motion video with 61 state-of-the-art trackers presented in the latest VOTC2016 challenge including purely color-based trackers **DAT** [75], **ASMS** [76], trackers that are variation of correlation filters including **SWCF** [77], **DSST2014** [34], **KCF2014** [33], **FCF** [19] and some that combined correlation filter outputs with color like **Staple** [22], **NSAMF** [35] and **ACT** [78]. We also compared our results with fragment-based trackers including frag-track **FRT** [36], optical flow clustering tracker **FCT** [37], **FoT** [38], **DPCF** [79] and **GGTv2** [80]. Tracker collection also contains **LoFT-Lite** [27] that is based on fusion of basic features optimized for aerial video, the multiple instance based tracker **MIL** [40], colour-aware complex cell tracker **CCCT** [81], structured SVM tracker **STRUCK'11** [39] and normalized cross-correlation tracker **NCC** [19]. Seven of the top ranked trackers are using deep learning techniques including **MLDF** [82], **TCNN** [83], **SSAT** [84], **DNT** [19], **DeepSRDCF** [85], **MDNet_N** [84] and **DDC** [19]. Raw results of the participating trackers are provided by VOTC2016 organizer.

As it is illustrated in Figure 2.14 SPCT ranked 11 among 62 trackers based on achieved robustness. Table 2.4 reports the computed measures for 62 considered trackers. The reported numbers are the average Accuracy (A) and Robustness(R) perfor-mance on all VOTC2016 60 sequences. Trackers have been ranked based on their robustness performance since SPCT main objective is performing persistent tracking and accurate object blob detection and segmentation is the next goal. As it is shown *SPCT* achieves robustness of 1.3 and accuracy of 0.458 and ranked 11.

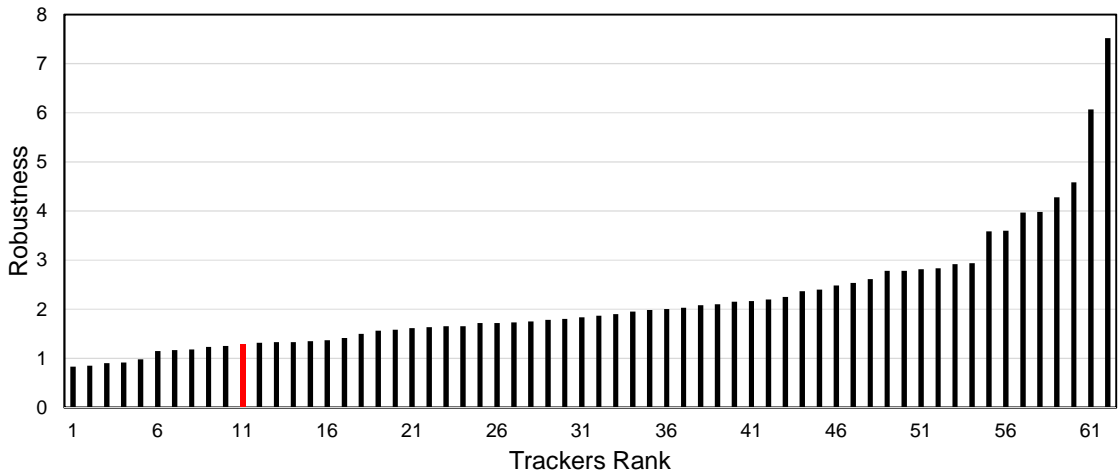

Figure 2.14: Robustness performance evaluation of VOTC2016 tested trackers. SPCT ranked 11 among 62 considered trackers.

### 2.8.5   SPCT Performance Evaluation on WAMI

In order to evaluate SPCT performance for Wide Aerial Motion Imagery, we have computed the same measures (accuracy and robustness) for SPCT and 9 other available trackers. Moreover, we have measured Missing Frame Rate (MFR) for all the 10 trackers that reports the number of failure over total number of frames. Table 2.5 reports the number of tracker failure per track ID on Argus WAMI dataset that contains ground truth bounding box annotations for 20 vehicles [73]. The last rows report the average computed Robustness(R), Accuracy(A) and MFR. SPCT ranked 1 among all trackers with low average robustness of 0.789, accuracy of 0.562 and missing frame rate as low as 0.4%.

Table 2.4: The table reports SPCT average accuracy and robustness performance on all VOTC2016 60 sequences compared with tested trackers discussed in VOTC2016.

| Rank | Tracker | A | R | Rank | Tracker | A | R |
|------|---------|---|---|------|---------|---|---|
| **1** | **MLDF** | 0.486 | 0.833 | 32 | ASMS | 0.485 | 1.867 |
| 2 | CCOT | 0.519 | 0.850 | 33 | HMMTxD | 0.093 | 1.900 |
| 3 | EBT | 0.437 | 0.900 | 34 | SCT4 | 0.453 | 1.950 |
| **4** | **TCNN** | 0.543 | 0.917 | 35 | TricTRACK | 0.443 | 1.983 |
| **5** | **SSAT** | 0.098 | 0.983 | 36 | LGT | 0.063 | 2.000 |
| **6** | **DNT** | 0.508 | 1.150 | 37 | KCF2014 | 0.483 | 2.033 |
| **7** | **DeepSRDCF** | 0.514 | 1.167 | 38 | CDTT | 0.415 | 2.083 |
| **8** | **MDNet_N** | 0.532 | 1.183 | 39 | SAMF2014 | 0.486 | 2.100 |
| **9** | **DDC** | 0.525 | 1.233 | 40 | OEST | 0.501 | 2.150 |
| 10 | SRBT | 0.091 | 1.250 | 41 | MWCF | 0.009 | 2.167 |
| **11** | **SPCT** | **0.458** | **1.300** | 42 | DPTG | 0.485 | 2.200 |
| 12 | STAPLEp | 0.547 | 1.317 | 43 | TGPR | 0.453 | 2.250 |
| 13 | RFD_CF2 | 0.473 | 1.333 | 44 | SWCF | 0.489 | 2.367 |
| 14 | SSKCF | 0.095 | 1.333 | 45 | ACT | 0.439 | 2.400 |
| 15 | Staple | 0.538 | 1.350 | 46 | MatFlow | 0.088 | 2.483 |
| 16 | SiamRN | 0.548 | 1.367 | 47 | MIL | 0.413 | 2.533 |
| 17 | SHCT | 0.015 | 1.417 | 48 | ART_DSST | 0.495 | 2.617 |
| 18 | SRDCF | 0.524 | 1.500 | 49 | DFST | 0.461 | 2.783 |
| 19 | NSAMF | 0.487 | 1.567 | 50 | SMPR | 0.436 | 2.783 |
| 20 | ColorKCF | 0.089 | 1.583 | 51 | FCT | 0.019 | 2.817 |
| 21 | BST | 0.084 | 1.617 | 52 | BDF | 0.083 | 2.833 |
| 22 | FCF | 0.070 | 1.633 | 53 | sKCF | 0.015 | 2.917 |
| 23 | CCCT | 0.444 | 1.650 | 54 | FoT | 0.068 | 2.933 |
| 24 | SiamAN | 0.501 | 1.650 | 55 | DFT | 0.439 | 3.583 |
| 25 | DAT | 0.089 | 1.717 | 56 | STC | 0.368 | 3.600 |
| 26 | GGTv2 | 0.013 | 1.717 | 57 | IVT | 0.408 | 3.967 |
| 27 | GCF | 0.510 | 1.733 | 58 | Matrioska | 0.087 | 3.983 |
| 28 | DPT | 0.479 | 1.750 | 59 | LT_FLO | 0.019 | 4.283 |
| 29 | KCF_SMXPC | 0.521 | 1.783 | 60 | LoFT_Lite | 0.332 | 4.583 |
| 30 | MAD | 0.480 | 1.800 | 61 | CMT | 0.371 | 6.067 |
| 31 | ANT | 0.096 | 1.833 | 62 | ncc | 0.451 | 7.517 |

Table 2.5: Table reports the computed average robustness, MFR and accuracy on ARGUS WAMI 20 sequences.

| Robustness (# tracker failure) | | | | | | | | | | | |
|---|---|---|---|---|---|---|---|---|---|---|---|
| Tr# | #Frs | SPCT | SWCF | FCT | NSAMF | SAMF14 | *LoFT** | KCF | NCC | DPCF | STAPLE |
| 0 | 40 | 0 | 0 | 0 | 0 | 0 | 0 | 0 | 6 | 4 | 5 |
| 2 | 997 | 1 | 1 | 0 | 2 | 1 | 1 | 1 | 3 | 3 | 3 |
| 3 | 329 | 0 | 0 | 0 | 0 | 1 | 2 | 2 | 3 | 0 | 1 |
| 4 | 806 | 5 | 8 | 7 | 8 | 8 | 8 | 8 | 10 | 9 | 8 |
| 5 | 97 | 0 | 3 | 4 | 4 | 3 | 3 | 3 | 4 | 4 | 9 |
| 6 | 131 | 2 | 5 | 6 | 5 | 5 | 6 | 5 | 10 | 8 | 11 |
| 7 | 128 | 1 | 4 | 4 | 4 | 5 | 6 | 5 | 5 | 7 | 11 |
| 8 | 172 | 3 | 9 | 10 | 9 | 8 | 8 | 9 | 8 | 10 | 9 |
| 10 | 107 | 1 | 5 | 6 | 6 | 5 | 7 | 6 | 7 | 6 | 6 |
| 11 | 147 | 0 | 4 | 3 | 3 | 4 | 4 | 4 | 5 | 8 | 4 |
| 12 | 114 | 0 | 2 | 5 | 5 | 5 | 3 | 3 | 14 | 11 | 12 |
| 13 | 97 | 0 | 4 | 2 | 2 | 4 | 1 | 5 | 7 | 9 | 7 |
| 14 | 112 | 0 | 3 | 3 | 5 | 4 | 3 | 5 | 14 | 11 | 13 |
| 15 | 273 | 0 | 0 | 0 | 0 | 0 | 1 | 0 | 0 | 0 | 0 |
| 16 | 143 | 0 | 3 | 3 | 3 | 4 | 3 | 4 | 5 | 8 | 10 |
| 17 | 141 | 0 | 3 | 4 | 4 | 3 | 5 | 5 | 5 | 5 | 4 |
| 18 | 124 | 0 | 1 | 0 | 1 | 1 | 1 | 2 | 2 | 7 | 9 |
| 19 | 249 | 0 | 2 | 2 | 2 | 3 | 3 | 3 | 4 | 4 | 5 |
| 20 | 141 | 2 | 3 | 2 | 2 | 2 | 3 | 2 | 3 | 3 | 2 |
| Total | 4348 | 15 | 60 | 61 | 65 | 66 | 68 | 71 | 115 | 117 | 129 |
| R | avg | 0.789 | 3.158 | 3.211 | 3.421 | 3.474 | 3.579 | 3.737 | 6.053 | 6.158 | 6.789 |
| MFR | avg | 0.004 | 0.021 | 0.022 | 0.023 | 0.024 | 0.023 | 0.025 | 0.048 | 0.048 | 0.055 |
| A | avg | 0.562 | 0.637 | 0.512 | 0.608 | 0.632 | 0.394 | 0.617 | 0.595 | 0.580 | 0.428 |

Table 2.6: The table reports SPCT average accuracy, MFR and robustness compared to several state-of-the art trackers on ABQ WAMI dataset.

| Tracker | **SPCT** | SWCF | NSAMF | SAMF14 | KCF2014 | **NCC** | *LoFT** | Staple | DPCF | FCT |
|---|---|---|---|---|---|---|---|---|---|---|
| R | **0.325** | 0.634 | 0.789 | 0.870 | 0.911 | 1.000 | 1.179 | 1.398 | 1.585 | 1.642 |
| MFR | **0.005** | 0.011 | 0.014 | 0.016 | 0.016 | 0.017 | 0.017 | 0.025 | 0.024 | 0.028 |
| A | 0.628 | 0.674 | 0.631 | 0.662 | 0.638 | **0.680** | 0.367 | 0.625 | 0.603 | 0.590 |

Table 2.6 reports the computed measures for ABQ aerial urban imagery which were collected over downtown Albuquerque, NM [74]. SPCT ranked 1 among the considered trackers with average robustness of 0.325, accuracy of 0.628 and low average MFR of 0.005 on all 140 sequences.

# Chapter 3

# Multi-scale Spatially Weighted Local Histograms in O(1)

In many image processing applications, histograms are commonly used to characterize and analyze the region of interest within the image. Histogram-based features are space efficient, simple to compute, robust to translation and particularly invariant to orientation for color-based features. However, when computing a plain histogram, spatial information are missed which makes it sensitive to noise and occlusion. Several techniques are proposed to preserve spatial information including color Correlograms [86], Spatiogram [87], Multiresolution histogram [88], locality sensitive histogram [21] and fragment-based approaches that exploit the spatial relationships between patches [36]. Spatially weighted histograms boost the performance of many image processing tasks at the expense of speed. In [64], Porikli generalized the concept of integral image and presented computationally very fast method to extract the plain histogram of any arbitrary region in constant time. Integral histogram provides an optimum and complete solution for the histogram-based search problem. Since then many novel approaches have been presented based on integral histogram to accelerate the performance of image processing tasks and incorporate the spatial information including filtering [1, 89, 2], classification and recognition [88, 90], and detection and tracking [91, 92].

Despite all different techniques that have been proposed to adaptively weight the contribution of pixels when computing local histograms by considering their distance from center pixel, the problem of how accurately extract the spatially weighted histogram of any arbitrary region within an image in constant time using integral histogram is still unsolved. Frag-track [36] proposes a discrete approximation scheme instead of the continuous kernel weighting approach to give higher weight to the contribution of inner rectangle compare to region margins for fast search.

I present a novel fast algorithm to accurately evaluate spatially weighted local histograms in O(1) time complexity using an extension of the integral histogram method (SWIH). The main idea is to (1) decompose the spatial filter into independent weights $w_i$ and split kernel into multiple quadrants $q_i$ subsequently, (2) for all $w_i$ compute candidate region weighted integral histogram $IH_{w_i}$, (3) for every quadrant $q_i$ compute its weighted local histogram using the corresponding $IH_{w_i}$ and considering its translation from center pixel, (4) normalize the local histograms, (5) finally add local histograms together to build the full kernel spatially weighted local histogram (Figure 3.5).

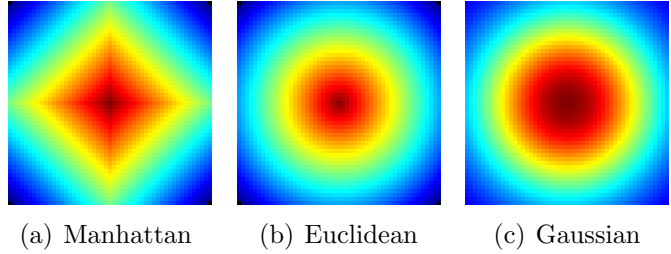

| (a) Manhattan | (b) Euclidean | (c) Gaussian |

Figure 3.1: Illustration of linear and non-linear distance kernels.

# 3.1  Spatially Weighted Local Histograms

Weighting pixel contributions is a key feature in many fundamental image processing tasks including filtering, modeling and matching to increase the accuracy of results in detection, tracking, recognition, etc..

The main idea is to assign lower weights to the pixels that most likely belong to background or occluding objects. One common technique is to define a weighting function $w(x, y)$ that assigns weights to pixels with respect to their distance from target center (since undesirable pixels are usually considered around the region contours) including Manhattan, Euclidean, Gaussian or exponential weighting distance functions (Figure 3.1). Then the spatially weighted histogram is computed as follow:

$$H(T, b_i) = \sum_{x,y \in T}^{w \times h} \delta(Q(f(x, y)) - b_i) \times w(x, y) \qquad (3.1)$$

where $T$ is the region of interest of size $w \times h$, $b_i$ is the histogram bin index, $\delta$ is the pulse function, $Q$ is the quantization function for feature values $f$.

Having such kernels enables us to adaptively weight the contribution of pixels and diminish the presence of background information when computing weighted local histograms. Figure 3.2 shows the accuracy of intensity feature likelihood maps

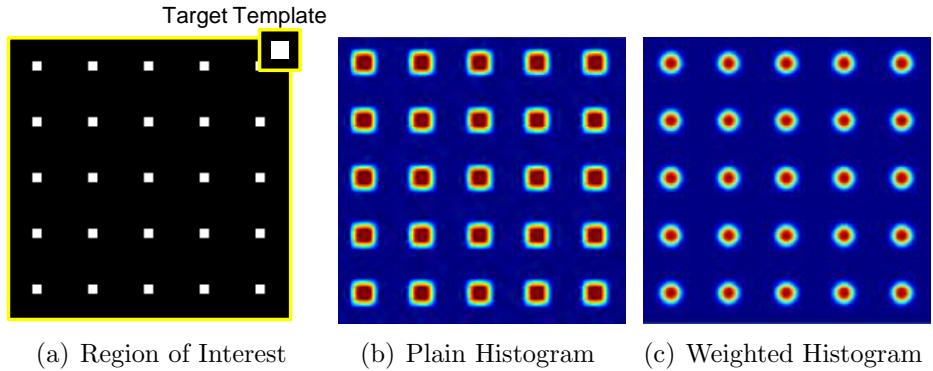

(a) Region of Interest     (b) Plain Histogram     (c) Weighted Histogram

Figure 3.2: Performance evaluation of intensity feature likelihood maps using sliding window (b) plain versus (c) spatially weighted histogram distance matching. Weighting pixel contribution considering its location is a key feature to increase the accuracy and boost the performance of detection, tracking and recognition systems

based on sliding window histogram matching when using plain local histograms versus spatially weighted local histograms. As it can be seen, using spatially weighted local histograms generates more robust matching results. In the following sections, we describe the straightforward convolution-based approach, the discrete approximation scheme and our proposed novel, fast and accurate algorithm based on weighted integral histogram to compute spatially weighted local histograms for fast search.

### 3.1.1 Brute-force Approach

The computational complexity of the brute-force approach to compute the adaptively weighted local histograms at each candidate pixel location is linear in the kernel size and the number of candidate pixels. Assuming a search window of size $w \times h$ and a neighborhood of size $k \times k$ and $b$-dimensional histogram, the computational complexity of finding the best matched pixel location is $O(b \times k^2 \times w \times h)$, which makes the system far away from real-time performance particularly when it comes to large scale high resolution image analysis.

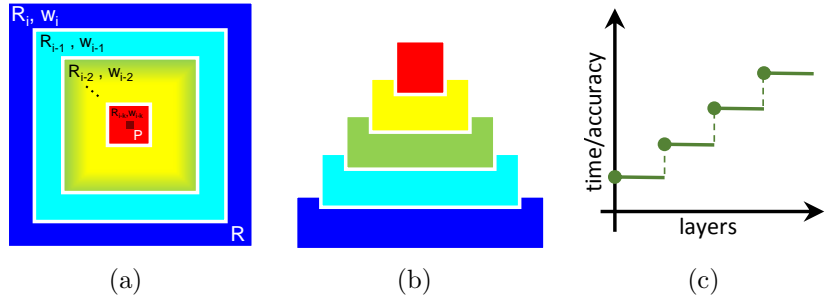

Figure 3.3: (a) Wedding-Cake Approach: the discrete approximation scheme to obtain the spatially weighted local histogram for the candidate region considering inner-nested windows and using integral histogram ($w_i < w_{i-1} < ... < w_{i-k}$). (b) Wedding-cake slice approximation. (c) Computational time complexity and accuracy increase by increasing the number of layers.

### 3.1.2 Wedding-Cake Approach

One solution to meet the demands of real-time implementation is to extract local histograms in constant time using integral histogram. However, as of our knowledge, there is still no solution to accurately and efficiently extract spatially weighted local histograms in O(1) using integral histogram but the discrete approximate scheme presented in [36]. Frag-track proposed a simple approach to approximate the kernel function with different weights instead of pixel-level kernel weighting. Assuming that we want to calculate a spatially weighted local histogram in the rectangular region $R$ centered at point $P$ using integral histogram. Such counting can be approximated by considering several inner-nested windows $R_i$ at multiple scales around $P$ (Figure 3.3(a)). The goal is to compute the counts of the rings between two adjacent windows $R_i$ and $R_{i-1}$ by subtracting their local histograms that are obtained in constant time using integral histogram. Then, rings histograms will be weighted appropriately with respect to their distance from $P$ and combined to provide an approximate spatially weighted local histogram on $R$.

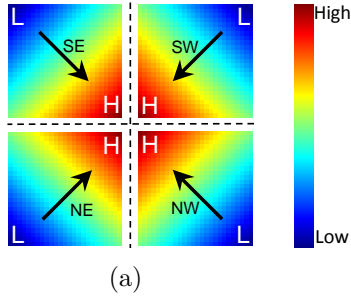

(a)

Figure 3.4: Tiling the kernel into four quadrants and decomposing the weights to accurately compute spatially weighted local histogram.

$$SWLH(R) = w_i \times (H(R_i) - H(R_{i-1})) + ... + w_{i-1} \times H(R_{i-k}) \qquad (3.2)$$

The accuracy of this approximation relies on the number of considered inner-nested windows. We presented a new approach to compute spatially weighted local histograms that is more accurate than the wedding-cake method and takes constant time using an extension of integral histograms.

### 3.1.3 Multi-scale Spatially Weighted Local Histograms in Constant Time Complexity (SWIH)

When using integral histogram, it is not clear how to weight pixel contributions when computing arbitrary rectangular region histogram in O(1). We propose to address the pixel-level weighting problem by tiling the kernel into multiple quadrants as well as decomposing the weights (Figure 3.4). In this section we describe our proposed algorithm in details when using Manhattan distance function (Figure 3.1(a)) to adaptively weight pixel contributions for fast matching. Assuming that we want to weight the contribution of each pixel within region $R$ centered at $P_c = (x_c, y_c)$ by its Manhattan distance from $P$ when computing histogram of region $R$. Manhattan or city-block weighting function measures the sum of the absolute distance between two points along each axis. In our case, Manhattan distance of any arbitrary point $P_i = (x_i, y_i)$ within region $R$ is

$$Dist_{Manhattan}(P_i, P_c) = \mid x_i - x_c \mid + \mid y_i - y_c \mid \tag{3.3}$$

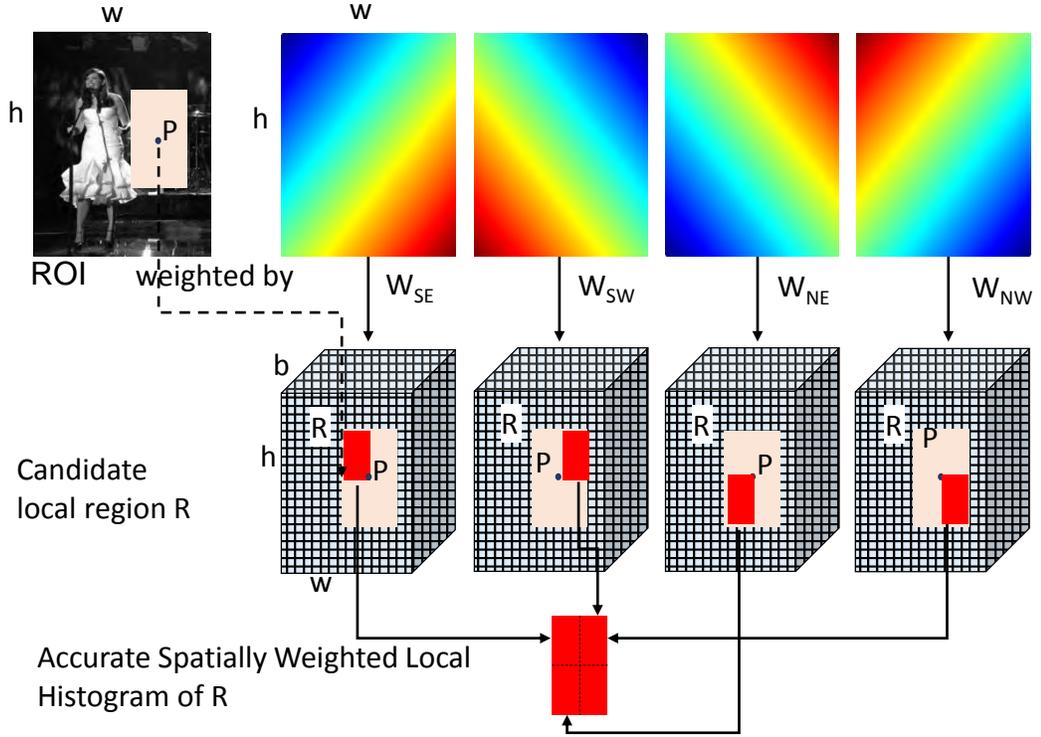

Figure 3.5: Computational flow of accurate spatially weighted local histograms using weighted integral histogram for the region of interest (ROI) of size $w \times h$.

Since the filter is rectilinear and symmetric, we propose to decompose it into four independent weighting functions in the shape of four quadrants:

TopLeft(TL), TopRight(TR), BottomLeft(BL) and BottomRight(BR) (Figure 3.3(b)).

As it is shown, weights linearly increase from one corner to its diagonally opposite corner in each of the quadrants covering four directions: {SE, SW, NE, NW}. We extend these weights for the region of interest and compute four differently weighted integral histogram. For each direction, we consider two correlated images $f$ and $w_{dir}$ to compute the weighted integral histogram up to point $(x, y)$:

$$IH_{w_{dir}}(x, y, b_i) = \sum_{i \leq x, j \leq y} \delta(Q(f(i, j)) - b_i) w_{dir}(i, j) \qquad (3.4)$$

$f$ contains image feature values, $Q$ is the quantization function that determines which bin to increase, $\delta$ is the impulse function and $w_{dir}$ is the pixel-wise weighing function that determine the value to increase at that bin. Having four differently weighted integral histogram, each of the quadrants spatially weighted local histogram will be computed in O(1) using its corresponding weighted integral histogram and considering its translation from the kernel center point. We will normalize the histograms and add them together to build the full region spatially weighted histogram. Figure 3.5 illustrates the flow of the computation.

It is noteworthy to mention that due to weights rectilinear changes, their values are independent of the pixel location in the region of interests. This characteristic enables us to appropriately normalize the computed weighted local histogram and match it with the target spatially weighted histogram regardless of its location. This new method provides multi-scale accurate spatially weighted local histogram in constant time and can be utilized for other spatial weighting functions. It can be easily adapted to any fast computation of integral histogram on GPUs to accelerate the computation [44].

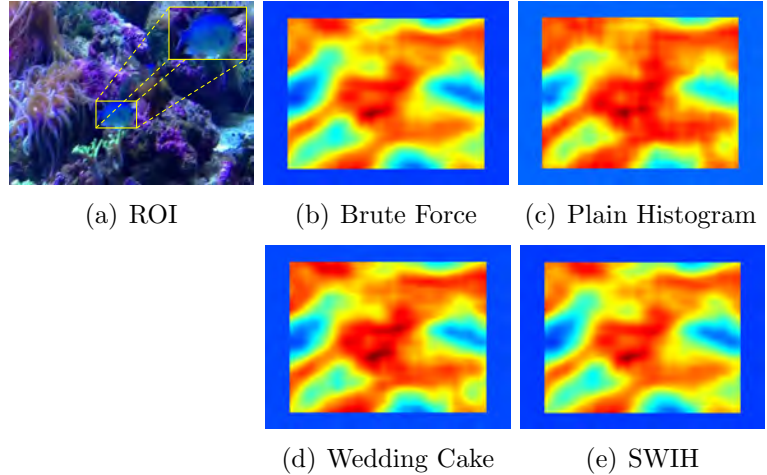

| (a) ROI | (b) Brute Force | (c) Plain Histogram |

| (d) Wedding Cake | (e) SWIH |

Figure 3.6: Performance evaluation of intensity likelihood maps estimation using sliding window histogram matching. Weighting pixel contribution considering its location results in more accurate and robust target localization as shown in (b) and (e).

## 3.2   Experimental Results and Performance Evaluation

In this section, we evaluate the performance of our approach and compare it with brute-force implementation and approximation scheme with respect to computational complexity and accuracy. Figure 3.6 illustrates the performance of the estimated intensity likelihood maps for a sample image from the VOT2016 data set [72] using sliding-window histogram matching. We compared the intensity likelihood map computed by the brute-force implementation with the matching results of the plain histogram, approximation scheme and our proposed accurate fast spatially weighted histogram. Background clutter is one of the main challenges in object detection systems relied on matching. We selected an image that contains background clutter to make the matching process very challenging. We calculated the mean-squared error (MSE) between the brute-force result which is our reference model and the two other

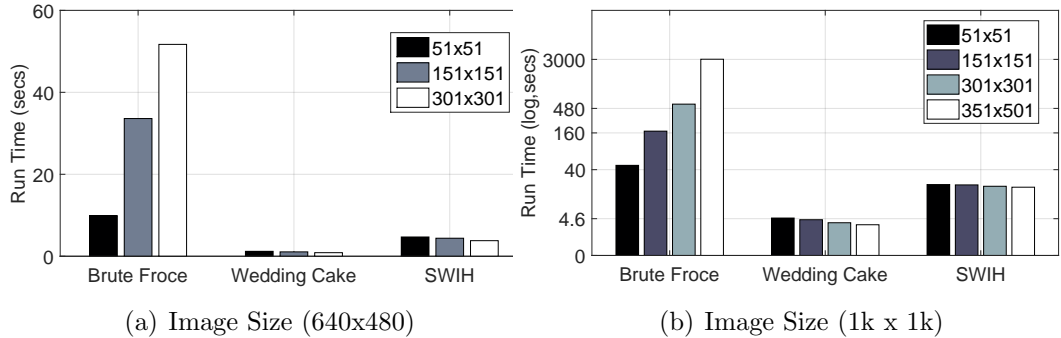

(a) Image Size (640x480)          (b) Image Size (1k x 1k)

Figure 3.7: Performance evaluation comparison by increasing the local histogram sliding window size. Computational time complexity of integral histogram-based methods are invariant of kernel size. However, Computational complexity and accuracy of Wedding Cake approach increases by increasing the number of layers

techniques. The MSE between brute-force and our results (SWIH) is 0 as we expected and 0.0012 using the approximation scheme with 3 layers respectively. It is proved that our proposed method not only provides exact results as the brute-force approach but is much faster and independent of sliding window size.

For a candidate region of size $345 \times 460$ and sliding window of size $61 \times 91$ (Figure 3.6(a)), SWIH is 4.5 times faster than brute-force implementation. Figure 3.7(a) and (b) shows the computational complexity of each of the discussed methods for standard image $640 \times 480$ as well as large scale image of size $1k \times 1k$ for different sliding window size from small scale to very large scale. It can be seen that the local histograms computational time using the brute-force implementation increases dramatically by enlarging the kernel size but is invariant of sliding window size for the approximation scheme and SWIH. The execution time of the integral histogram based methods are invariant of sliding window size, however there is a small drop in execution time when increasing the sliding window size. The reason is that the number of sliding windows and consequently the number of computed local histograms is reducing by increasing the sliding window size.

This chapter presents our novel fast algorithm to accurately evaluate spatially weighted local histograms in constant time using an extension of the integral histogram method (SWIH). We have shown that SWIH produces exact local histograms compared to brute-force approach and is much faster. Utilizing the integral histogram makes it to be fast, multi-scale and flexible to different weighting functions. This technique can be applied to fragment-based approaches to adaptively weight object patches considering their location. SWIH can be integrated into any detection or tracking system to provide an efficient exhaustive search and achieve more robust and accurate target localization.

# Chapter 4

# Automatic Moving Object Detection

Automatic moving object detection and segmentation is a critical low-level task for many video analysis and tracking applications. Fast and accurate foreground motion estimation provides primary useful information for a number of image and video analysis including urban traffic monitoring [93, 94], object classification [95, 96, 95], registration and tracking [23, 97]. Moreover, many of the WAMI trackers use motion based cueing either as the primary module or fused with other modalities to enhance the performance of detecting and tracking moving objects.

## 4.1   Moving Object Detection Approaches

I worked on utilizing different techniques to estimate the foreground motion including spatio-temporal median-based background modeling and the trace of the flux tensor.

### 4.1.1 Background Subtraction Using 3D Median Background Modeling

Median filtering performs as well as other more complicated techniques (i.e.GMM) and avoids creating unrealistic pixel values when blending pixel values. Spatio-temporal median-based background modeling is simple, computationally efficient and robust to noise. Median computation methods can be grouped as sorting-based or histogram-based approaches. We investigated the computational time and memory complexity of either sorting-based or histogram-based approaches based on image size, temporal window size and histogram bin size. For example, computing 64-bins integral histogram for a $2k \times 2k$, 8 bit images for a temporal window of size 17 requires 256 Megabytes of memory while following sorting-based approach takes only 44 Megabytes of memory and the median computation is still fast.

**3D Median Computation Using Integral Histogram for Full Motion Video**

In the early works, we proposed a fast multi-scale 3D median computation algorithm that can handle large temporal windows for standard full motion videos utilizing GPU implementation of integral histogram. The integral histogram of an image provides the local histogram of every arbitrary target region in constant time. Taking advantages of this property enables us to compute the medians in constant time for all target pixels using its local histogram. Algorithm 4 describes spatio-temporal median computations using the integral histogram (assuming sequences of $k$ images are transferred to the GPU using double buffering and a spatio-temporal median of size $m \times n \times (T + 1)$ is computed). In the initialization phase, the individual integral histograms are computed for each image array. Meanwhile the joint integral histogram is computed for the first $T + 1$ individual integral histograms (assume that $T$ is an even number). After creating the joint integral histogram, the median calculation phase started from frame $\frac{T}{2} + 2$. In each iteration, first, the joint integral histogram is updated by adding the integral histogram of head image and subtracting the integral histogram of tail image (Fig. 4.1). Then medians are computed for each pixel using the joint local histograms of kernel size $m \times n$ where local kernel histograms can be obtained in $O(1)$. Figure 4.2 demonstrates background subtraction results for PETS2013 benchmark background dataset including an indoor hallway motion, outdoor motion and a tracking scenario (seq. irw01) from OTCBVS Benchmark IR Database [31]. The performance in terms of speedup has been discussed in chapter 5.

---

**Algorithm 4:** 3D Median Computation Using Integral Histogram

**Input :** Image sequences **I[k]** of size $h \times w$,
number of bins $b$,
size of image history $T + 1$

**Output :** Medians **M[k − T]]** of size $h \times w$

1: **Initialize JointIH**
2: IH_tail=JIH=Integral_Hist(Quantized(image(1))); //Compute the first joint integral histogram for the first $T + 1$ frames
3: **for** $Fr = 2 : T + 1$ **do**
4: $IH[Fr] = Integral\_Hist(Quantized(image(Fr)));$
5: JIH = JIH + IH[Fr];
6: **end for**

7: //Calculate the Median of current frame
8: **for** $Fr = T + 2 : k$ **do**
9: //Update the IH_head
10: IH_head=Integral_Hist(Quantized(image(Fr)));

11: //Update the Joint Integral Histogram
12: JIH= JIH + IH_head - IH_tail;

13: //Update the IH_tail
14: IH_tail=Integral_Hist(Quantized(image(Fr-T)));

15: //Compute Median $[Fr - \frac{T}{2}]$
16: Median $[Fr - \frac{T}{2}]$ = ComputeLocalMedian(JIH)
17: **end for**

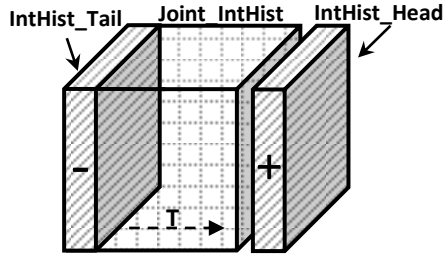

Figure 4.1: Updating the joint integral histogram for spatio-temporal median computation.

**Sorting-Based 3D Median Computation for Wide Aerial Motion Imagery**

Although using integral histogram boost the median computation performance, but GPU memory allocation for integral histogram tensor for high resolution, large scale aerial motion imagery, is not feasible and efficient when the histogram bin size is much larger than the median temporal window size. For example, computing 64-bins integral histogram for a $2k \times 2k$, 8 bit images requires 256 Megabytes while following sorting-based approach for a temporal window of size 17 takes only 44 Megabytes of memory and the median computation is still fast. Therefore, I used the sorting-based approach to compute the spatio-temporal median for high resolution WAMI images to avoid complexities of *Tiling* the image for GPU computation and hold the computation fast enough due to small temporal window size. Figure 4.3 illustrates moving object detection results for high resolution $2k \times 2k$ WAMI images collected over downtown Albuquerque, NM. We model the background using median of 17 images (eight frames before and eight frames after the target image), and perform background subtraction to estimate the moving foreground objects. Every pixel is classified as moving versus stationary by thresholding the estimated foreground image. Contrast enhancement and morphology operations are applied to improve the motion detection results and to filter out the spurious noises caused by illumination changes. However, it is seen that approximately 90.2% of the motion responses are induced by parallax effects of tall structures (area under blue mask shown in last row of Fig. 4.3) which significantly degrade the precision of the motion detection results. The low average precision of 12.9% is reported for the first 200 frames of Albuquerque sequence.

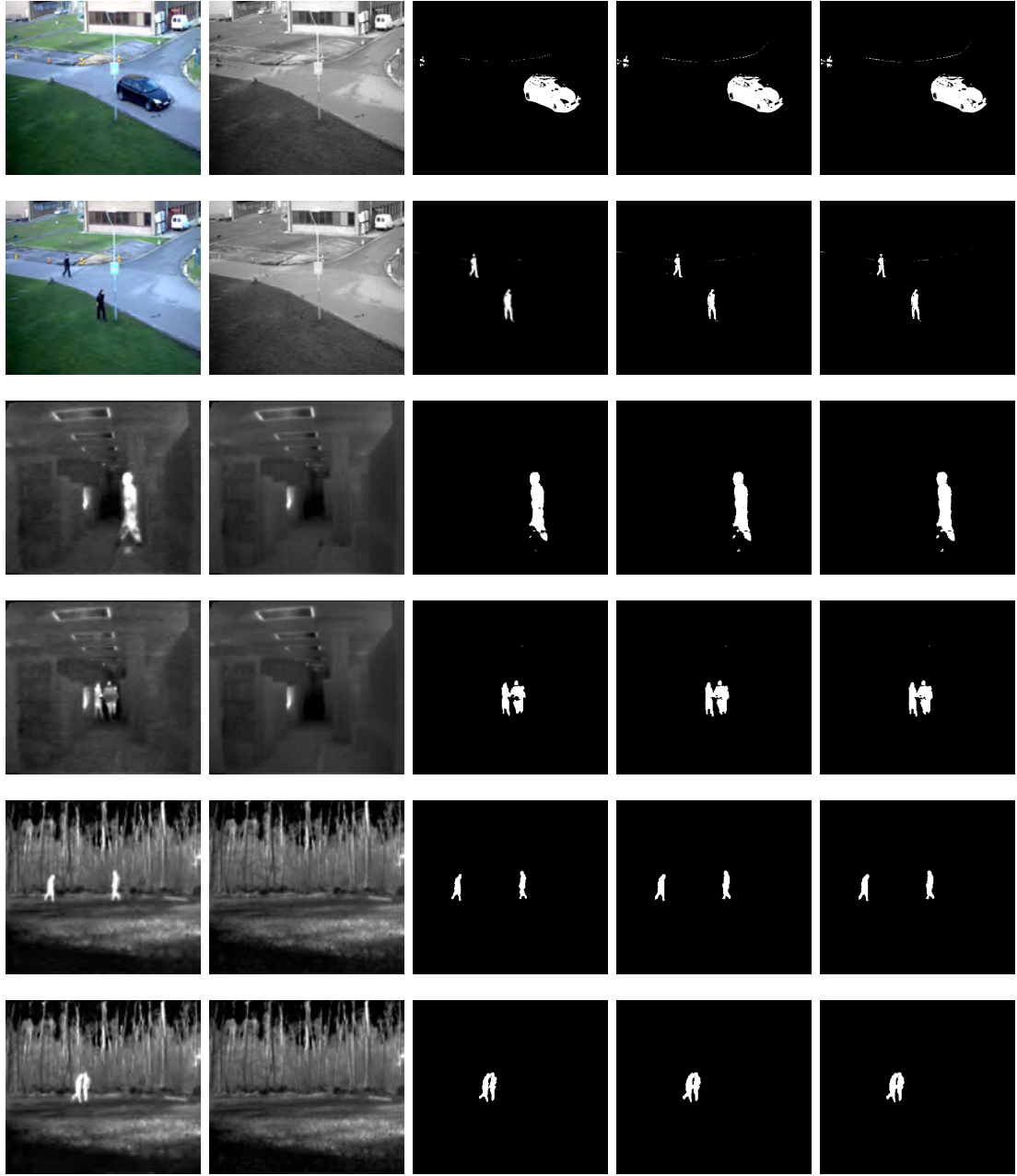

Figure 4.2: Background subtraction for PETS2013 Benchmark background dataset (first row frame 73, second row frame 161) [98], an indoor hallway motion (seq. irin01) from OTCBVS Benchmark IR Database (row third frame 5150 and row four frame 8329) [99], and outdoor motion and tracking scenarios (seq. irw01) from OTCBVS Benchmark IR Database (row five frame 332, row six frame 399)[99] using 3D median filter based on GPU integral histogram using different number of bins, from left column to the right column: original image, background model (256 bins), foreground using 16 bins integral histogram, foreground using 128 bins integral histogram, foreground using 256 bins integral histogram.

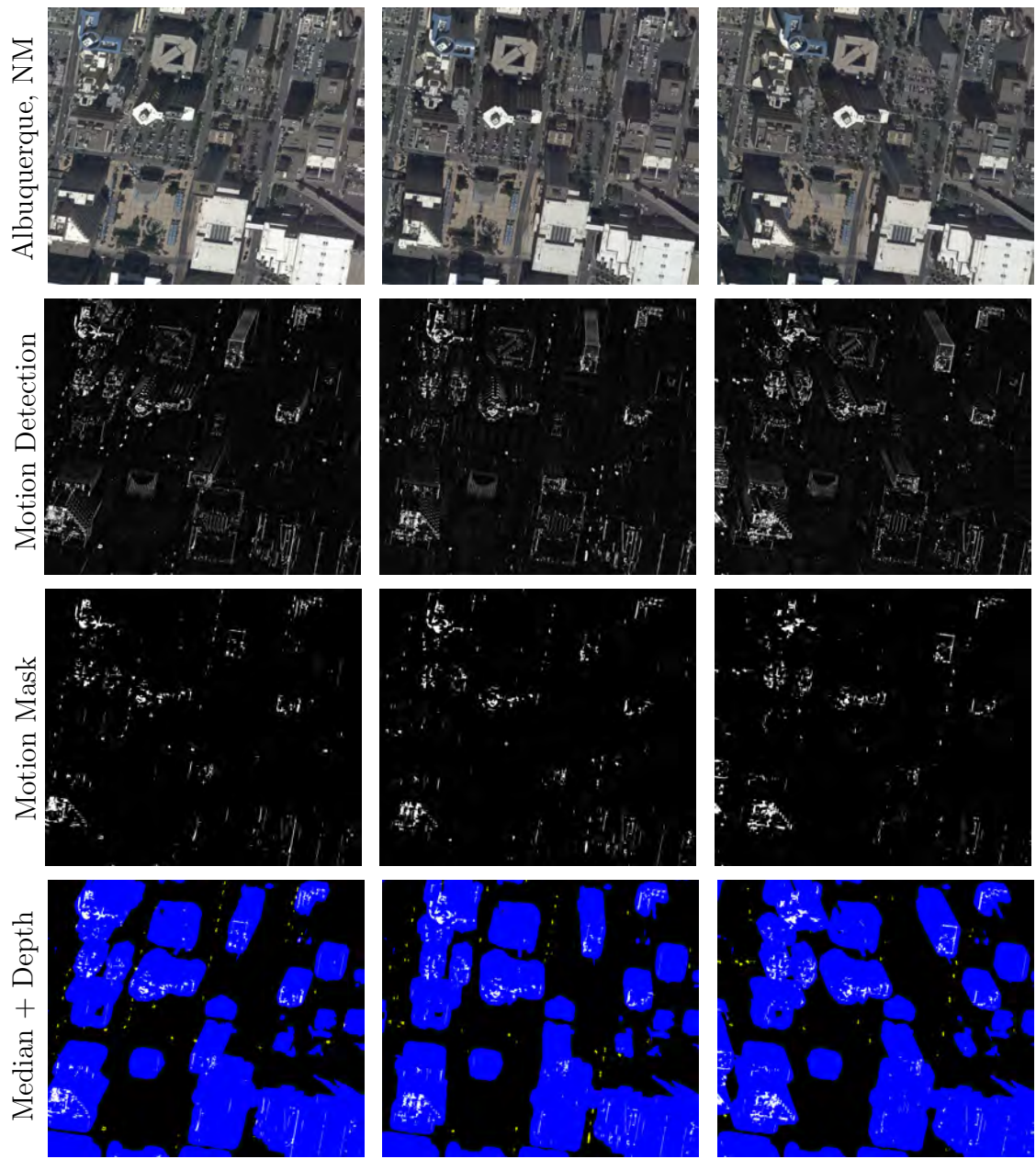

Figure 4.3: Illustration of motion detection results using spatio-temporal Median: First row shows the registered $2k \times 2k$ cropped frames (57, 123, 182) from Albuquerque aerial imagery. Second row shows moving object detection results for the corresponding frames. Third and fourth row present the motion detection masks obtained from 3D median background subtraction and the building masks in blue, respectively.

### 4.1.2 Moving Object Detection Using the Trace of the Flux Tensor

In a similar work, I exploited the trace of the flux tensor to detect moving vehicles in urban aerial imagery. Flux tensor is presented as an extension of 3D structure tensor that allows reliable motion segmentation without expensive eigenvalue decomposition [100, 101]. Under constant illumination model, optical-flow equation of a spatiotemporal image volume $\mathbf{I}(\mathbf{x})$ centered at location $\mathbf{x} = [x, y, t]$ is

$$
\begin{aligned}
\frac{d\mathbf{I}(\mathbf{x})}{dt} &= \frac{\partial \mathbf{I}(\mathbf{x})}{\partial x}\,v_x + \frac{\partial \mathbf{I}(\mathbf{x})}{\partial y}\,v_y + \frac{\partial \mathbf{I}(\mathbf{x})}{\partial t}\,v_t \\
&= \nabla \mathbf{I}^T(\mathbf{x})\,\mathbf{v}(\mathbf{x})
\end{aligned}
\tag{4.1}
$$

taking the derivative of Eq. 4.1 with respect to $t$, we obtain Eq. 4.2

$$
\begin{aligned}
\frac{\partial}{\partial t}\left(\frac{d\mathbf{I}(\mathbf{x})}{dt}\right) &= \frac{\partial^2 \mathbf{I}(\mathbf{x})}{\partial x \partial t}\,v_x + \frac{\partial^2 \mathbf{I}(\mathbf{x})}{\partial y \partial t}\,v_y + \frac{\partial^2 \mathbf{I}(\mathbf{x})}{\partial t^2}\,v_t \\
&\quad + \frac{\partial \mathbf{I}(\mathbf{x})}{\partial x}\,a_x + \frac{\partial \mathbf{I}(\mathbf{x})}{\partial y}\,a_y + \frac{\partial \mathbf{I}(\mathbf{x})}{\partial t}\,a_t
\end{aligned}
\tag{4.2}
$$

which can be written in vector notation as,

$$
\frac{\partial}{\partial t}(\nabla \mathbf{I}^T(x)\mathbf{v}(\mathbf{x})) = \frac{\partial \nabla \mathbf{I}^T(\mathbf{x})}{\partial t}\,\mathbf{v}(\mathbf{x}) + \nabla \mathbf{I}^T(\mathbf{x})\,\mathbf{a}(\mathbf{x})
\tag{4.3}
$$

where $\mathbf{v}(\mathbf{x}) = [v_x, v_y, v_t]$ is the optical-flow vector and $\mathbf{a}(\mathbf{x}) = [a_x, a_y, a_t]$ is the acceleration of the image brightness located at $\mathbf{x}$. Usually $\mathbf{v}(\mathbf{x})$ is estimated by minimizing Eq. 4.3 over a local 3D image patch $\mathbf{\Omega}(\mathbf{x}, \mathbf{y})$:

$$\frac{\partial \nabla \mathbf{I}^T(\mathbf{x})}{\partial t} \mathbf{v}(\mathbf{x}) + \nabla \mathbf{I}^T(\mathbf{x}) \, \mathbf{a}(\mathbf{x}) = 0 \tag{4.4}$$

Assuming a constant velocity model subject to the normalization constraint $||\mathbf{v}(\mathbf{x})|| = 1$ and consequently zero acceleration, a least-squares error measure $e_{ls}(\mathbf{x})$ (Eq. 4.5) is used to minimize the Eq. 4.4

$$\begin{aligned}
e_{ls}(\mathbf{x}) = \int_{\mathbf{\Omega}(\mathbf{x}, \mathbf{y})} \; &\left( \frac{\partial (\nabla \mathbf{I}^T(\mathbf{y})}{\partial t} \mathbf{v}(\mathbf{x}) \right)^2 d\mathbf{y} \\
&+ \lambda \left( 1 - \mathbf{v}(\mathbf{x})^T \mathbf{v}(\mathbf{x}) \right)
\end{aligned} \tag{4.5}$$

Differentiation of $e_{ls}(\mathbf{x})$ with respect to $v$, leads to eigenvalue decomposition problem $\mathbf{J_F}(\mathbf{x}) \, \hat{\mathbf{v}}(\mathbf{x}) = \lambda \, \hat{\mathbf{v}}(\mathbf{x})$. The 3D flux tensor $\mathbf{J_F}$ for the spatiotemporal volume centered at $(x, y)$ can be written in expanded matrix format as

$$\mathbf{J_F} = \begin{bmatrix}
\int_{\mathbf{\Omega}} \left\{ \frac{\partial^2 \mathbf{I}}{\partial x \partial t} \right\}^2 d\mathbf{y} & \int_{\mathbf{\Omega}} \frac{\partial^2 \mathbf{I}}{\partial x \partial t} \frac{\partial^2 \mathbf{I}}{\partial y \partial t} d\mathbf{y} & \int_{\mathbf{\Omega}} \frac{\partial^2 \mathbf{I}}{\partial x \partial t} \frac{\partial^2 \mathbf{I}}{\partial t^2} d\mathbf{y} \\[2mm]
\int_{\mathbf{\Omega}} \frac{\partial^2 \mathbf{I}}{\partial y \partial t} \frac{\partial^2 \mathbf{I}}{\partial x \partial t} \, d\mathbf{y} & \int_{\mathbf{\Omega}} \left\{ \frac{\partial^2 \mathbf{I}}{\partial y \partial t} \right\}^2 d\mathbf{y} & \int_{\mathbf{\Omega}} \frac{\partial^2 \mathbf{I}}{\partial y \partial t} \frac{\partial^2 \mathbf{I}}{\partial t^2} d\mathbf{y} \\[2mm]
\int_{\mathbf{\Omega}} \frac{\partial^2 \mathbf{I}}{\partial t^2} \frac{\partial^2 \mathbf{I}}{\partial x \partial t} d\mathbf{y} & \int_{\mathbf{\Omega}} \frac{\partial^2 \mathbf{I}}{\partial t^2} \frac{\partial^2 \mathbf{I}}{\partial y \partial t} d\mathbf{y} & \int_{\mathbf{\Omega}} \left\{ \frac{\partial^2 \mathbf{I}}{\partial t^2} \right\}^2 d\mathbf{y}
\end{bmatrix} \tag{4.6}$$

The elements of the flux tensor incorporate information about temporal gradient changes which leads to efficient discrimination between stationary and moving image features. Thus the trace of the flux tensor matrix which can be compactly written and computed as, $\mathbf{trace}(\mathbf{J_F}) = \int_{\mathbf{\Omega}} ||\frac{\partial}{\partial t} \nabla \mathbf{I}||^2 d\mathbf{y}$ can be directly used to classify moving

and non-moving regions without the need for expensive eigenvalue decompositions.

Figure 4.4 illustrates moving object detection results for high resolution WAMI images described in Figure 4.3 using the trace of the flux tensor that provides information about temporal gradient changes or moving edges. First row presents the original cropped ROI and the trace of flux tensor based moving edges results are shown in second row. Every pixel is classified as moving versus stationary by thresholding the trace of the corresponding flux tensor matrix. However, using this method 70% of the detected motions (averaged on 200 frames) are induced by parallax effect of tall structures which significantly degrades the precision of the motion detection results. The low precision of 20% is reported using only the trace of the flux tensor.

## 4.2 Context-aware Moving Vehicle Detection Using 2D Depth Maps

We studied that using purely conventional moving object detection methods would not be sufficient for a wide aerial motion imagery in which there are strong traces of parallax induced by tall buildings. The trace of the flux tensor or 3D median provides robust spatio-temporal information of moving objects but a large percentage of the detected motions are induced by parallax effects of tall structures as the camera viewpoint changes (Fig. 4.3 and Fig. 4.4). In order to reject undesirable detections due to tall structures, we develop a context-based semantic fusion approach to identify and remove such non-vehicle detections by using the depth map information with an active contour boundary refinement and filtering process.

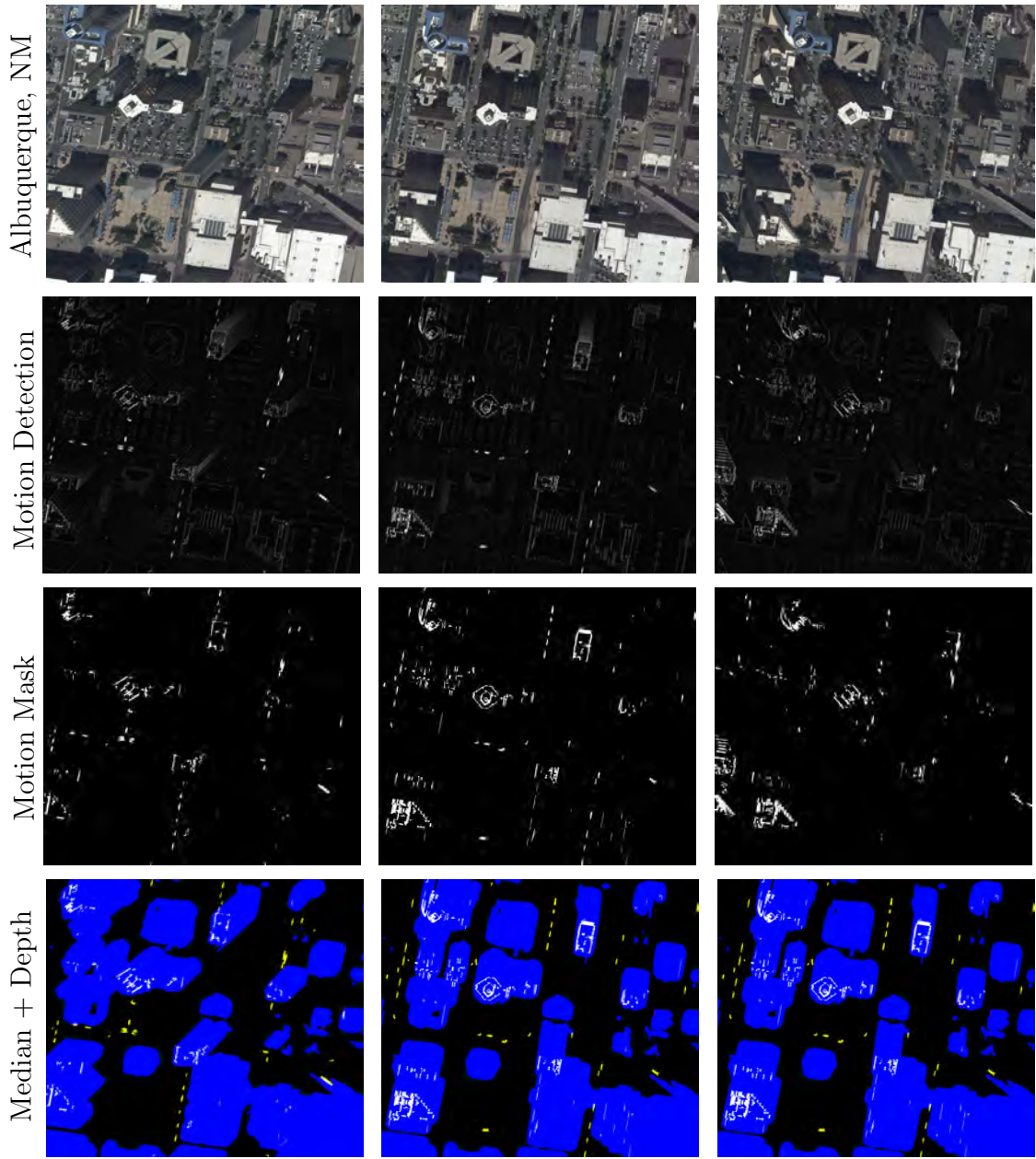

Figure 4.4: Illustration of motion detection results using the trace of the flux tensor: First row shows the registered $2k \times 2k$ cropped frames (003, 100, 199) from Albuquerque aerial imagery. Second row shows moving object detection results for the corresponding frames. Third and fourth row present the motion detection masks obtained from the trace of the flux tensor and the building masks in blue, respectively.

The accurate height of every pixel in the orthorectified temporal frames can be estimated using 3D point clouds or meshes resulting from dense multiview 3D reconstruction algorithms. In order to produce a frame specific building mask, the 3D point cloud or mesh is projected to produce a depth map that is then thresholded. Image pixels with a height value greater than a threshold value are identified as part of tall structures or buildings which will be used to remove flux tensor motion responses. Figure 4.5(a) illustrates the true motion detection produced by flux tensor (in yellow color) and undesirable moving detection caused by parallax in white color. The areas of tall structures are filtered by building mask and shown in blue. Provided ground-truth bounding-boxes are drawn in red to enable visual evaluation of the detection performance.

2D depth maps are projected from 3D point clouds that are obtained by 3D reconstruction of the scene. These point clouds have lower resolution compared to the analyzed images. Low resolution combined with 2D projection inaccuracies may result in filtering out correctly detected vehicles positioned close to tall structure (zoomed in Fig. 4.5(a)).

In order to refine the coarse building map $B_{dmap}$, we proposed to fuse the high resolution moving edges information from trace of the flux tensor with $B_{dmap}$ through a level-set based geodesic active contours framework.

The trace of flux tensor is used to construct an edge indicator function $g_F$ which will guide and stop the evolution of the geodesic active contour when it arrives at tall structure boundaries,

$$g_F(\mathbf{trace}(\mathbf{J_F})) = \frac{1}{1 + \mathbf{trace}(\mathbf{J_F})} \tag{4.7}$$

The edge indicator function is a decreasing function of the image gradient that rapidly

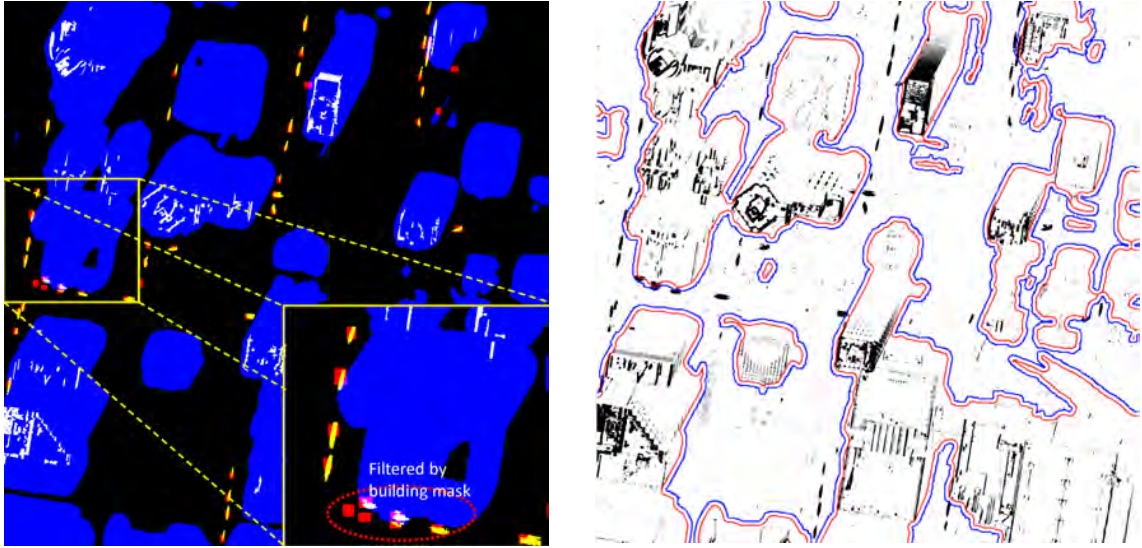

(a) Illustration of motion detection results: true motion detection produced by flux tensor (in yellow color) and false detection caused by parallax in white color. The areas with high altitude are filtered by building mask and are shown in blue. Provided ground-truth bounding-boxes are shown in red.

(b) Improved building mask using level-set based geodesic active contours: blue lines are the initial building contours which are evolved and stopped at building actual boundaries (red lines).

Figure 4.5: Building Mask Refinement Using Level Set Based Active Contours.

goes to zero along building edges and holds high values elsewhere.

Active contours evolve a curve $\mathcal{C}$, subject to constraints from a given image. In level set based active contour methods the curve $\mathcal{C}$ is represented implicitly via a Lipschitz function $\phi$ by $\mathcal{C} = \{(x, y) | \phi(x, y) = 0\}$, and the evolution of the curve is given by the zero-level curve of the function $\phi(t, x, y)$. Evolving $\mathcal{C}$ in a normal direction with speed $F$ amounts to solving the differential equation [102],

$$\frac{\partial \phi}{\partial t} = |\nabla \phi| F; \quad \phi(0, x, y) = \phi_0(x, y) \tag{4.8}$$

Unlike parametric approaches such as classical snake, level set based approaches ensure topological flexibility since different topologies of zero level-sets are captured implicitly in the topology of the energy function $\phi$. Topological flexibility is crucial for our application since we want to guide the coarse thresholded building mask to the actual building contours and reveal the filtered moving vehicles next to buildings. We used geodesic active contours [103] that are effectively tuned to trace of flux tensor edge information. The level set function $\phi$ is initialized with the signed distance function of the coarse building mask ($B_{dmap}$) and evolved using the geodesic active contour speed function,

$$\frac{\partial \phi}{\partial t} = g_F(\mathbf{trace}(\mathbf{J_F}))(c + \mathcal{K}(\phi))|\nabla \phi| + \nabla \phi \cdot \nabla g_F(\mathbf{trace}(\mathbf{J_F})) \qquad (4.9)$$

where $g_F(\mathbf{trace}(\mathbf{J_F}))$ is the fused edge stopping function (Eq. 4.7), $c$ is a constant, and $\mathcal{K}$ is the curvature term,

$$\mathcal{K} = div\left(\frac{\nabla \phi}{|\nabla \phi|}\right) = \frac{\phi_{xx}\phi_y^2 - 2\phi_x\phi_y\phi_{xy} + \phi_{yy}\phi_x^2}{(\phi_x^2 + \phi_y^2)^{\frac{3}{2}}} \qquad (4.10)$$

The force $(c + \mathcal{K})$ acts as the internal force in the classical energy based snake model. In this work, the constant velocity $c$ pushes the curve inwards to the tall structures. The regularization term $\mathcal{K}$ ensures boundary smoothness. The external image dependent force $g_F(\mathbf{trace}(\mathbf{J_F}))$ is used to stop the curve evolution at building boundaries edges. The term $\nabla g_F \cdot \nabla \phi$ introduced in [103] is used to increase the basin of attraction for evolving the curve to the boundaries of the objects.

Figure 4.5(b) shows the improved building contours results in red. The blue line are the initial building contours which are evolved and stopped at building actual boundaries. As it can be seen, the previously filtered detected cars by initial building mask are revealed and counted as true detections.

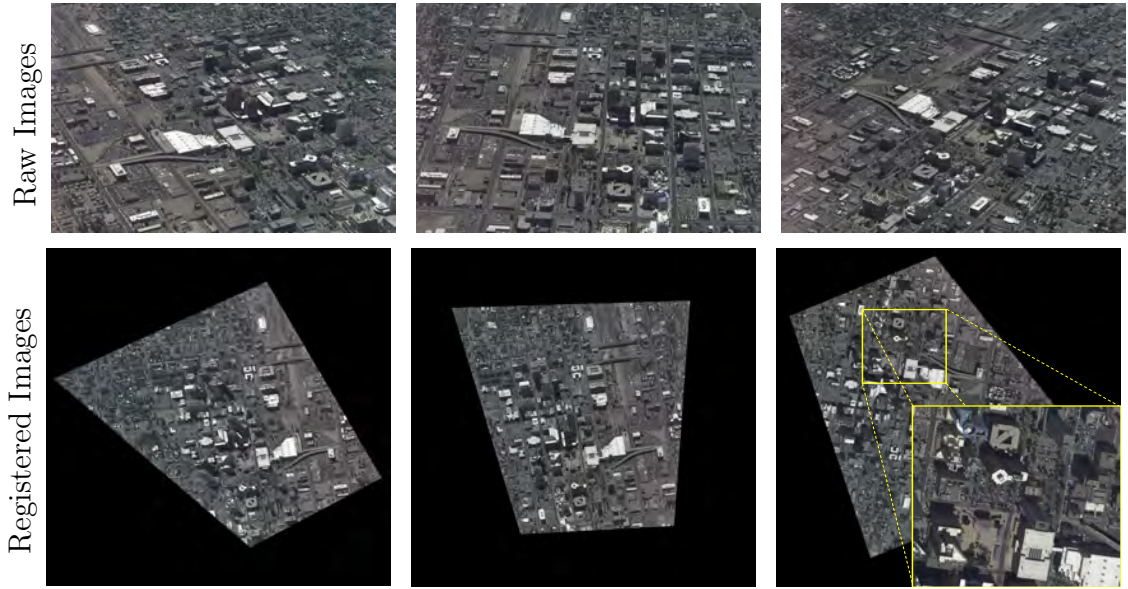

Figure 4.6: Top row shows original ultra-high resolution images (6600×4400) collected from an airborne WAMI platform flying over downtown Albuquerque, NM. Bottom row shows corresponding registered images (12000 × 12000) using the MU BA4S ortho-rectification module, an extremely fast bundle adjustment algorithm that avoids RANSAC iterations, and uses less than 12 minutes for 1071 images [54, 55].

## 4.3 Experimental Results

We elaborate and evaluate our proposed vehicle moving object detection results for ABQ aerial urban imagery which were collected by TransparentSky [74] using an aircraft with on-board IMU and GPS sensors flying 1.5 km above ground level of downtown Albuquerque, NM on September 3, 2013. Imaging was done at frame rate of 4Hz and 2.6 km orbit radius. Figure 4.6 shows samples of raw ultra high resolution images (6400 × 4400) with nominal ground resolution of 25cm and the corresponding registered images using *MU BA4S* registration approach which processes the total sequence of 1071 images in very short amount of time (less than 12 minutes). For evaluation, we carried out the experiments on the first 200, 2000 × 2000 cropped frames from location (4761, 5800) upper left corner in the 12K × 12K orthorectified images for which the ground-truth are provided by Kitware (Fig. 4.3 and Fig. 4.4).

In the first step, we applied the state of the art registration algorithm, *MU BA4S*, to orthorectify image sequences into a global reference system and to produce dense 3D point clouds [54, 55]. Then, *depth* or *height* maps are computed by projecting the 3D points into each camera view. In the third step, motion detection masks are obtained using either median-based background subtraction or the trace of the flux tensor. Finally, we fused building masks information extracted from depth maps with motion detection masks information to identify moving objects on the ground from motion induced by parallax effects of tall buildings and reject the false motion responses.

Figure 4.7 shows the original cropped ROI and the trace of flux tensor results. Figure 4.8 presents the results of the trace of flux tensor motion detection filtered by building mask. The left most image in Fig. 4.8 shows the flux tensor motion detection results in 2 colors; motion detections due to parallax are shown in white color and the rest are in yellow. In order to enable visual evaluation of the results ground truth bounding boxes are overlaid on flux tensor mask in red color. Height mask corresponding to the ortho-rectified ROI is shown in the middle. All the pixels with height values greater than a fixed threshold are considered as tall structures and are shown in blue in the rightmost image.

As discussed in Section 4.2 level-set based geodesic active contours is used to improve the building mask and reveal the filtered moving vehicles positioned next to the buildings. Improved building mask and final motion detection results are shown in Figure 4.9. In order to find the best 2D depth map thresholding value so that to achieve high precision and maintain high recall, we evaluated the performance of

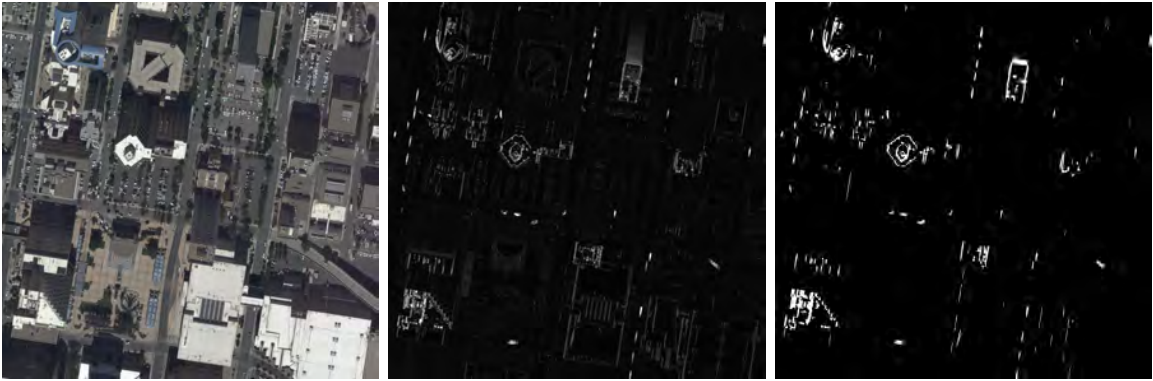

Figure 4.7: Illustration of motion detection using trace of flux tensor only: From left to right, cropped ROI of Albuquerque aerial imagery ($fr_{100}$), the spatio-temporal motion information computed by trace of flux tensor for the selected image, and flux tensor mask in which each pixel is identified as moving or stationary by thresholding the trace of flux tensor. Morphology is applied to improve the result.

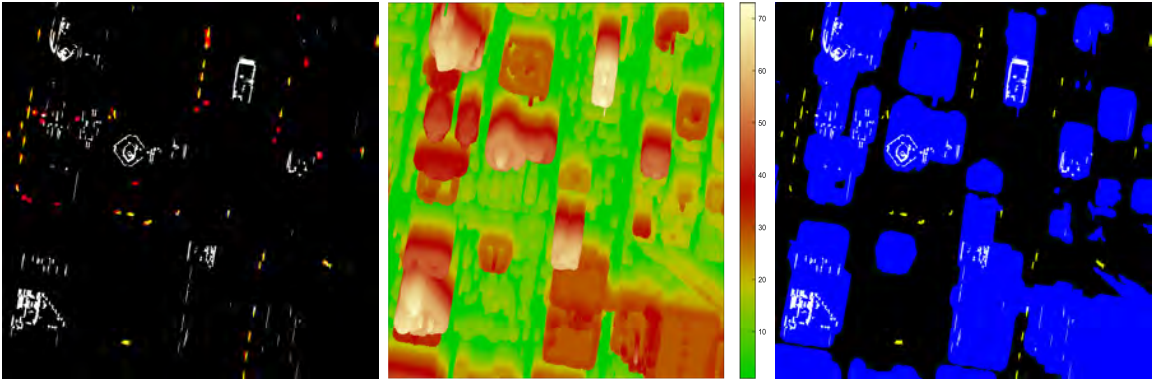

Figure 4.8: Illustration of motion detection results using trace of flux tensor filtered by depth mask. Left most image presents the motion detection results by thresholding the trace of flux tensor in 2 colors; motion detections due to parallax are shown in white besides other detection results in yellow color. In order to enable visual evaluation of the detection results ground-truth bounding boxes are overlaid on flux tensor mask in red color. Altitude mask corresponding to the orthorectified image is shown in the middle. All the pixels with altitude values greater than 20 meters are considered as tall structures and are shown in blue in the rightmost image.

the fused 3D median-based motion detection mask and 2D depth map when using

different height value to threshold the 2D depth map. Figure 4.10 shows the ROC

curve that represents the relation between *recall* (sensitivity) and *precision* using object-level detection evaluation methodology. As it is shown, threshold value of 20 is selected to achieve high precision of 85% and recall of 75%.

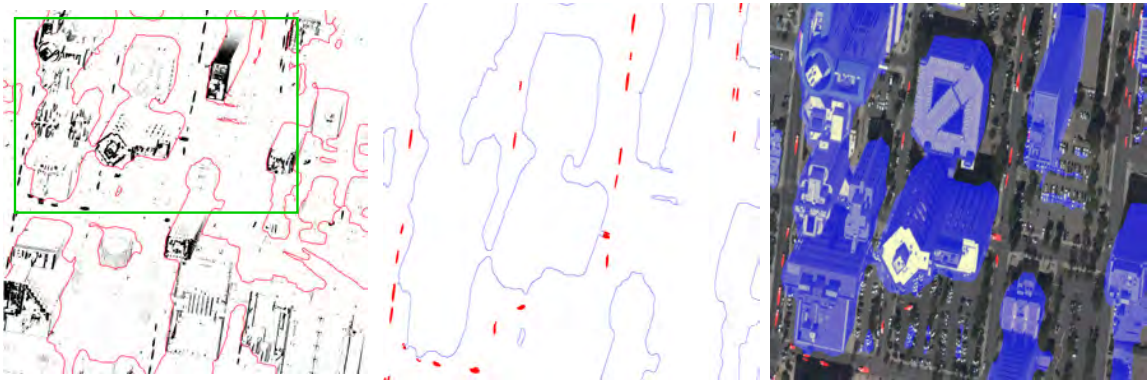

Figure 4.9: Moving object detection results using the proposed semantic fusion-based approach. The evolved building contours are shown in red in the left most image. The final moving object detection results of the region bounded in green box are shown in middle in red and final building contours in blue. Results are superimposed on the original image where building masks are shown in blue.

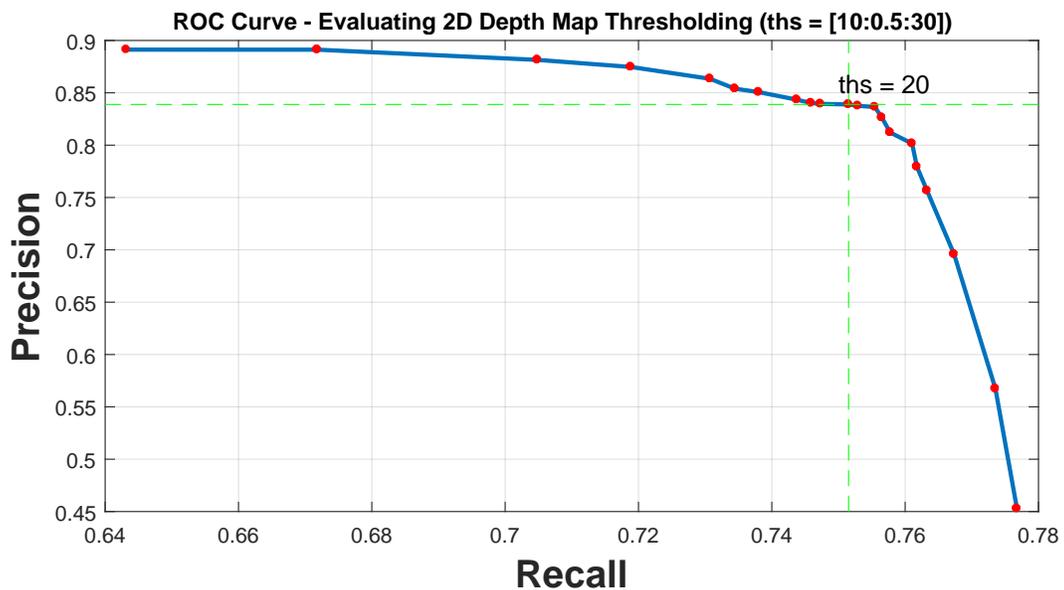

Figure 4.10: ROC curve Performance evaluation of moving object detection mask using different 2D depth map thresholding.

## 4.4 Evaluation Methodology

The general requirement for moving object detection algorithms is providing reasonable precision in terms of the number of true detected objects as well as high recall in objects contour detection [104]. Therefore we evaluated the performance of motion detection mask using two methodology: pixel-wise and object-wise detection performance evaluation.

### 4.4.1 Pixel-wise Performance Evaluation

The performance of motion detection results are evaluated after each stage of the fusion by computing the spatial precision and recall as

$$Precision = \frac{\sum_{i=1}^{N_D} |G_i \cap D_i|}{\sum_{i=1}^{N_D} |D_i|} = \frac{|TP|}{|TP| + |FP|} \tag{4.11}$$

$$Recall = \frac{\sum_{i=1}^{N_D} |G_i \cap D_i|}{\sum_{i=1}^{N_D} |G_i|} = \frac{|TP|}{|TP| + |FN|} \tag{4.12}$$

$$F_{measure} = 2 \times \frac{Precision \times Recall}{Precision + Recall} \tag{4.13}$$

where $G_i$ is the moving object bounding box presents in Ground Truth and $D_i$ is the segmented moving object obtained by motion detection algorithm. $N_D$ is the cardinality of the detected objects. Figure 4.11 presents the computed measures for the first 200 frames from Albuquerque sequence using 3D median and Figure 4.12 shows the results using the trace of the flux tensor. Table 4.1 reports the computed measures averaged on first 200 frames. The average precision of 73.2%, recall of 48.8% and $F_{measure}$ of 58.4% is reported using fusion of 3D median and 2D depth map.

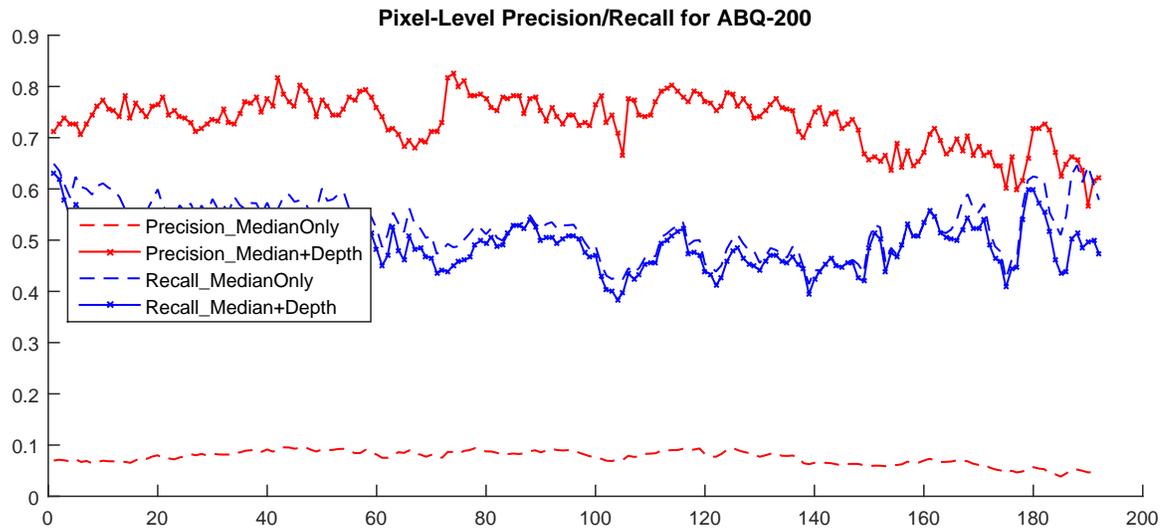

(a) Precision/Recall

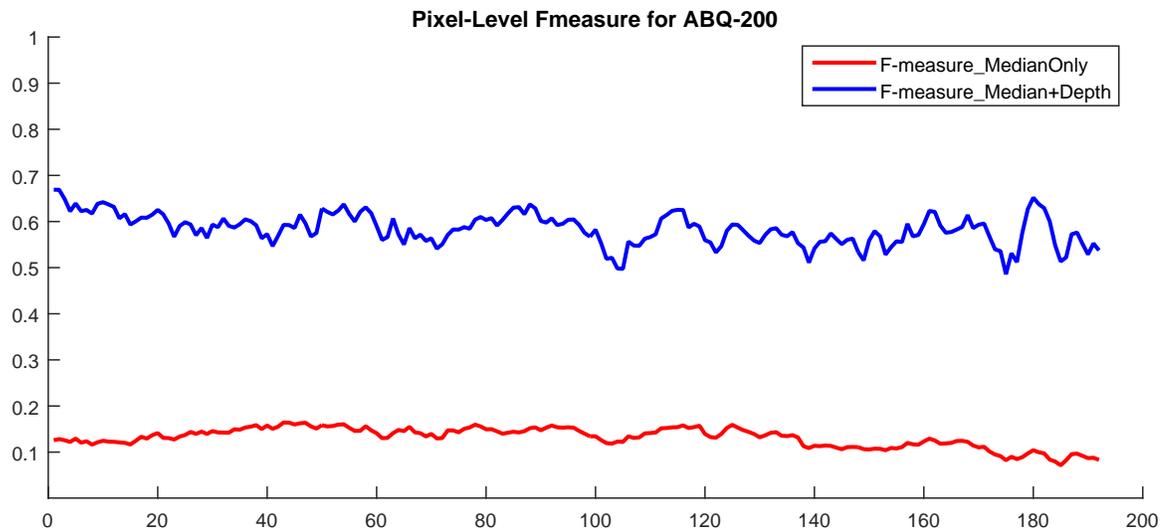

(b) $F_{measure}$

Figure 4.11:   Pixel-wise performance evaluation of proposed fused motion detection method. Top graph presents the computed precision and recall using 3D median motion mask and bottom graph illustrate the computed $F_{measure}$.

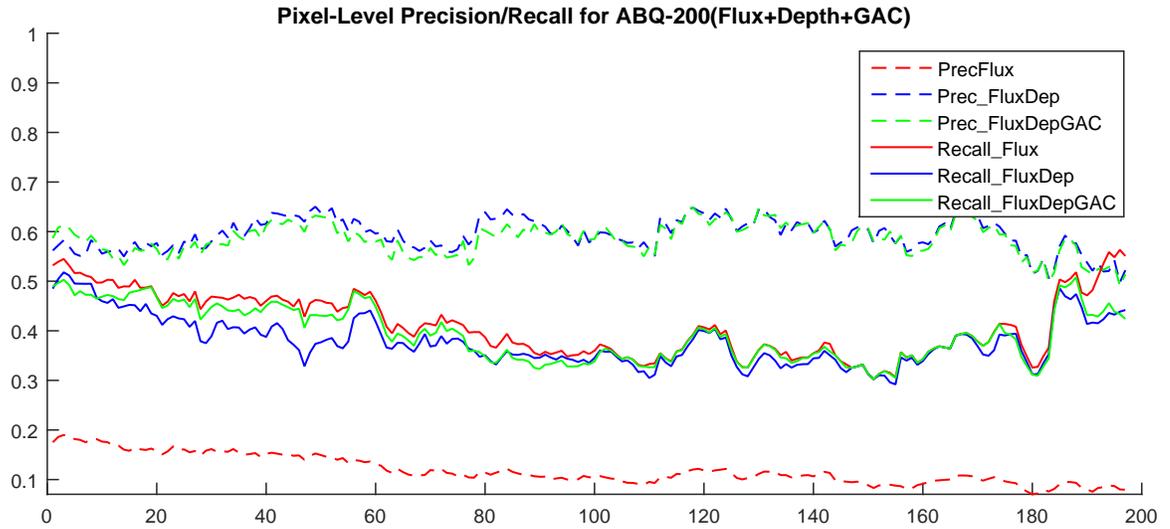

(a) Precision/Recall

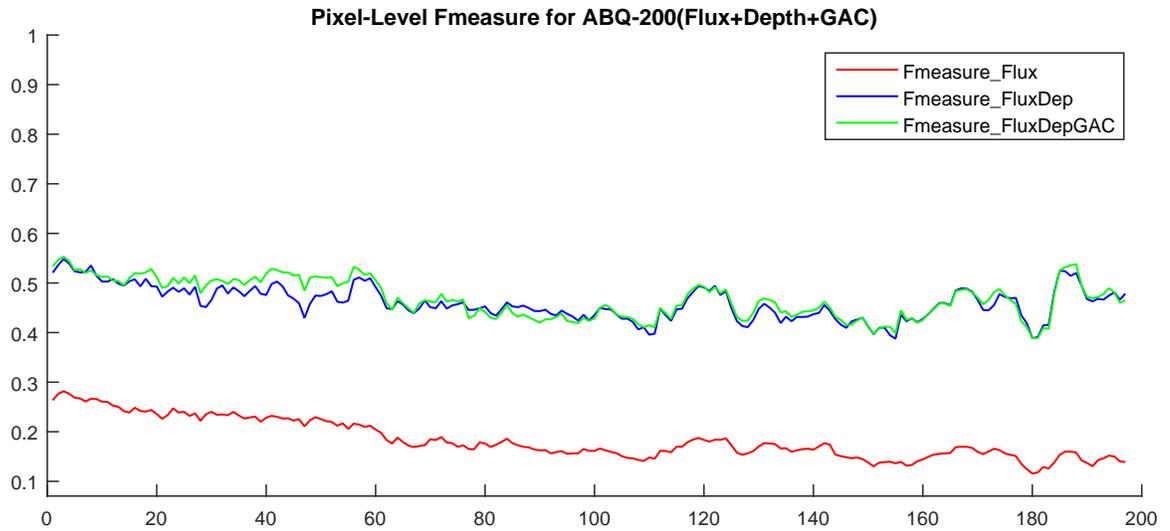

(b) $F_{measure}$

Figure 4.12: Pixel-wise performance evaluation of proposed fused motion detection method. Top graph presents the computed precision and recall using the trace of the flux tensor motion mask and bottom graph illustrate the computed $F_{measure}$.

Table 4.1: Average moving object detection performance results using pixel-level evaluation method.

| | Pixel-wise Evaluation | | | | |
|---|---|---|---|---|---|
| Average | Median Only | Median+Depth | Flux Only | Flux+Depth | Flux+Depth+GAC |
| Precision | 7.560 | 73.200 | 0.119 | 0.594 | 0.585 |
| Recall | 52.800 | 48.800 | 0.411 | 0.379 | 0.394 |
| $F_{measure}$ | 13.200 | 58.400 | 0.184 | 0.463 | 0.471 |

Table 4.2: Average moving object detection performance results using object-level evaluation method.

| | Object-wise Evaluation | | | | |
|---|---|---|---|---|---|
| Average | Median Only | Median+Depth | Flux Only | Flux+Depth | Flux+Depth+GAC |
| Precision | 0.129 | 0.839 | 0.200 | 0.867 | 0.836 |
| Recall | 0.801 | 0.751 | 0.779 | 0.735 | 0.759 |
| $F_{measure}$ | 0.222 | 0.792 | 0.318 | 0.796 | 0.796 |

## 4.4.2 Object-wise Performance Evaluation

Since the ultimate goal of the proposed motion detection system is to perform persistent tracking of moving vehicles, we have evaluated detection performance using object level measures as well. Associations of the detected moving blobs to ground truth objects is performed using a bidirectional correspondence analysis described in [105, 106] that handles not only one-to-one matches but also merge and fragmentation cases. Figure 4.13 presents the performance evaluation results using object-level methods and 3D Median Motion mask. It can be seen that precision has been drastically increased while recall remained almost the same. Average precision of 83.9% and recall of 75.1% is achieved. Table 4.2 reports the average of computed statistics for 200 frames.

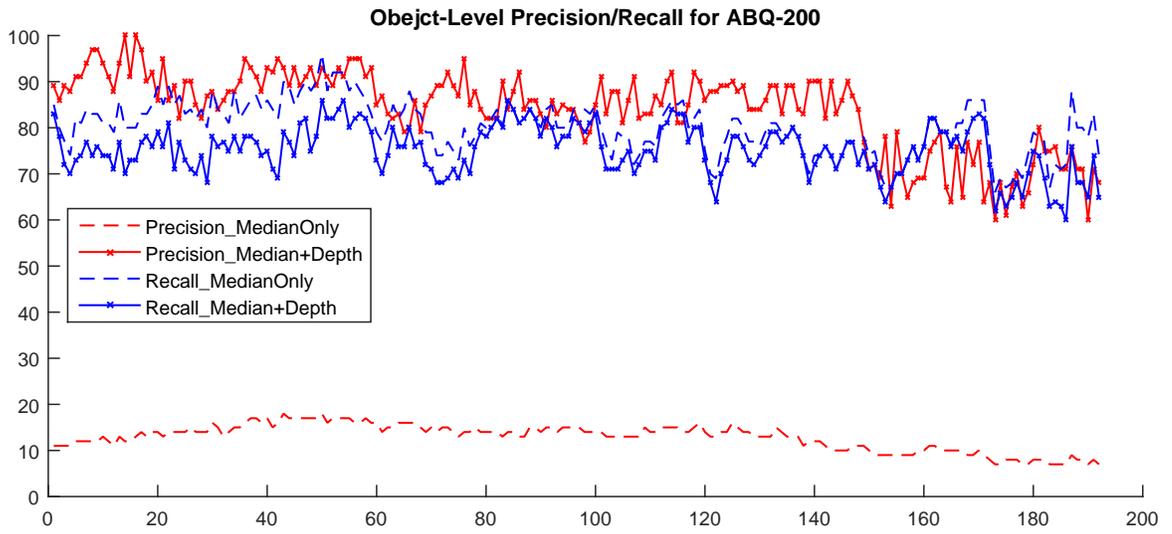

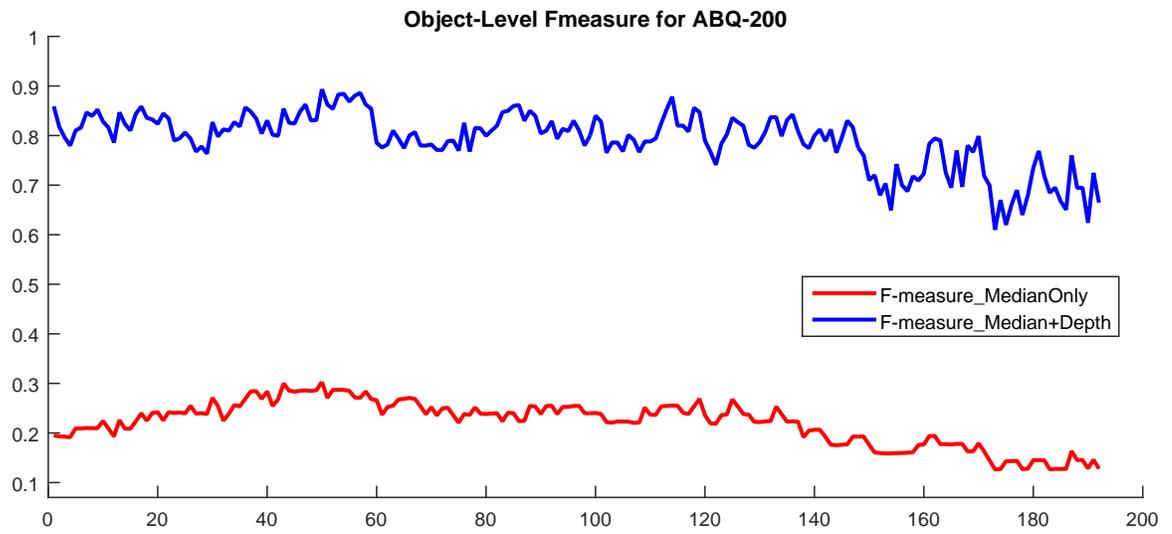

Figure 4.13: Performance evaluation of our proposed fused motion detection method.

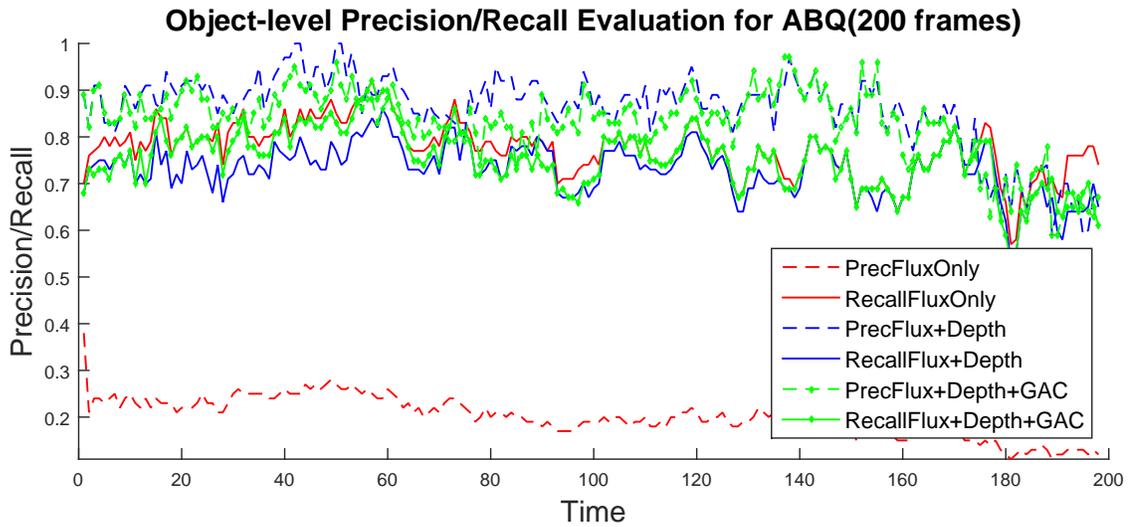

(a) Precision and Recall

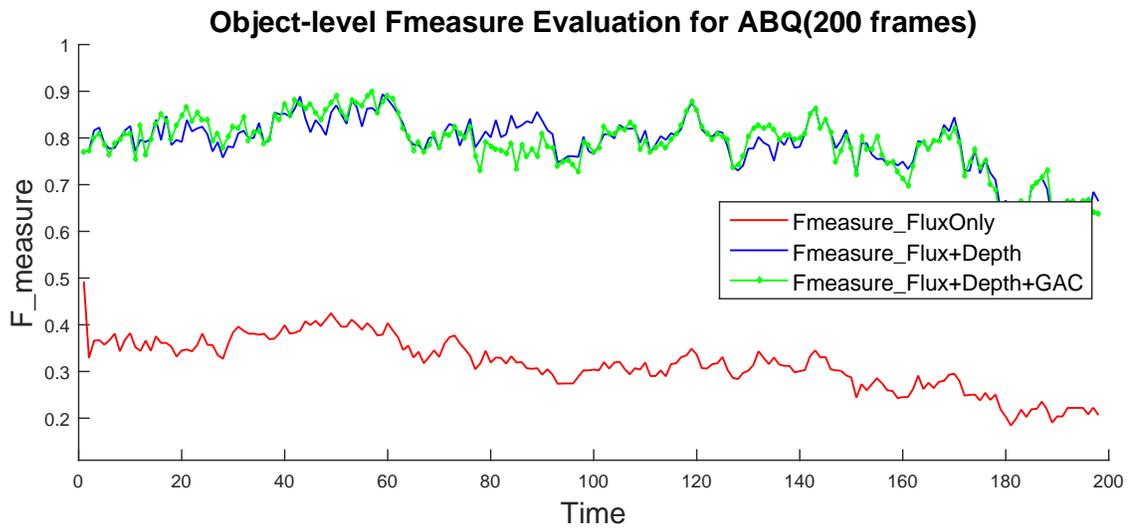

(b) $F_{measure}$

Figure 4.14: Object-Level performance evaluation of proposed fused moving object detection method using the trace of the flux tensor motion mask.

# Chapter 5

# Fast GPU Implementation of Integral Histogram

Integral histogram provides an optimum and complete solution for the histogram-based search problem. Porikli [64] generalized the concept of integral image and presented computationally very fast method to extract the plain histogram of any arbitrary region in constant time. Integral histograms provide not only an efficient computational framework for extracting histogram-based regional descriptors, but also enable low cost multi-scale image analysis. Descriptors for different scales can be generated in constant time without recomputing the integral histogram, since regional histograms for any-size rectangular regions can be derived from a given integral histogram.

Many novel approaches have been presented based on integral histogram to accelerate the performance of image processing tasks including filtering [1, 89, 2], recognition and classification [88, 107, 108], image retrieval [109], object segmentation

**Algorithm 5:** Sequential Integral Histogram

**Input :** Image **I** of size $h \times w$, number of bins b
**Output :** Integral histogram tensor **H** of size $b \times h \times w$
 1: **Initial H:**
   $\mathbf{H} \leftarrow 0$
 2: **for** k=1:b **do**
 3:   **for** x=1:h **do**
 4:     **for** y=1:w **do**
 5:       $H(k, x, y) \leftarrow H(k, x - 1, y) + H(k, x, y - 1)$
         $-H(k, x - 1, y - 1) + Q(k, I(x, y))$
 6:     **end for**
 7:   **end for**
 8: **end for**

[110, 111, 112], object detection [31, 113, 114, 115, 116], visual tracking [92, 117, 118, 119, 25, 27, 28, 120], etc.

Our proposed visual tracking system *SPCT* utilized integral histogram as the building block to compute candidate regions local histograms in constant time. Although integral histogram enables fast exhaustive search but it is still considered as the most compute intensive image processing task for the presented tracking system. The sequential implementation of the integral histogram uses an $O(N)$ recursive row-dependent method, for an image with $N$ pixels (5). Therefore, we explored different techniques to efficiently compute integral histograms on GPU architecture using the NVIDIA CUDA programming model [121, 122].

The contributions of the work can be summarized as follows:

- We described four GPU implementations of the integral histogram: Cross-Weave Baseline (CW-B), Cross-Weave Scan-Transpose-Scan (CW-STS), customized Cross-Weave Tiled horizontal-vertical Scan (CW-TiS) and Wave-Front Tiled Scan (WF-TiS). All the implementations rely on parallel cumulative sums on row and column histograms. The first three designs operate a cross-weave scan, and the latter does a wave-front scan.

- In our implementations, we show a trade-off between productivity and efficiency. In particular, the less efficient CW-B and CW-STS solutions rely on existing open-source kernels, whereas the most efficient CW-TiS and WF-TiS designs are based on custom parallel kernels. We show analogies between the computational pattern of the integral histogram and that of the bioinformatics Needleman-Wunsch algorithm. We leverage their similarity to design our most efficient implementation (WF-TiS).

- We relate the performances of our proposed implementations to their utilization of the underlying GPU hardware, and use this analysis to gradually improve over the naive CW-B scheme.

- When computing the 32-bin integral histogram of a $512 \times 512$ image, our custom implementation WF-TiS achieves a frame rate of 135 $fr/sec$ on Tesla K40c (Fig. 5.16(c)) and 351 on a GeForce GTX Titan X graphics card (Fig. 5.16(d)). Further, our GPU WF-TiS design reports a 60X speedup over a serial CPU implementation, and a 8X to 30X speedup over a multithreaded implementation deployed on an 8-core CPU server (Fig. 5.18(b)).

- We have exploited task-parallelism to overlap computation and communication across the sequence of images. Using dual-buffering improved the performance by a factor of two when computing 16-bins integral histogram of HD ($1280 \times 720$) images versus no dual-buffering on GeForce GTX 480 (Fig. 5.12).

- We evaluated utilizing multiple GPUs for large scale images due to the limited GPU global memory. The integral histogram computations of different bins are distributed across available GPUs using a task queue. We achieved the increasing speedup range from 3X for HD images to 153X for the large 64MB images and 128 bins over the single threaded CPU implementation (Fig. 5.17).

## 5.1    Kernel Optimization for Integral Histogram

In this section, we first describe the integral histogram data structure and its layout in GPU memory and then present different optimization strategies. In our first implementations [44], we reuse existing parallel kernels from the NVIDIA Software Development Kit (SDK); we refer to these as generic kernels. We point out the limitations of such an approach, and progressively refine our implementation in order to better utilize the architectural features of the GPU. This leads to the evolution of four techniques to compute the integral histogram on GPUs that trade-off productivity with efficiency and a discussion of how the performance of the proposed implementations reflect their utilization of the underlying hardware. The first three implementations perform cumulative sums on row and column histograms in a cross-weave (CW) fashion, whereas the fourth one performs a wavefront (WF) scan.

### 5.1.1    Data Structure Design

An image with dimensions $h \times w$ produces an integral histogram tensor of dimensions $b \times h \times w$, where $b$ is the number of bins in the histogram. This tensor can be represented as a 3-D array, which in turn can be mapped onto a 1-D row major ordered array as shown in Figure 5.1. It is well known that the PCI-express connecting CPU and GPU is best utilized by performing a single large data transfer rather than many small data transfers. Therefore, whenever the 1-D array representing the integral histogram fits the available GPU global memory, we transfer it between GPU and CPU with a single memory transaction. The computation of larger integral histograms is tiled along the bin-direction and distributed between available GPUs: portions of the 1-D array corresponding to the maximum number of bins that fit the GPU capacity are transferred between GPU and CPU in a single transaction. For all the considered image sizes a single bin fits the GPU memory; however, our implementation can be easily extended to images exceeding the GPU capacity by tiling the computation also column-wise. Finally, we experimentally verified that initializing the integral histogram on GPU is more efficient than initializing it on CPU and then transferring it from CPU to GPU. Therefore, in all our GPU implementations, we initially transfer the image from CPU to GPU, then initialize and compute the integral histogram on GPU, and finally transfer it back from device to host.

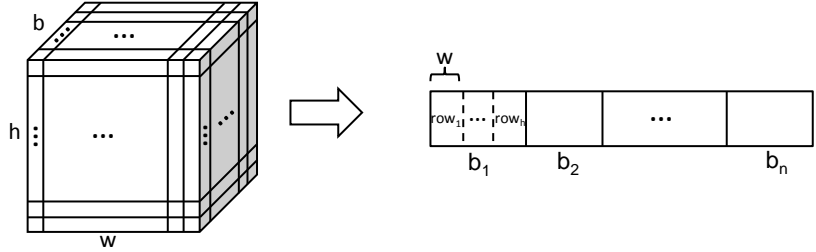

Figure 5.1: Integral histogram tensor represented as 3-D array data structure (left), and equivalent 1-D array mapping (right).

### 5.1.2 Naive Cross-weave Baseline Parallelization(CW-B)

The sequential implementation of the integral histogram is represented by Algorithm 5. As can be seen, the algorithm can be trivially parallelized along the $b$- dimension, since the computation of different bins can be done fully independently. However, the algorithm presents loop-carried dependences along both the $x-$ and the $y-$ dimensions. Therefore, we need an intelligent mechanism for inter-row and inter-column parallelization. The cross-weave scan mode enables cumulative sum tasks over rows (or columns) to be scheduled and executed independently allowing for inter-row and column parallelization. We observe that NVIDIA CUDA SDK provides an efficient implementation of the *all-prefix-sums* [123] and of the *2-D transpose* [124] operations. Therefore, leveraging these existing open source kernels, we can quickly implement the integral histogram. Algorithm 6 presents our cross-weave baseline approach (CW-B) to compute the integral histogram, combining cross-weave scan mode with the existing parallel prefix sum and 2-D transpose implementations. In this approach, we first apply prefix-sums to the rows of the histogram bins (horizontal cumulative sums or prescan), then transpose the array corresponding to each bin using a 2-D transpose, and finally reapply the prescan to the rows of the transposed histogram to obtain the integral histograms at each pixel. We now briefly describe the two parallel kernels available in NVIDIA SDK.

**Parallel Prefix Sum Operation on the GPU**

The core of our CW-B approach is the parallel prefix sum algorithm [123]. The *all-prefix-sums* operation (also refered as *scan*) applied to an array generates a new array where each element $k$ is the sum of all values preceding $k$ in the scan order. Given an array $[a_0, a_1, ..., a_{n-1}]$ the prefix-sum operation returns

$$[0, a_0, (a_0 + a_1), ..., (a_0 + a_1 + ... + a_{n-2})] \tag{5.1}$$

---
**Algorithm 6:** CW-B: Naive Cross-weave Baseline Parallelization
---

**Input** : Image $\mathbf{I}$ of size $h \times w$, number of bins b
**Output** : Integral histogram tensor $\mathbf{IH}$ of size $b \times h \times w$

1: **Initialize IH**
   **IH** $\leftarrow 0$
   $\mathbf{IH}(\mathbf{I}(\mathbf{w}, \mathbf{h}), \mathbf{w}, \mathbf{h}) \leftarrow 1$
2: **for** all bins $b$ **do**
3:    **for** all rows $x$ **do**
4:       //horizontal cumulative sums
           $IH(x, y, b) \leftarrow IH(x, y, b) + IH(x, y - 1, b)$
5:    **end for**
6: **end for**
7: **for** all bins $b$ **do**
8:    //transpose the bin-specific integral histogram
        $IH^T(b) \leftarrow$ 2-D Transpose$(IH(b))$
9: **end for**
10: **for** all bins $b$ **do**
11:   **for** all rows $y$ of $IH^T$ **do**
12:     //vertical cumulative sums
         $IH^T(y, x, b) \leftarrow IH^T(y, x, b) + IH^T(y, x - 1, b)$
13:   **end for**
14: **end for**
---

The parallel prefix sum operation on GPU consists of two phases: an *up-sweep* and a *down-sweep* phase (see Fig. 5.2). The *up-sweep* phase builds a balanced binary tree on the input data and performs one addition per node. Scanning is done from the leaves to the root. In the *down-sweep* phase the tree is traversed from root to the leaves and partial sums from the up-sweep phase are aggregated to obtain the final scanned (prefix summed) array. Prescan requires only $O(n)$ operations: $2 \times (n - 1)$ additions and $(n - 1)$ swaps. Padding is applied to shared memory addresses to avoid bank conflicts.

### GPU-based 2D Transpose Kernel

The integral histogram computation requires two prescans over the data: a horizontal prescan that computes cumulative sums over rows of the data, followed by a vertical prescan that computes cumulative sums over the columns of the first scan output. Taking the transpose of the horizontally prescanned image histogram enables us to reapply the same (horizontal) prescan algorithm effectively on the columns of the data. The transpose operation can be performed using the efficient 2D kernel described in [124]. This tiled implementation uses shared memory to avoid uncoalesced read and write accesses to global memory, and uses padding to avoid shared memory conflicts and thereby optimize the shared memory accesses.

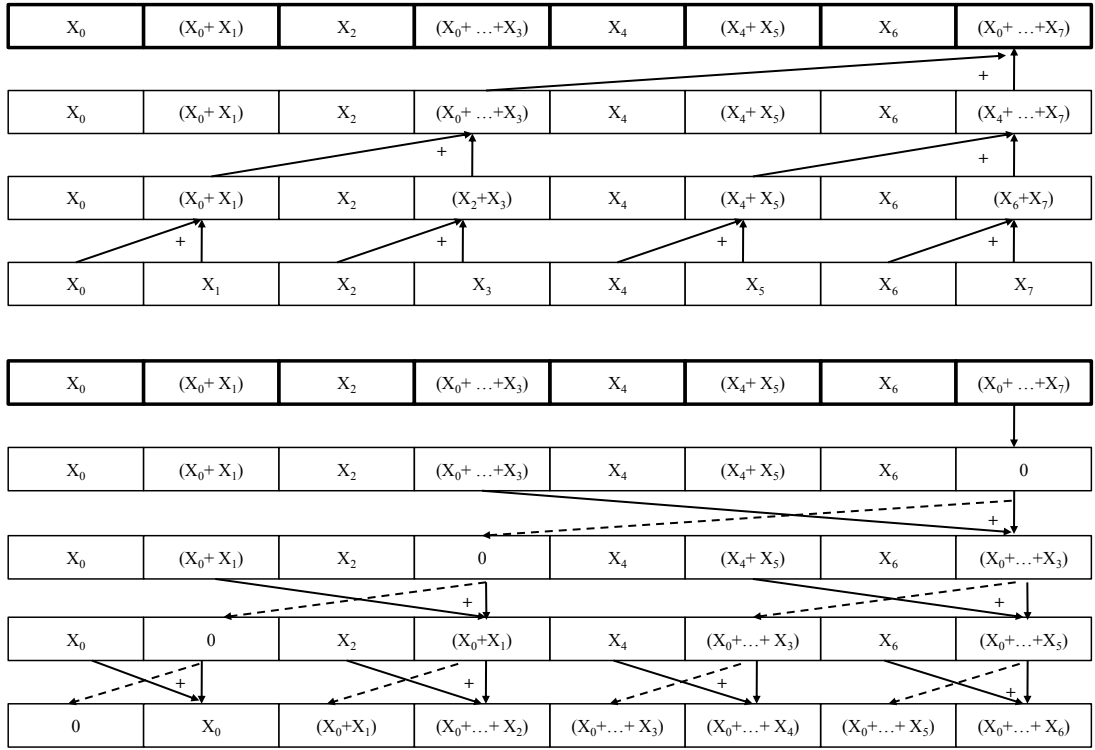

Figure 5.2: Parallel prefix sum operation, commonly known as exclusive scan or prescan [123]. Top: Up-sweep or reduce phase applied to an 8-element array. Bottom: Down sweep phase.

### 5.1.3 Cross-weave Scan-Transpose-Scan Parallelization (CW-STS)

As can be observed in Algorithm 6, the CW-B implementation performs many kernel invocations: $b \times h$ times horizontal scans each of size $w$, $b$ times 2-D transposes each of size $w \times h$, and $b \times w$ times vertical scans each of size $h$. In other words, this approach is based on the use of many parallel kernels, each of them performing very little work and therefore greatly under-utilizing the GPU. With the regular image sizes in consideration, for example, the scan kernels are invoked on arrays of size varying from 512 to 2048; however, the all-prefix-sum kernel has been designed to perform well on arrays consisting of millions of elements. Therefore, an obvious way to improve the integral histogram implementation is to increase the amount of work performed by each kernel invocation, and reduce kernel invocation overheads. This, in turn, will improve the GPU utilization. However, as shown in Algorithm 7, the computation can be easily broken into three phases: a single horizontal scan, a single 3-D transpose, and a single vertical scan. We call this solution cross-weave scan-transpose-scan (CW-STS) parallelization. We observe that the CW-STS approach

---

**Algorithm 7:** CW-STS: Single Scan-Transpose-Scan Parallelization

---
**Input :** Image $\mathbf{I}$ of size $h \times w$, number of bins b
**Output :** Integral histogram tensor $\mathbf{IH}$ of size $b \times h \times w$
  1: **Initialize IH**
          $\mathbf{IH} \leftarrow 0$
          $\mathbf{IH(I(w,h),w,h)} \leftarrow 1$
  2: **for** all $b \times h$ blocks in parallel **do**
  3:    **Prescan**$(IH)$
  4: **end for**
  5: //transpose the histogram tensor
          $IH^T \leftarrow$ 3D Transpose$(IH)$
  6: **for** all $b \times w$ blocks in parallel **do**
  7:    **Prescan**$(IH^T)$
  8: **end for**

---

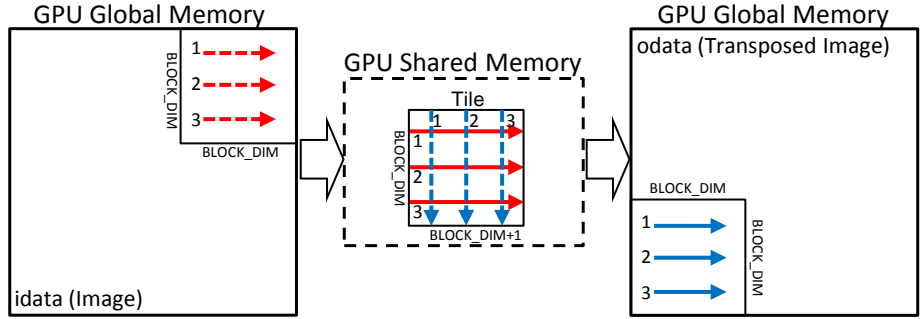

Figure 5.3: Data flow between GPU global memory and shared memory within the transpose kernel; stage 1 in red, stage 2 blue, reads are dashed line, writes are solid lines.

does not require rewriting the prescan kernel, but only invoking it judiciously. As explained above, the all-prefix-sum operation consists of two phases: the up-sweep computes partial sums, and the down-sweep aggregates them into the final result. However, the problem is that in the early stages of the up-sweep phase much more threads are involved in the computation than in the later stages (leading to GPU under-utilization). The down-sweep phase has the same problem except that the thread utilization, initially minimal, increases with the computational stage (Figure 5.2). To avoid this problem, we proposed the CW-TiS and WF-TiS approaches which compute all row- or column- wise scans with a single kernel call. Specifically, in the horizontal and vertical scan phase we invoke the custom prescan kernel *once* using a 2-D grid of size $(b, \frac{w \times h}{2 \times Num\_Threads})$.

In order to allow a single transpose operation, we need to transform the existing 2-D transpose kernel into a 3-D transpose kernel. This can be easily done by using the bin offset in the indexing. The 3-D transpose kernel is launched using a 3-D grid of dimension $(b, \frac{w}{\text{BLOCK\_DIM}}, \frac{h}{\text{BLOCK\_DIM}})$, where BLOCK_DIM is the maximum number of banks in shared memory (32 for all graphics card used). Figure 5.3 shows the data flow in the transpose kernel. A tile of size $BLOCK\_DIM * BLOCK\_DIM$ is written to the GPU shared memory into an array of size $BLOCK\_DIM * (BLOCK\_DIM + 1)$. This pads each row of the 2-D block in shared memory so that bank conflicts do not occur when threads address the array column-wise. Each transposed tile is written back to the GPU global memory to construct the full histogram transpose.

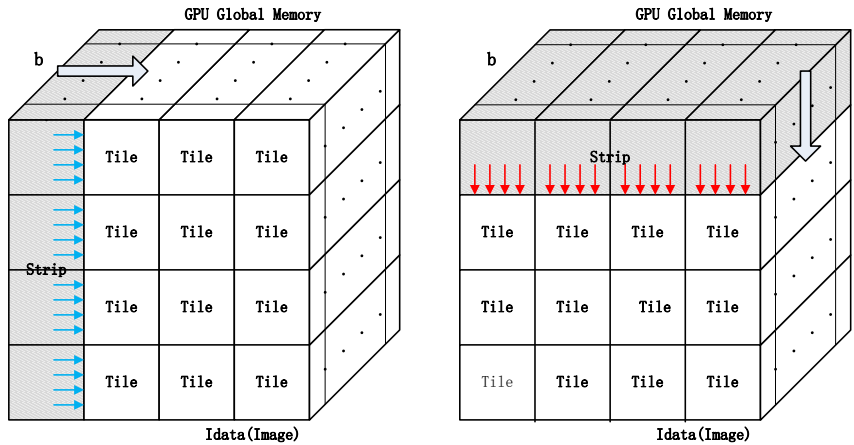

Figure 5.4: Cross-weave Tiled Horizonal-Vertical Scan (CW-TiS): Tiled horizontal scan (left), Tiled vertical scan (right). Both scans are performed in tile and the shadow area presents one strip.

### 5.1.4 Cross-weave Tiled Horizontal-Vertical Scan Parallelization (CW-TiS)

To understand how to further improve the integral histogram computation, we can observe the following. First, the use of the transpose kernel in the CW-STS implementation is motivated by the reuse of the prescan kernel in the vertical scan phase. However, the transpose operation can take considerable time compared to the prescan. For instance, when the image size is $512 \times 512$ and the histogram consists of 32 bins, the transpose takes about 20% of the whole computation time, and almost 50% of the time of a single prescan (Figure 5.7). Therefore, the execution time can be greatly reduced by combining the transpose and the vertical prescan into a single parallel kernel.

Second, the prescan kernel has its own limitations. While in the early stages of the up-sweep phase many threads are involved in the computation, in later stages only a few threads are active. For instance, as can be observed in Fig. 5.2, the number of active threads decreases from 4 at the beginning to 1 at the end. The down-sweep phase has the same problem (except that the thread utilization, initially minimal, increases with the computational stage). In general, the up-sweep and down-sweep phases of a scan on an array of $n$-element will consist of $2 \times log_2(n)$ iterations. In the up-sweep the number of active threads, initially equal to $\frac{n}{2}$, halves at every iteration. The number of working cycles of all active threads is therefore equal to $3 \times (n-1)$. The efficiency, defined as the ratio of the total working cycles over the product of the number of threads and the number of iterations will be

$$\frac{3 \times (n-1)}{n \log_2 n} \approx \frac{3}{\log_2 n} \tag{5.2}$$

For example, the efficiency of the scan on a 1024-element 1-D array is only 30%. To achieve a better efficiency, we can leverage the data-level parallelism underlying the integral histogram computation. As mentioned before, different bins can be processed fully independently. In addition, for each bin, the horizontal scan can be performed fully independently on the $h$ rows, and the vertical scan can be performed in parallel on the $w$ columns. Based on these observations, we propose a cross-weave tiled horizontal and vertical scan (CW-TiS) method that: (i) eliminates the need for the transpose operation and benefits less memory, and (ii) better utilizes the data-level

---

**Algorithm 8:** CW-TiS: Cross-weave Tiled Horizonal-Vertical Scan Parallelization

---

**Input :** Image **I** of size $h \times w$, number of bins b
**Output :** Integral histogram tensor **IH** of size $b \times h \times w$

1: **Initialize IH**
    **IH** $\leftarrow 0$
    **IH**(**I**(**w**, **h**), **w**, **h**) $\leftarrow 1$
2: **for** all bins b in parallel **do**
3:    **for** all vertical strips $v_s$ of width $TILE\_SIZE$ **do**
4:       **Tiled_Horizontal_Scan(IH)**
5:    **end for**
6:    **for** all horizontal strips $h_s$ of height $TILE\_SIZE$ **do**
7:       **Tiled_Vertical_Scan(IH)**
8:    **end for**
9: **end for**

---

parallelism of the integral histogram. Algorithm 8 represents how CW-TiS operates. First, each of the $b$ matrices of size $(h \times w)$ corresponding to different bins is divided into *tiles*. Each tile must be small enough to fit in shared memory and large enough to contain sufficient amount of data for computation work. In our implementation, we use squared tiles. The processing is divided into two stages: the horizontal scan (Fig. 5.4 (left)) and the vertical scan (Fig. 5.4 (right)). In each stage, the computation is performed strip-wise until the whole matrix has been processed. A kernel call operates on one strip of size $tile\_width \times image\_height$ in horizontal scan versus $image\_width \times tile\_height$ during the vertical scan. The number of vertical strips in horizontal scan equals to $VStrips = \frac{w_{Image}}{w_{Tile}}$. Whereas the number of horizontal strips during the vertical scan is $HStrips = \frac{h_{Image}}{h_{Tile}}$. Therefore, The total number of image tiles or blocks being processed is given by:

$$Tiles = \frac{w_{Image} \times h_{Image}}{w_{Tile} \times h_{Tile}} = VStrips \times HStrips \qquad (5.3)$$

We expect the image sizes to be evenly divisible by the tile sizes otherwise the image will be appropriately padded. In the kernel implementation, each thread-block is assigned to a tile and each thread to a row/column. Shared memory is used to allow efficient and coalesced memory accesses. Threads belonging to the same block push the cross-weave forward (either from left to right or from top to bottom (Fig. 5.4). Since each thread-block consists of warps, in order to avoid thread divergence within warps and GPU underutilization, the tile size is set to be a multiple of the warp size (32).

### 5.1.5 Wave-front Tiled Scan Parallelization(WF-TiS)

The use of separate horizontal and vertical scan kernels in the CW-TiS method has a drawback: it causes each tile to be transferred multiple times between global and shared memory. In fact, in both scan kernels, each tile is first moved from global into shared memory, processed and then moved back to global memory. As a consequence, combining the horizontal and vertical scans into a single kernel will allow accessing

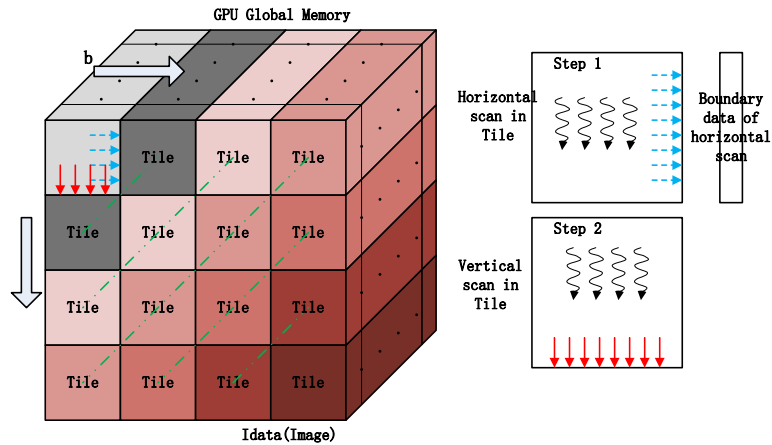

Figure 5.5: "WF-TiS" implementation. (Left) Tiles with the same color belong to the same stride and are executed in the same kernel launch iteration. (Right) The horizontal and vertical scan of a tile. It starts from the horizontal scan, store the boundary data into extra memory and finally the vertical scan.

---

**Algorithm 9:** WF-TiS: Wave-front Tiled Scan Parallelization

---

**Input :** Image **I** of size $h \times w$, number of bins b
**Output :** Integral histogram tensor **IH** of size $b \times h \times w$
  1: **Initialize IH**
       **IH** $\leftarrow 0$
       **IH(I(w, h), w, h)** $\leftarrow 1$
  2: **for** all diagonal strips $s$ of width $TILE\_SIZE$ **do**
  3:    **for** all bins b in parallel **do**
  4:       **Tiled_Wavefront_Scan()**
  5:    **end for**
  6: **end for**

---

global memory only twice per tile (once in read, and once in write mode). Before introducing the detailed implementation, let us briefly analyze the data dependences of the integral histogram. For the horizontal scan, the data in each row rely on the data on their left; for the vertical scan, the data in each column rely on the data on their upper position. This data access pattern is quite similar to that in the GPU implementation of the Needleman-Wunsch algorithm [125] in the Rodinia Benchmark Suite [126]. Therefore, we can arrange the computation in a similar fashion, and compute the integral histogram using a wave-front scan. Algorithm 9 represents the latest approach which we called Wave-front Tiled Scan Parallelization (WF-TiS).

Similarly to the CW-TiS implementation, we divide the $h \times w$ matrix into different tiles as shown in Fig. 5.5. Again, each tile should be small enough to fit in shared memory, and large enough to contain non-trivial amount of computation work. All the tiles lying on the same diagonal line with different bins (they are presented with the same color) are considered part of the same strip and processed in parallel. Therefore, for an image of size $w \times h$ and tile size $TILE\_SIZE$ the total number of iterations are

$$\lceil \frac{w}{TILE\_SIZE} \rceil + \lceil \frac{h}{TILE\_SIZE} \rceil - 1 \qquad (5.4)$$

Within the parallel kernel, each thread block will process a tile, and each thread will process a row (during horizontal scan) and a column (during the vertical scan) of the current tile. The tricky part of this implementation is that after the horizontal scan, the last column of each tile must be preserved for horizontal scan of the next strip before being overwritten during the vertical scan. This can be achieved by storing the extra data in global memory (the additional required memory is an array of $h$ elements). The WF-TiS method can potentially be preferable to the CW-TiS by eliminating unnecessary data movements between shared and global memory.

## 5.2   Experimental Results

In this section, we present a performance evaluation of our proposed GPU implementations of the integral histogram. Our experiments were conducted on four GPU cards:

- GeForce GTX Titan X graphics card - equipped with $24 \times 128 - core$ SMs, maxwell architecture, $12GB$ of global memory and compute capability 5.2.

- Nvidia Tesla K40c - equipped with $15 \times 192 - core$ SMs, kepler architecture, 11GB of global memory and compute capability 3.5.

- Nvidia Tesla C2070 - equipped with $14 \times 32 - core$ SMs, fermi architecture and has about 5GB of global memory, compute capability 2.0.

- Nvidia GeForce GTX 480 - consists of $7 \times 48 - core$ SMs, fermi architecture with 1GB global memory, compute capability 2.1.

Our discussion is organized as follows: First, we present a comparative evaluation of the four proposed GPU implementations focusing on the processing time. Second, we show how to tune the parameters of our most efficient implementation to achieve better performances (the WF-Tiled solution). Third, we discuss the impact of the data transfers on the overall performances. Fourth, we exploit double-buffering to overlap the computation and communication across sequences of images and calculate the frame rate of our proposed methods. We have extended our experiments utilizing multiple GPUs for large scale images. To conclude, we compare our GPU implementations using different GPU architectures with the multi-threaded CPU implementation. We tune all kernel functions to achieve the best performances.

### 5.2.1 Kernel Performance Evaluation

Figure 5.6 reports the cumulative kernel execution time of the four proposed GPU implementations on different image sizes for a 32-bin integral histogram. For readability, the data in the y-axis are reported in logarithmic scale. As it is obvious, due to its extremely poor GPU utilization, the CW-B approach performs extremely poor, and is outperformed by all other approaches by a factor in excess of 30X. CW-TiS outperforms CW-STS by a factor between 2X and 3X depending on the image size. Finally, WF-TiS leads to a further performance improvement up to about 1.5X over CW-TiS (Fig. 5.6).

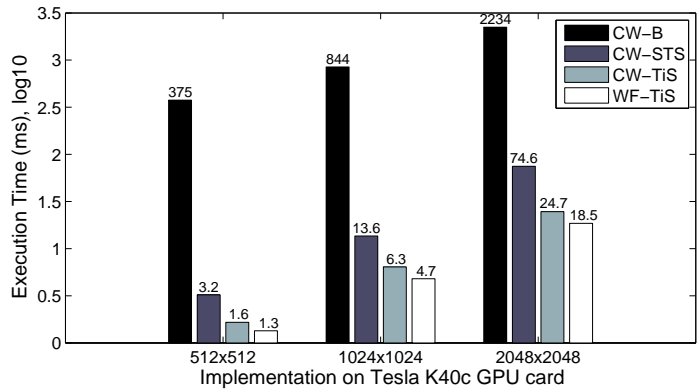

Figure 5.6: Cumulative kernel execution time of the four proposed GPU implementations on different image sizes for a 32-bin integral histogram on Tesla K40c.

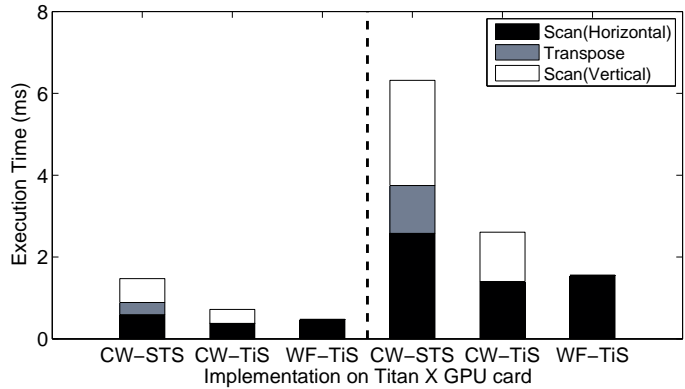

Figure 5.7: Kernel execution time for calculating integral histogram of size 512x512x32 (left), and 1024x1024x32 (right) on GTX Titan X.

To allow a better understanding of these results, Fig. 5.7 breaks down the execution times of different processing tasks. We can make the following observations. First, even when performed on large arrays (consisting of $b \times w \times h$ elements), the prescan kernel provided by CUDA SDK is less effective than the custom prescan kernel that we implemented in our CW-TiS and WF-TiS solutions. In fact, the execution time of our custom prescan kernel is comparable to that of the transpose kernel. Second, additional performance boost is achieved by merging the horizontal and the vertical scan into a single scan kernel, reducing the total number of global memory accesses and halving the required memory by removing the transpose phase.

## Performance Tuning of WF-TiS

In this section, we discuss the tuning of two important parameters used in our WF-

TiS design: the tile size and the thread block configuration. Our considerations can be generalized to other parallel kernels using tiling.

## Configuring the Thread Blocks

CUDA kernels can run with different thread block configurations. At runtime, each thread block is mapped onto a SM in a round-robin fashion. The execution of blocks mapped onto the same SM can be interleaved if their cumulative hardware resource requirements do not exceed those available on the SM (in terms of registers and shared memory). In case of interleaved execution, context-switch among thread blocks can help hiding global memory latencies. If interleaved execution is not possible, memory latencies can be hidden only by context switching within a single thread block. Therefore, properly setting the block configuration can help achieving better performances. NVIDIA provides a 'CUDA Occupancy Calculator' to assist the programmer in finding the kernel configuration that maximizes the resource utilization of the GPU. Although low occupancies (typically below 50%) indicate possible bad performances, a full occupancy does not ensure the optimal configuration. This fact is highlighted in Fig. 5.8, which shows the kernel execution time and the GPU occupancy using different thread block configurations. These results were reported on a 512×512 image and a 32-bin integral histogram. As can be observed(Fig. 5.8), both the best and the worst configurations, 512 and 1024 threads, in terms of execution time are characterized by a 100% GPU occupancy. In addition, the lowest execution time is achieved using 512-thread blocks.

## Configuring the Tile Size

The tile size determines the amount of shared memory used by each thread block. Increasing the tile size will reduce the number of iterations (or strips), and will increase

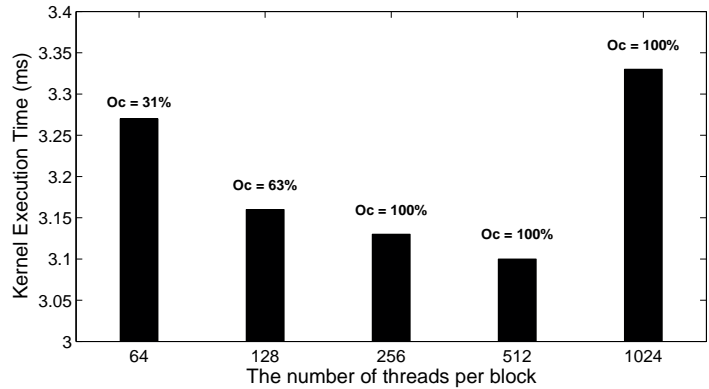

Figure 5.8: Kernel execution time and occupancy for different thread block configurations for an integral histogram of size 512x512x32 on Tesla K40c.

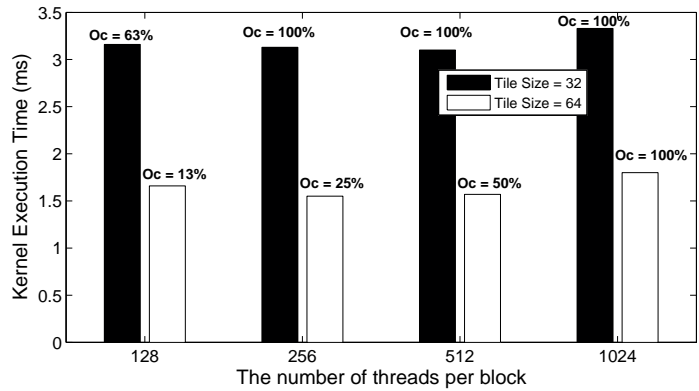

Figure 5.9: Performance evaluation of the WF-TS kernel for two tile sizes and several thread block configurations for an integral histogram of size 512x512x32 on Tesla K40c card.

the amount of work performed by each thread block. But larger tiles may limit inter-block parallelism and decrease the opportunity to hide global memory latencies by context switching across thread blocks. We tried different tile configurations ($16 \times 16$, $32 \times 32$, $64 \times 64$). However, in the case of $16 \times 16$, the performance is much worse than the two others since the tile data is processed linearly and each line is limited to only 16 elements. This causes that only half of the threads warp be active for each tile. Figure 5.9 reports a performance analysis of our WF-TiS implementation using two

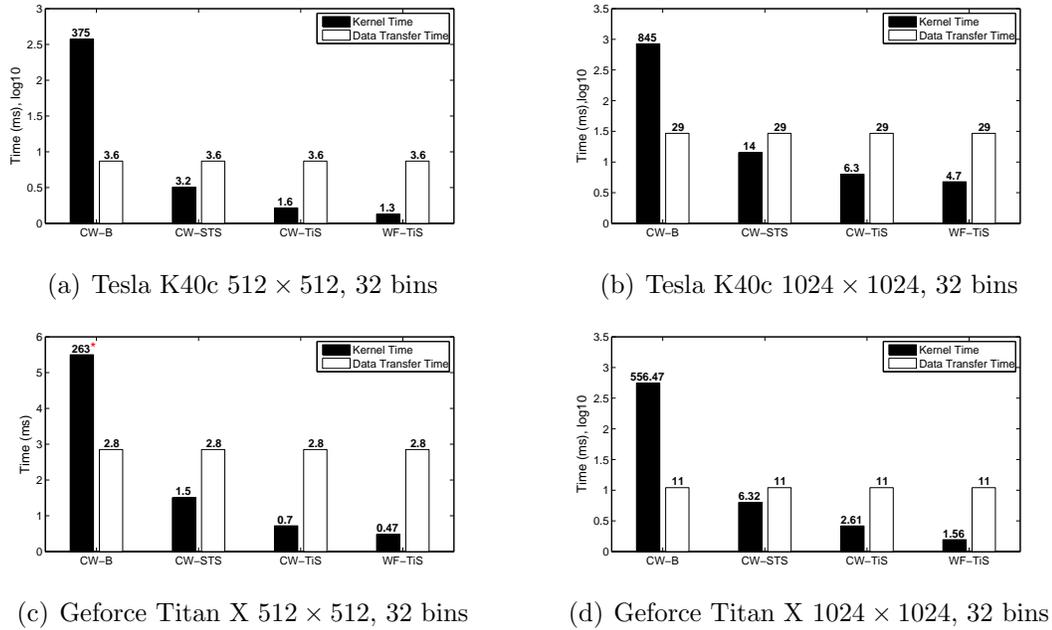

(a) Tesla K40c $512 \times 512$, 32 bins

(b) Tesla K40c $1024 \times 1024$, 32 bins

(c) Geforce Titan X $512 \times 512$, 32 bins

(d) Geforce Titan X $1024 \times 1024$, 32 bins

Figure 5.10: Kernel execution time versus data transfer time for
$64 \times 64$ tile configuration.

tile sizes ($32 \times 32$ and $64 \times 64$) and several thread block configurations. The $64 \times 64$ tile configuration performs better than $32 \times 32$ tile configuration by better use of the limited shared memory.

## 5.2.2 Communication Overhead Analysis

In this section, we discuss the overhead due to data transfers between CPU and GPU. The experiments were performed on Tesla k40c with kepler architecture and a Geforce GTX titan X with maxwell architecture.

Figure 5.10 shows the results. We make the following observations for both GPUs: the CW-B implementation is compute bound (that is, the kernel execution time is larger than the CPU-GPU data transfer time), whereas the other solutions are data-transfer-bound. These experiments suggest that further kernel optimizations may not be advisable. On the other hand, potential performance improvements may result from a reduction in the communication overhead coming from advances in interconnection technologies. However, it must be observed that the data transfer overhead is significant only when considering the integral histogram a stand-alone application. *In most cases, the integral histogram is part of a more complex image processing pipeline that can be implemented on GPU.* In these scenarios, since the integral histogram does not need to be transferred back to CPU, optimizing the kernel processing is still relevant.

## Overlap Communication and Computation Using Dual-Buffering for Sequence of Images

So far we have focused on distributing integral histogram data across available GPU cores. We can also leverage dual-buffering to overlap CPU-GPU communication and GPU computation across different images for further performance improvement. This can be accomplished by using CUDA streams along with page-locked memory and asynchronous data transfers between host and device. Algorithm 10 represents the pseudo-code of our dual-buffering based solution. All operations issued to stream 1

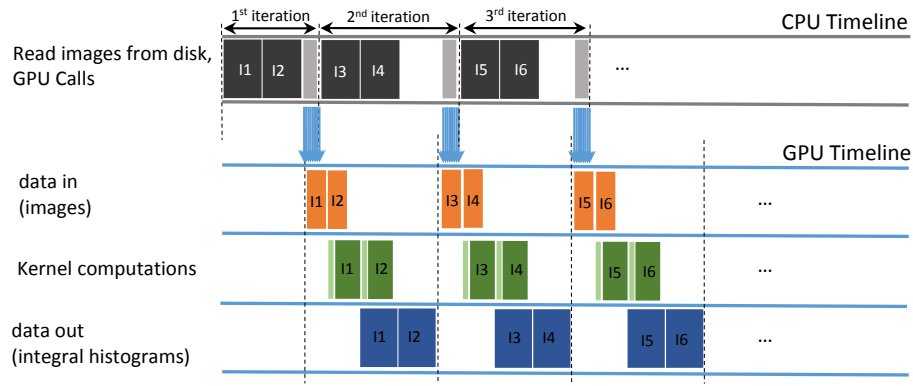

Figure 5.11: Pipeline parallelism: communication and computation overlap using dual-buffering.

**Algorithm 10:** Integral histogram computation on sequences of images using dual-buffering

```
 1: for all pair of images do
 2:     /* Read two images from disk to CPU */
 3:     h_Img1 = CopyImageFromDisk(Img1);
 4:     h_Img2 = CopyImageFromDisk(Img2);
 5:
 6:     /* Transfer two images from CPU to GPU */
 7:     cudaMemcpyAsync(d_Img1, h_Img1, stream1);
 8:     cudaMemcpyAsync(d_Img2, h_Img2, stream2);
 9:
10:     /* Generate and intialize the histograms on GPU */
11:     init_kernel(d_IntHist1, d_Img1, stream1);
12:     init_kernel(d_IntHist2, d_Img2, stream2);
13:
14:     /* Compute integral histograms */
15:     IntHistComputation_kernel(d_IntHist1, stream1);
16:     IntHistComputation_kernel(d_IntHist2, stream2);
17:
18:     /* Transfer integral histogram from GPU to CPU */
19:     cudaMemcpyAsync(h_IntHist1, d_IntHist1, stream1);
20:     cudaMemcpyAsync(h_IntHist2, d_IntHist2, stream2);
21: end for
```

are independent of those issued to stream 2. In each iteration, the following operations are performed: first, two images are read from disk and stored in page-locked host memory; second, those images are transferred to the GPU with asynchronous data transfers; then, the initialization and histogram computation kernels are launched; finally, the computed integral histograms are copied back to CPU (again, using asynchronous data transfers). We enqueue the memcpy and kernel execution operations breadth-first across streams rather than depth-first to avoid blocking the copies or kernel executions of a stream with another stream. This sequence of operations allows the effective overlapping of operations belonging to different CUDA streams. Figure 5.11 exploits pipeline parallelism to address the communication and computation overlap when computing 32-bins integral histogram for HD images. Our experiments show that the memory copies from disk to host are twice as fast as the GPU operations. Therefore, in our implementation, such copies are completely hidden behind the GPU tasks. Figure 5.12 shows the effect of dual-buffering on the frame rate for a sequence of 100 HD images (1280 × 720) using the WF-TiS kernel. As can be seen, dual-buffering improves the performance by a factor of two for 16-bins integral histogram computations. However, as the number of bins increases, the performance improvement decreases, and becomes negligible at 128 bins. This can be attributed to the fact that the use of page-locked memory on very large memory regions leads to performance degradation.

### 5.2.3 Integral Histogram for Large Scale Images Using Multiple GPUs

We have extended our WF-TiS kernel implementation to process large images (e.g. WHSXGA: 6400×4800). For such large images, limited GPU global memory becomes the bottle-neck since it can not hold the whole integral histogram with all bins. In our approach, bins are equally grouped into distinct tasks which are enqueued. Each time CPU picks one task from the queue and issues a kernel to the available GPU in the system. Each kernel computes the integral histogram of task bins using the WF-TiS approach. When a device is available, CPU will dispatch another task from the queue to GPU and meanwhile, results will be copied back to CPU. This iterative process will continue until the queue is empty.

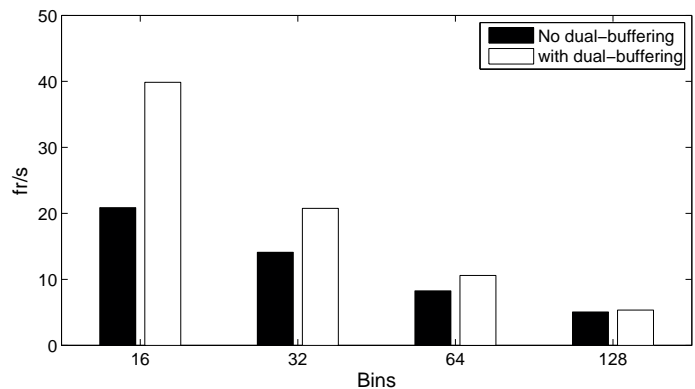

Figure 5.12: Effect of dual-buffering on the frame rate of a sequence of 100 HD images (1280 × 720) and different bin sizes. The experiments are conducted using the WF-TiS kernel.

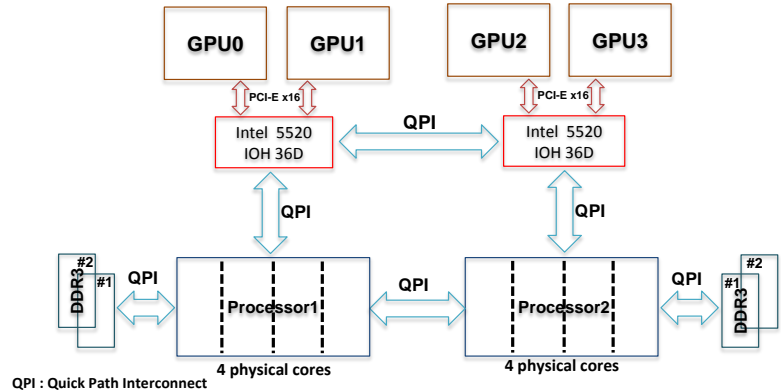

Figure 5.13: General block diagram of superserver configuration

- It is an easy-to-scale approach which can utilize multiple GPUs in the node

- The computation time (on GPU) is overlapped with communication time (copy result) via dual-buffering

- It can handle the imbalanced computation capability of heterogeneous system in which GPUs may have different hardware configuration

The experiments are conducted on our superserver equipped with 4 GTX 480 GPUs (Fig. 5.13). In our system, tasks will be distributed evenly. For instance, if there are 64 bins, each set of 16 bins will be performed on one of the GPUs. The portion of the

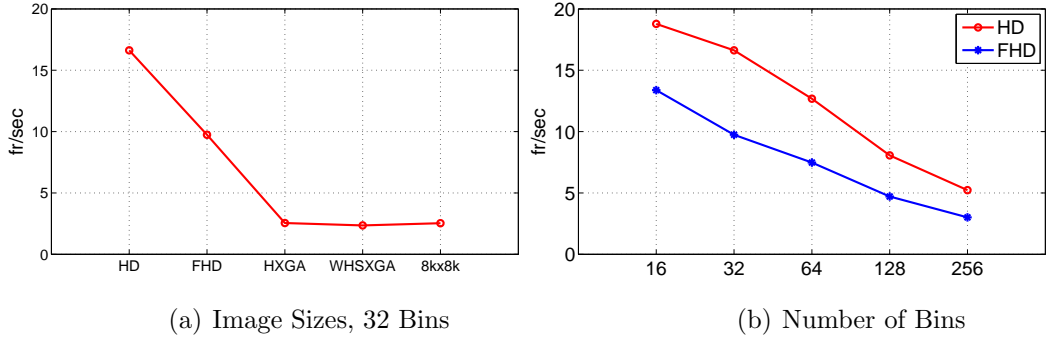

(a) Image Sizes, 32 Bins  (b) Number of Bins

Figure 5.14: Frame rate when computing (a) 32-bins integral histogram for different high definition large standard images (b) integral histogram for HD and FHD images with different number of bins using four Geforce GTX 480 GPUs.

16 bins that fit the GPU capacity will be performed in parallel. The most efficient WF-TiS kernel is invoked to compute the integral histogram for each bin.

Figure 5.14(a) shows the frame rate when computing 32-bins integral histogram for selected high definition image sizes (HD: $1280 \times 720$, FHD: $1920 \times 1080$, HXGA: $4096 \times 3072$, WHSXGA: $6400 \times 4800$ and 64MB: $8k \times 8k$). Figure 5.14(b) represents the frame rate for HD and FHD images with different number of bins.

Figure 5.17 shows the speedup of computing 128 bins integral histogram. The increasing speedup range from 3X (for HD) to 153X for the large 64MB images and 128 bins (total 32GB of 4 byte integer) over a single threaded CPU implementation is achieved. There are several advantages of this approach:

## 5.2.4 Frame Rate

The frame rate is defined as the maximum number of images processed per second. In this section, we compute the frame rate in the assumption that the integral histogram is a stand-alone application (rather than part of a more complex image processing

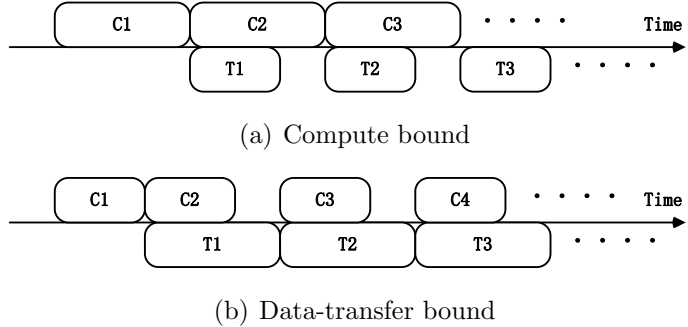

(a) Compute bound

(b) Data-transfer bound

Figure 5.15: Overlapping of computation and communication using double buffering for (a) compute bound and (b) data-transfer bound implementations. In the diagram, $Ci$ and $Ti$ shown in blocks represent kernel computation and data transfer, respectively, for the $i$th integral histogram

pipeline on GPU). We consider data transfers between CPU and GPU, but since we use double buffering, the processing of the current image and the transfer of the previously computed integral histogram from GPU to CPU can be overlapped (see Figure 5.15). Therefore, the frame rate equals to the $(cumulative\_kernel\_execution\_time)^{-1}$ for compute-bound implementations, and to $(data\_transfer\_time)^{-1}$ for data-transfer-bound ones. Fig. 5.16 $a$ and $b$ show the frame rate for different image sizes on the considered GPUs. The GTX Titan X allows faster data transfers between CPU and GPU. All data are reported on 32-bins integral histograms. In Fig 5.16 $a$ and $b$, the CW-STS, CW-TiS and WF-TiS implementations are data-transfer-bound for considered images. Figure 5.16 $c$ and $d$ show the frame rate reported on $512 \times 512$ images with different number of bins. As can be seen in Fig. 5.16$c$, the three best implementations are data-transfer-bound. Since increasing the number of bins means increasing the amount of data to be transferred (from GPU to CPU), the performances degrade linearly with the number of bins.

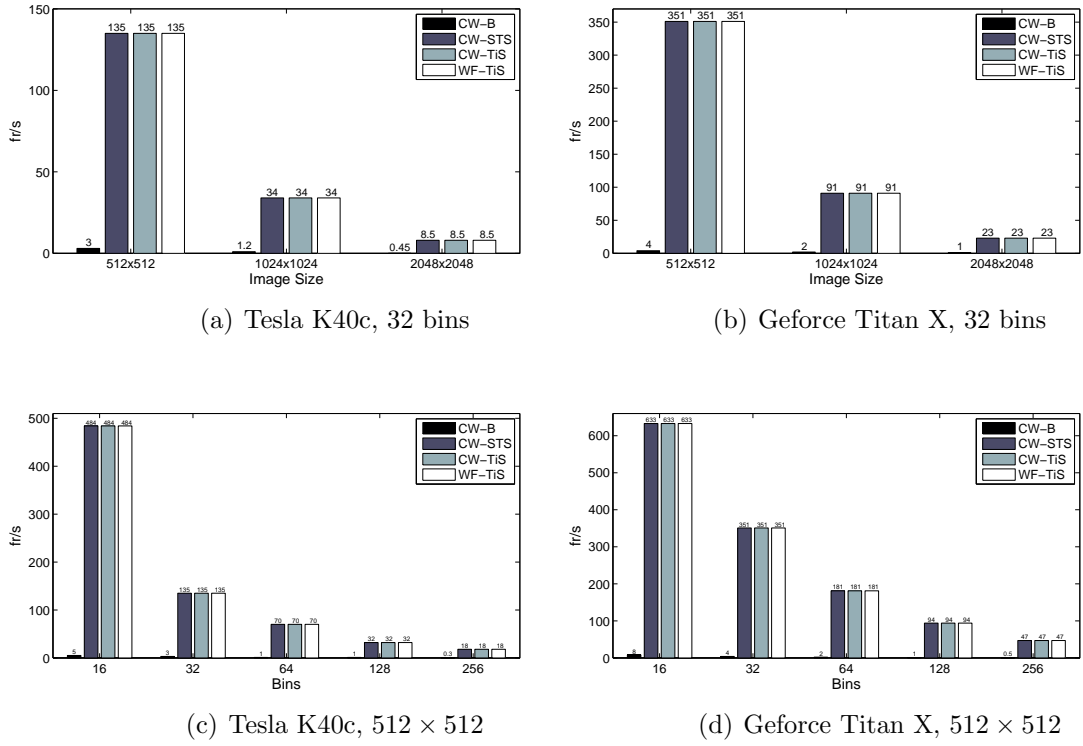

(a) Tesla K40c, 32 bins

(b) Geforce Titan X, 32 bins

(c) Tesla K40c, $512 \times 512$

(d) Geforce Titan X, $512 \times 512$

Figure 5.16: Frame rate for 32x32 tile configurations: (a) and (b) for different image sizes, (c) and (d) for $512 \times 512$ image and different numbers of bins.

### 5.2.5 Speedup over CPU

In this section, we report the speedup of our GPU implementation of the integral histogram over a parallel CPU implementation. The speedup is defined in terms of frame rate considering data transfers between CPU and GPU. As mentioned before, these data are conservative, since in most cases the integral histogram is part of a more complex image processing pipeline, which does not require transferring the computed integral histogram back to CPU.

Figure 5.18 shows the speedup of GPU over CPU for image sizes varying from 256x256 to 2048x2048 and 32-bins. The GPU implementation is run on Tesla K40c

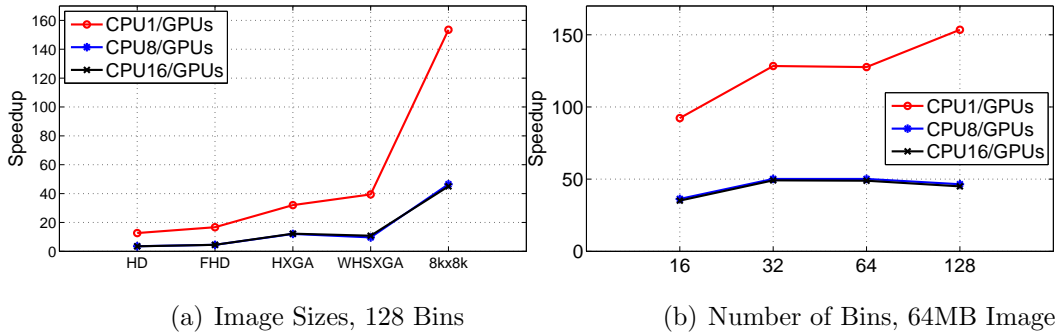

(a) Image Sizes, 128 Bins

(b) Number of Bins, 64MB Image

Figure 5.17: Speedup of integral histogram computations over a CPU implementation using different degrees of multi-threading for large scale images (up to 64MB) using four Geforce GTX 480

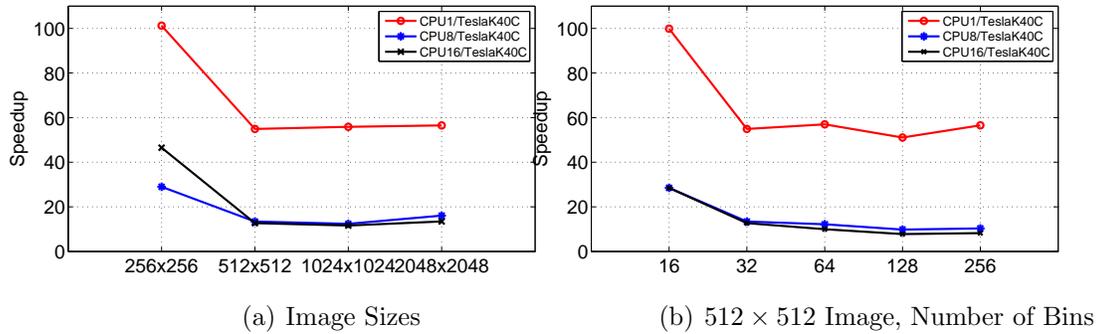

(a) Image Sizes

(b) $512 \times 512$ Image, Number of Bins

Figure 5.18: Speedup of the GPU designs over a CPU implementation using different degrees of multi-threading

GPU card. Our CPU implementation, parallelized using OpenMP, is run on an 8-core Intel Xeon E5620. Since the CPU cores are hyper-threaded, the best CPU configuration consists of 16 threads. Although the GPU implementation is data-transfer-bound for large images and/or number of bins, the speedup over the single threaded CPU implementation is about 60X, and over a 16-thread implementation, it varies between 8X and 30X.

In figure 5.19, we have compared our WF-TiS performances to the best performance of the IBM Cell/B.E integral histogram parallel implementation using wavefront(WF) and cross-weave(CW) scan mode [127] for standard $640 \times 480$ image size.

To calculate the frame rate of our implementation, we chose the maximum value between kernel time and data transfer time since dual buffering is applied. For our GPU implementation, in most cases the performance is data-transfer-bound: the achieved frame rate is limited by transferring the data between CPU and GPU over the PCI-express interconnect, rather than from the parallel execution on the GPU.

It is shown that the Titan X with maxwell architecture outperforms all the other GPU devices, CPU and the Cell/BE processors.

## 5.3 Conclusion

In this chapter, we have evaluated four GPU implementations of the integral histogram and proposed the fastest approach to accelerate computer vision applications utilizing integral histogram. All our designs – namely CW-B, CW-STS, CW-TiS, and WF-TiS – compute parallel cumulative sums on row and column histograms either in a cross-weave or in a wavefront scan. While CW-B and CW-STS kernels are based on pre-existing scan and transpose kernels, CW-TiS and WF-TiS are based on our custom scan kernels. Our progressive optimizations were based on a careful analysis of how each alternative leverages the hardware resources offered by the GPU architecture. Our kernel optimizations coupled with the use of dual-buffering allow us to achieve a frame-rate bounded by the data transfer time over PCI-Express connecting CPU and GPU for smaller images. In particular, when computing the 32-bin integral histogram of a $512 \times 512$ image, our most efficient implementation reached a frame rate of 351 fr/sec on an Geforce GTX Titan X graphics card. However, for a sequence of larger images (HD size), which were bounded by kernel execution time, the frame rate has been improved by a factor of two for 16 bins integral histogram using dual-buffering. For large scale images, mapping integral histogram bins computations on multiple GPUs enables us to process 32 giga bytes of image data with a frame rate of 0.73 Hz. These results strengthen the idea of high performance computing to distribute the data/compute intensive tasks between multiple nodes. Furthermore, our optimized WF-TiS kernel had a 60X speedup over a serial single-threaded CPU implementation for standard image size, and a 8X to 30X speedup over a multi-threaded implementation deployed on an 8-core CPU server.

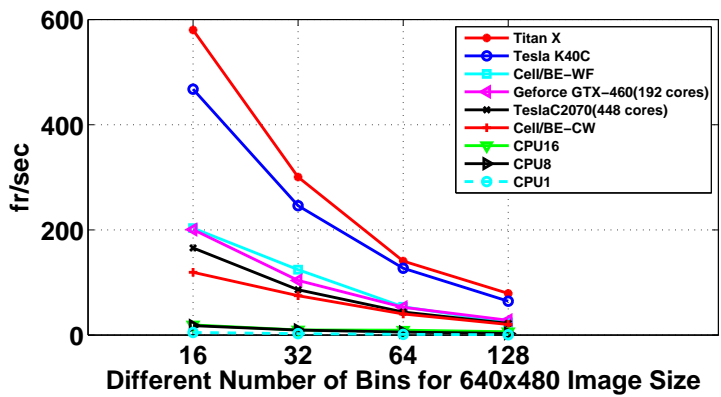

Figure 5.19: Frame rate performance comparison of proposed GPU WF-TiS design versus CPU implementation using different degrees of multi-threading, CPU1, CPU8, CPU16 and Cell/B.E. performance results presented for wavefront(WF) and cross-weave(CW) scan mode using 8 SPEs in[127].

We have extended the GPU integral histogram approach to design and implement spatio-temporal median filter algorithm for fast motion detection in full motion video, as well as for 3D face reconstruction texturing.

# Chapter 6

# Conclusion and Future Research

## 6.1 Summary

Disciplined or informed intelligent fusion of different kinds of information is useful for a general purpose tracking system across modalities and different computer vision tasks. We present a collaborative tracking system consists of a master tracker and two auxiliary trackers. The main idea is to have a pool of trackers that are working together in an intelligent fusion framework to improve tracking performance by being called dynamically. The input of the systems varies from a standard definition video to very large scale airborne imagery collected over urban areas. The output will be target tracklets that are computed using object visual features and object temporal motion information. The visual feature-based tracker usually takes the lead as long as object is visible and presents discriminative visual features. Otherwise, tracker will be assisted by motion information. Motion prediction will be used to localize the object when being partially or fully occluded by trees or tall structures. The estimated motion detection mask can be fused intelligently with visual object features to increase tracking localization accuracy or being applied to initialize and perform persistent multi-object tracking. The main objective of the presented tracker is to accommodate to object appearance changes due to scale, pose, orientation and illumination and perform persistent tracking under background noises (clutter, dynamics) and occlusion as well as camera motion effects. Two measures are used to analyze the performance of the visual tracking: Accuracy and Robustness. Accuracy is the average overlap between the predicted and ground truth bounding boxes during successful tracking periods. Robustness measures the number of times that tracker loses the target during tracking. We weight robustness more than accuracy since the ultimate goal of visual tracking is performing persistent tracking. It is also required to achieve real-time performance on low-power computing platforms (laptops, PCs).

Image spatial context can be modeled as a hierarchy of abstractions by increasing the spatial scale. We utilized image spatial context at different level to make our video tracking system resistant to occlusion and background noise and improve target localization accuracy and robustness. Pixel-level spatial information are used to build intensity spatially weighted histogram or compute object foreground and background color histogram tensor. Spatial layout of image fragments are preserved when constructing the spatial pyramid of HoG. The structure-level spatial context (i.e. road network, building maps) can be applied to filter out the false object detections by distinguishing background from moving objects in full motion videos. Therefore, our proposed visual tracker is named Spatial Pyramid Context-aware Tracker (SPCT). We chose a pre-selected seven-channel complementary features including RGB color, intensity and spatial pyramid of HoG to encode object color, shape and spatial layout information. Integral histogram is the building block to encode candidate regions feature information and achieve fast, multi-scale local histogram computation in constant time. A novel fast algorithm is presented to accurately evaluate spatially weighted local histograms in constant time complexity using an extension of the integral histogram method.

The experiments on extensive VOTC2016 benchmark dataset and aerial video confirm that combining complementary tracking cues in an intelligent fusion framework enables persistent tracking for Full Motion Video (FMV) and Wide Aerial Motion Imagery (WAMI). SPCT ranked 11 among 62 trackers based on achieved average robustness of 1.3 and accuracy of 0.458 on all VOTC2016 60 sequences. SPCT ranked 1 among all trackers with low average robustness of 0.789, accuracy of 0.562 and missing frame rate as low as 0.4% on Argus WAMI dataset and robustness of 0.325, accuracy of 0.628 and low average MFR of 0.005 on ABQ aerial urban imagery.

We proposed a multi-component framework based on semantic fusion of motion information with projected building footprint map to significantly reduce the false alarm rate in urban scenes with many tall structures. It was shown that using purely conventional motion detection methods would not be sufficient for a wide area aerial imagery in which there are strong traces of parallax induced by tall buildings. In order to reject undesirable detections due to tall structures we used depth map information – obtained from the fast SFM followed by a dense 3D point clouds algorithm – in a boundary refinement and filtering processing stage. Using the proposed fusion approach a high average precision of 83.9% and recall of 75.1% is achieved using 3D median background modeling which promises a reliable persistent tracking.

## 6.2 Contributions

The main contribution of the work are summarized as follows:

- **Pool of trackers with a smart context-based fusion scheme**: A collaborative tracking system consists of a master tracker and two auxiliary trackers is developed using multi-channel Features predictors and temporal motion information.

- **Spatial pyramid appearance tracking:** we utilize spatial Pyramid of Histogram of Gradient Orientation (PHoG) to encode object local shape and spatial layout of the shape so that to make tracking resistant to occlusion and invariant to illumination changes.

- **Spatially weighted local histograms in O(1) using weighted integral histogram:** we proposed a novel fast algorithm to accurately evaluate spatially weighted local histograms in O(1) time complexity using an extension of the integral histogram method (SWIH) that encode both spatial and feature information.

- **Parallel GPU implementation of integral histogram:** we utilize integral histogram as the building block to encode candidate regions feature information and achieve fast, multi-scale histogram computation in constant time. The custom implementation WF-TiS achieves a frame rate of 135 $fr/sec$ on Tesla K40c and 351 on a Geforce GTX Titan X graphics card when computing the 32-bin integral histogram of a $512 \times 512$ image. Furthermore, the GPU WF-TiS design reports a 60X speedup over a serial CPU implementation, and a 8X to 30X speedup over a multithreaded implementation deployed on an 8-core CPU server [44, 31].

- **Context-based semantic fusion of motion information with projected building footprint information:** we proposed a multi-component framework based on semantic fusion of motion information with projected building footprint information to significantly reduce the false alarm rate in urban scenes with many tall structures. Moving object detection in wide-area aerial imagery is very challenging since fast camera motion prevents direct use of conventional moving object detection methods and strong parallax induced by tall structures in the scene causes excessive false detections [45, 46].

- **Orientation-Aligned Template Matching by Learning the Object Direction:** Experimental results show that most of the orientation sensitive features fail to detect the object when object template and search window are not aligned for example when computing features likelihood maps using normalized cross correlation of target template and search window. I proposed an orientation-aligned template matching particularly for vehicle detection in wide aerial imagery using vehicle's non-holonomic constraints.

- **Target object initialization refinement using CAMSHIFT:** If the ground truth bounding box annotated around the object is not tight, not oriented aligned or not centered around the object (drifted), it will contain background information that will be incorporated into object descriptors which is not desirable. Incorporating background information will lead to less accurate target localization and rapidly loss of the target being tracked. We used the Contin-uously Adaptive Mean Shift (CAMSHIFT) algorithm to partially correct the drifted and loose ground truth bounding box and improve tracking robustness.

- **Offline feature selection test-bed using tracking context:** A separate test-bed is developed for filtering-based feature selection in order to decouple feature performance from the rest of the tracking system where the final outcome depends not only on the features used but also on the other parameters like the predictor performance. Based on this experiment, a 7-channel complementary features including RGB(3), gradient orientation and magnitude (2) and edges(1) are chosen to characterize the object appearance model [25, 27].

- **Automatic Detection of candidate regions using motion prediction:** Automatic prediction of ROI in a complex image or video is a key task for visual tracking that enables fast search and avoids background clutter, particularly for large scale aerial imagery. When target motion dynamics is linear or approximately linear during the intervals between observations then a motion prediction filter like Kalman filter can be used to automatically determine the search window in the next image.

- **Top performance on benchmark datasets including VOTC2016 and WAMI data:** SPCT ranked 11 among 62 trackers based on achieved average robustness of 1.3 and accuracy of 0.458 on all VOTC2016 60 sequences. SPCT ranked 1 among all trackers with low average robustness of 0.789, accuracy of 0.562 and missing frame rate as low as 0.4% on Argus WAMI dataset and robustness of 0.325, accuracy of 0.628 and low average MFR of 0.005 on ABQ aerial urban imagery.

## 6.3 Future Research

Many of the moving object detection and tracking algorithms are sensitive to orientation and scale variations and usually fail to accommodate to visual appearance changes due to pose, orientation or scale changes. Accurate and robust scale and orientation estimation is still a challenging problem in visual tracking. A reliable orientation estimation would significantly improve the feature likelihood map computation and consequently the target localization accuracy. Orientation information can be used to provide orientation aligned tracking results rather than traditional axis aligned bounding boxes. An accurate scale estimation is also required to improve SPCT accuracy performance.

We presented an orientation aligned template matching by learning the direction of the vehicle using four angles including $\{0°, 90°, 180°, 360°\}$. However, orientation aligned matching performance can be improved by considering more angles for alignment and incorporating more information including mean and standard deviation of the orientation estimations history and getting feedback from motion prediction.

Online feature selection is the next component that can be integrated into SPCT to adaptively select the most discriminative features along image sequences.

In the past few years, discriminative trackers based on deep learning techniques outperform the generative trackers performance. One of the immediate directions of this work is to employ learning algorithms to improve the performance of tracking cues particularly on aerial imagery.

# VITA

Mahdieh Poostchi was born in Mashhad, Iran. After earning her high school diploma in mathematics in 2000, she attended Azad University of Mashhad where she got her B.Sc in Computer Science. She received her M.Sc in Artificial Intelligence and Robotics from Iran University of Science and Technology in 2009 while serving as an instructor for several high educational institute. After completing her M.Sc, she worked for the ITS department of the Mashhad municipality in Iran. In 2011, she got admitted to the doctoral program at Computer Science Department, University of Missouri-Columbia. She started working as a research assistant in Computational Imaging and Visualization Analysis Laboratory supported by grants from the US Air Force Research Lab, NIH, NGA, NASA and NSF. She participated UCLA IPAM Graduate Computer Vision Summer School in 2013. She had an internship in 2014 with Metaio, an augmented reality company that has since been purchased by Apple. She spent the Summer of 2015 as an intern research fellow at the National Library of Medicine NIH conducting research on microscopy image analysis of blood smears. She received her second Master degree from University of Missouri-Columbia in 2016. Her current research interests include image/video processing, computer vision, machine learning and high performance computing with emphasis on moving object detection and tracking for full motion video and wide aerial imagery. She is a member of the IEEE, UPE, CSGSC, ACM women and Computer Vision Foundation. She has served as a referee for international journals and conferences.

She received her Ph.D. degree in computer science from the university of Missouri-Columbia in 2017.